\documentclass[journal]{IEEEtran}
\usepackage[colorinlistoftodos]{todonotes}

\usepackage{subcaption}

\ifCLASSINFOpdf
\else
\fi
\usepackage{url}
\usepackage{color,soul}

\renewcommand\hl[1]{#1} % <<<  Makes highlighting invisible

\begin{document}
\title{Hybrid Encoding For Generating Large Scale Game Level Patterns With Local Variations} % Using a GAN}

\author{Jacob Schrum,
        Benjamin Capps,
        Kirby Steckel,
        Vanessa Volz,
        and~Sebastian Risi% <-this % stops a space
\thanks{J.~Schrum, B.~Capps, and K.~Steckel are from Southwestern University in Georgetown, TX USA. Corresponding author J.~Schrum ({\texttt schrum2@southwestern.edu}) is an Associate Professor, B.~Capps and K.~Steckel are recent graduates. V.~Volz is an AI researcher at modl.ai and an Honorary Lecturer at Queen Mary University London, UK, and S.~Risi is a co-founder of modl.ai and a Full Professor at ITU Copenhagen in Denmark.}}


\markboth{IEEE Transactions on Games}%
{J.~Schrum, B.~Capps, K.~Steckel, V.~Volz, and S.~Risi}



\maketitle

\begin{abstract}
Generative Adversarial Networks (GANs) are a
powerful indirect genotype-to-phenotype mapping for evolutionary search.
\hl{Much previous work applying GANs to level generation focuses on 
fixed-size segments combined into a whole level, but individual
segments may not fit together cohesively.
In contrast, segments in human designed levels are often repeated,
directly or with variation, and organized into patterns 
(the symmetric eagle in Level~1 of \emph{The Legend of Zelda}, 
or repeated pipe motifs in \emph{Super Mario Bros}).} 
Such patterns can be produced with Compositional Pattern
Producing Networks (CPPNs). CPPNs define
latent vector GAN inputs as a function of geometry, 
\hl{organizing} segments output by a GAN into 
complete levels. However, collections of latent vectors can also be
evolved directly, producing more chaotic levels. We propose
a hybrid approach that evolves CPPNs first, but allows
latent vectors to evolve later, combining the benefits of
both approaches. These approaches are evaluated in \emph{Super Mario
Bros}.~and \emph{The Legend of Zelda}. We previously demonstrated via %divergent search (MAP-Elites)  
a Quality-Diversity algorithm that CPPNs better cover the space
of possible levels than directly evolved levels. Here, we show that
the hybrid approach \hl{(1)} covers areas that neither of the other
methods can, and \hl{(2)} achieves comparable or superior QD scores.
\end{abstract}


% \fi
\IEEEpeerreviewmaketitle

\section{Introduction}
\IEEEPARstart{G}{enerative} Adversarial Networks (GANs \cite{goodfellow2014generative}), a type of generative neural network trained in an unsupervised way, are capable of reproducing certain aspects of a given training set. For example, they can generate diverse high-resolution samples of a variety of different image classes \cite{brock2018large}.

Several recent works \hl{show} that it is possible to learn the structure of video game levels using GANs \cite{volz:gecco2018,park:cog19,torrado2019bootstrapping,gutierrez2020zeldagan}, but these approaches only generate small level segments.
\hl{GANs have also generated levels of arbitrary size {\cite{awiszus:aiide2020}}, {\cite{awiszus:cog2021}}, but global patterns between segments, i.e.~symmetry and repetition, are difficult to capture with purely GAN-based approaches {\cite{volz2020capturing}}.}

%In contrast, complete game levels can consist of many segments, sometimes repeated, often with variation. A valid question regarding such \hl{segment-based GAN} approaches is whether they can scale to generate arbitrarily large artefacts that have a modular structure, such as these \emph{complete} game levels. 

%% Removed: GAN-based generation techniques 

%To scale up 
%from generating level segments
%to generating complete levels, 

%\todo{add "with regularities"?}

\hl{To generate complete levels with global patterns,} we combine Compositional Pattern Producing Networks (CPPNs \cite{stanley:gpem2007}) with GANs.
A CPPN is a special type of neural network that generates patterns with regularities such as symmetry, repetition, and repetition with variation. CPPNs have succeeded in many domains \cite{secretan:ecj2011,hoover2012generating,clune:ecal11,cellucci20171d,hastings2009automatic,tweraser:gecco2018}. 

\begin{figure*}[t]
\centering
\includegraphics[width=0.9\textwidth]{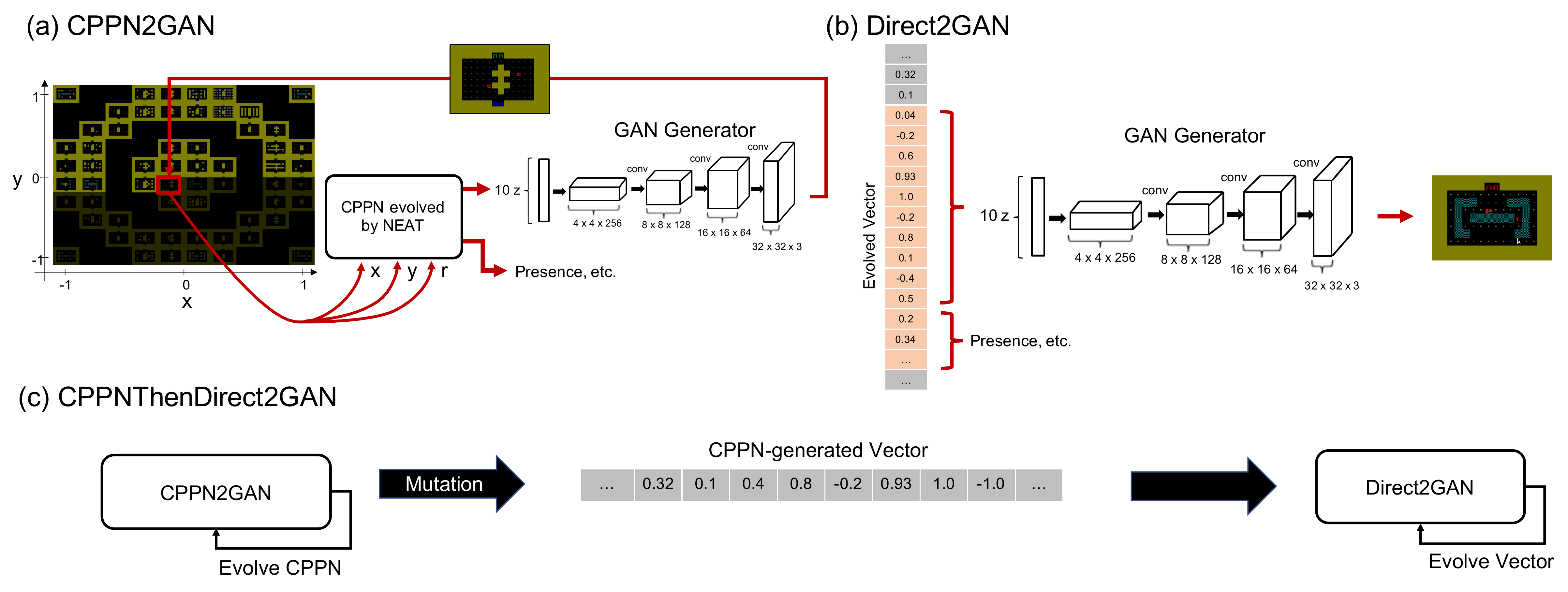}
\caption{\textbf{Three Genotype Encodings Applied to Zelda}. \normalfont (a)
%\caption{CPPN2GAN applied to Zelda. 
In \textbf{CPPN2GAN}, the CPPN takes as input the Cartesian coordinates of a level segment ($x,y$) and its distance from the center $r$, and for each segment produces a different latent vector $z$ that is fed into the generator of a GAN pre-trained on existing level content. 
The CPPN also outputs additional information determining whether the room \hl{is present}, %how its doors connect to other rooms, 
\hl{door placement}, and other miscellaneous information.
The approach captures patterns in the individual level segments, but also creates complete maps with global structure, \hl{e.g}.\ imperfect radial symmetry as above. (b) In \textbf{Direct2GAN}, one real-valued vector is evolved, which is chopped into %individual vectors to generate each tile. 
\hl{one vector per segment.}
A part of the vector (size 10) is input into the GAN, while additional values specify the presence of a room and \hl{miscellaneous information}. %additional room-specific properties. 
(c) Finally, in \textbf{CPPNThenDirect2GAN}, vectors are first generated by evolving CPPNs (CPPN2GAN), facilitating the discovery of global patterns, and then later switched to evolving a vector encoding (Direct2GAN) that allows for local variations.}
\label{fig:overview}
\end{figure*}

Our approach, CPPN2GAN, was first introduced in 2020 \cite{schrum:gecco2020cppn2gan}. 
It is a doubly-indirect encoding. An evolved CPPN represents a pattern on the geometry of a level. The CPPN maps coordinates of level segments to vectors in a latent space, where the next level of indirection occurs. These vectors are fed as inputs to a pre-trained GAN, which outputs the level segment that belongs at the specified location (Fig.~\ref{fig:overview}).

This paper goes further by hybridizing CPPN2GAN with directly evolved latent vector inputs, a method we called Direct2GAN. The new approach takes inspiration from HybrID \cite{CluneBPO09:HybrID}, and begins evolution with CPPN genomes, but lets them transition into latent vector genomes to gain the benefits of both approaches. In fact, starting with a CPPN-based focus on global patterns, and only later switching to a vector encoding that allows for local variations, can discover solutions that would not be discovered by either approach in isolation. This combination approach is called CPPNThenDirect2GAN.

We evaluate the three methods using several different evaluation metrics proposed in previous work, and consider both performance as well as their ability to cover the space of possible solutions. We thus consider the ability of the different methods to optimize for specific characteristics (controllability), as well as to generate diverse levels, both important qualities of level generators \cite{pcgbook12}. The Quality Diversity algorithm MAP-Elites \cite{mouret:arxiv15} is used to conduct these evaluations.
Our experiments reaffirm our previous results \cite{schrum:gecco2020cppn2gan} with CPPN2GAN and Direct2GAN
in \emph{Super Mario Bros.}~and \emph{The Legend of Zelda}, and we add results both with an additional diversity characterization for each game, and with
the newly proposed hybrid CPPNThenDirect2GAN approach.\footnote{Code available at \url{https://github.com/schrum2/GameGAN}}
Results show that Direct2GAN is usually inferior to CPPN2GAN, and always inferior to the new CPPNThenDirect2GAN approach in terms of level \hl{fitness} and coverage of design space. CPPNThenDirect2GAN is as good or better than CPPN2GAN depending on the game and diversity characterization.

Ultimately, CPPN2GAN and CPPNThenDirect2GAN could be relevant not only for game levels, but other domains requiring large-scale pattern generators, e.g.\ texture generation, neural architecture search, computer-aided design, etc.

\section{Previous and Related Work}

%%% Will remove if necessary
This paper combines Generative Adversarial Networks (GANs) and Compositional Pattern Producing Networks (CPPNs) into a new form of Latent Variable Evolution (LVE).

%\subsection{Compositional Pattern Producing Networks}
\subsection{CPPNs} % Jacob: The word wrap of the full name is just too ugly, and wastes space
\label{sec:cppnsAndNEAT}

Compositional Pattern Producing Networks (CPPNs \cite{stanley:gpem2007}) are %artificial 
\hl{neural networks with various} %varying 
activation functions. %per node. 
They are repeatedly queried across a geometric space of %possible 
inputs and are %thus 
well suited to generating %geometric 
patterns, \hl{e.g}.\ %. For example, 
a CPPN can generate an %2D 
image by taking pixel coordinates $(x,y)$ as input and outputting intensity values for each corresponding pixel.

%(Fig.~\ref{fig:cppn}).\todo{figure caption: There is no "f" in the figure. This is kind of confusing. Maybe use $CPPN(x,y)$ instead?}

CPPNs typically include %many different 
\hl{various} activation functions biased towards specific patterns and regularities, \hl{e.g}.\ % . For example, 
a Gaussian function %allows a CPPN output pattern to be symmetric, 
\hl{for symmetry} and a periodic \hl{sine function for} %creates 
repeating patterns. %Other patterns, such as 
\hl{Repetition} with variation (e.g.\ fingers of a hand) can be created by combining functions (e.g.\ sine and Gaussian). CPPNs have been adapted to produce a variety of patterns in domains such as 2D images \cite{secretan:ecj2011}, musical accompaniments \cite{hoover2012generating}, 3D objects \cite{clune:ecal11}, animations \cite{tweraser:gecco2018}, physical robots \cite{cellucci20171d}, particle effects  \cite{hastings2009automatic}, and flowers \cite{risi2012combining}. 

%%% Jacob: I think this is redundant given Figure 1.
%\begin{figure}
%\centering
%\begin{subfigure}{0.3\textwidth} 
%    \includegraphics[width=1.0\textwidth]{Figures/new_gpem_funcpat2.pdf}
%    \caption{Mapping}
%\end{subfigure}
%\begin{subfigure}{0.17\textwidth}
%    \includegraphics[width=1.0\textwidth]{Figures/new_2d_cppn.pdf}
%    \caption{Composition}
%\end{subfigure}
%\caption{A CPPN encoding example for 2D images. \normalfont (a) The CPPN takes arguments $x$ and $y$ as inputs, which are coordinates in a two-dimensional space. A spatial pattern is generated by drawing the point at position $(x,y)$ with the corresponding intensity from output $f(x,y)$.
%When all the coordinates are drawn with an intensity corresponding to the output of $f$, the result is a spatial pattern.
%(b) The CPPN is essentially a graph that determines which functions are connected. The connections are weighted such that the output of a function is multiplied by the weight of its outgoing connection}
%\label{fig:cppn}
%\end{figure}

%The benefit of NEAT is that it optimizes both the neural architecture and weights of the network at the same time.

CPPNs are traditionally optimized through NeuroEvolution of Augmenting Topologies (NEAT \cite{stanley:ec02}). 
\hl{NEAT optimizes both the neural architecture and weights of networks at the same time. The population starts simple (inputs directly connected to outputs), but mutations later add nodes and connections. NEAT also allows for %efficient 
crossover between structural components with a shared origin.}
More recently, CPPN-inspired neural networks have also been optimized through gradient descent-based approaches  \cite{ha:iclr2017,fernando2016convolution}.  

%%%% Replaced with below
%While CPPNs can create patterns with complex regularities, training CPPNs to recreate particular images is difficult \cite{woolley2011deleterious}. However, GANs do not share this weakness. 
\hl{Evolved CPPNs can create global patterns with complex regularities, but struggle with precise local variations. GANs trained via gradient-descent do not have this problem.}

%\hl{While evolved CPPNs can create high-level patterns with complex regularities, they have trouble creating precise low-level variations. However, GANs trained via gradient-descent do not have this problem.}

\subsection{Generative Adversarial Networks}
\label{sec:gan}
% Jacob: Seems to have been sufficiently shortened/changed
%\todo[inline]{Content below is copied from Mario GAN paper, and is currently also in the non-CPPN interactive evolution paper that will also be submitted to this GECCO. We should change the text a bit.}

The training process of Generative Adversarial Networks (GANs \cite{goodfellow2014generative}) is like a two-player adversarial game in which a generator $G$ % (faking samples decoded from a random noise vector)
and a discriminator $D$ %(distinguishing real/fake samples and outputting 0 or 1)
are trained at the same time by playing against each other. The discriminator $D$'s job is to classify samples as being generated (by $G$) or sampled from the training data. %real.
The discriminator aims to minimize classification error, but the generator tries to maximize it. % at maximizing that probability.
Thus, the generator is trained to deceive the discriminator by generating samples that are indistinguishable from the training data. % good enough to be classified as genuine.
After training, the discriminator $D$ is discarded, and the generator $G$ is used to produce novel outputs that capture the fundamental properties present in the training examples. Input to $G$ is \hl{usually} some fixed-length vector from a latent space.

Generating content by parts, as is done by our CPPN2GAN approach, has also been investigated with conditional GANs (COCO-GAN~\cite{lin2019coco}). However, in contrast to our approach, in which the CPPN \hl{maps} coordinates of segments to latent vectors (generating patterns with regularities such as repetition and symmetry), COCO-GAN is conditioned on fixed coordinates with a common latent vector to produce segments of an image. \hl{Convolutional GANs have also been used to generate game levels of arbitrary size out of patches in both Mario {\cite{awiszus:aiide2020}} and Minecraft {\cite{awiszus:cog2021}}. Alternative generators, such as Variational Autoencoders {\cite{sarkar:fdg2020}} and Bayes nets {\cite{summerville2015sampling}}, have also been used to sequentially generate levels one segment at a time.} 

%\todo{added Bayes nets}

%a means of searching within the real-valued latent space of the GAN is needed, as described next. 

For a properly trained GAN, %randomly 
sampling latent vectors produces outputs that could pass as real \hl{content.} % images or levels. 
However, to find content with certain properties (\hl{i.e}.\ specific game difficulty, number of enemies), the latent space needs to be searched. %, as described in the next section.

%\subsection{Latent Variable Evolution}

%The first latent variable evolution (LVE) approach was by Bontrager et al.~\cite{bontrager2017deepmasterprint}, who

The first latent variable evolution (LVE) \hl{approach~{\cite{bontrager2017deepmasterprint}} trained} a GAN on %real 
fingerprint images and used \hl{evolution} %ary search 
to find latent vectors \hl{matching} %that match with 
subjects in the dataset.
Our previous work \cite{schrum:gecco2020cppn2gan} introduces the first indirectly encoded LVE approach. Instead of searching %directly 
for latent vectors, parameters for CPPNs are sought. These CPPNs %can 
generate a variety of %different 
latent vectors conditioned on the locations of level segments.

%Generating content by parts, as is done by our CPPN2GAN approach, has also been investigated with conditional GANs (COCO-GAN~\cite{lin2019coco}). However, in contrast to our approach, in which the CPPN can learn how to map coordinates of parts to latent vectors (facilitating generating patterns with regularities), COCO-GANs use fixed coordinates with a common latent vector to produce components of an image.

%the coordinates that COCO-GANs are conditioned on stay fixed.

\section{Video Game Domains}

%The games in this paper rely on data 

\hl{Game data comes} from the Video Game Level Corpus (VGLC \cite{summerville:vglc2016}). GAN models \hl{from previous work in \emph{Super Mario Bros.}~{\cite{Volz19}} and \emph{The Legend of Zelda} {\cite{gutierrez2020zeldagan}} are used because they 
are popular representatives of two distinct genres. Mario is a platformer with linear levels, and Zelda is a dungeon crawler with 2D levels. Thus different types of patterns are required in each game, demonstrating
the broad applicability of % the 
CPPNThenDirect2GAN.} % approach}. %, though in each case some specialized processing of the data is required.

\subsection{Super Mario Bros.}
\label{sec:mario}

\emph{Super Mario Bros.}~(1985) is a platform game that involves moving left to right while running and jumping. Levels are visualized with the 
Mario AI framework.\footnote{\url{https://github.com/amidos2006/Mario-AI-Framework}} 

%At the start of evaluation, Mario is in his \emph{fire flower} state, which means he can launch fire balls and is the size of two tiles on screen. Contact with an enemy reverts him to \emph{big} state (no more fire balls), and subsequent contact shrinks him to \emph{small} state, which takes up only one tile. Mario has access to more areas on screen when he is \emph{small}, but any further enemy contact causes him to die, ending evaluation.
%Past work \cite{volz:gecco2018} used a skilled agent from the Mario AI Competition \cite{togelius:cig2010:marioAI} to evaluate the generated levels, but due to the high computational cost of such simulation, this paper applies A* search to the tile-based level representation.

The tile-based level representation from VGLC uses a particular character symbol to represent each possible tile type. The encoding is extended to more accurately reflect the data in the original game, for example by adding different enemy types. % (Table \ref{tab:mariotiles}). For example, VGLC does not distinguish between different enemy types. Besides adding symbols for more enemy types, 
The representation of \emph{pipes} is adjusted to avoid the broken \emph{pipes} seen in previous work \cite{volz:gecco2018}.
Instead of using four different tile types for a \emph{pipe}, a single tile is used as an indicator for the presence of a \emph{pipe} and extended automatically downward as required. A detailed explanation of all modifications made to the encoding %as well as a full description 
can be found in work by Volz \cite[Chap. 4.3.3.2]{Volz19}. %, and a visual representation of how tiles are mapped is in Table \ref{tab:mariotiles}.

%\end{comment}

\subsection{The Legend of Zelda}
\label{sec:zelda}

\emph{The Legend of Zelda} (1986) is an action-adventure dungeon crawler. The main character, Link, explores dungeons full of enemies, traps, and puzzles. In this paper, the game is visualized  with an ASCII-based Rogue-like game engine used in previous work \cite{gutierrez2020zeldagan}. Details on the mapping between original VGLC game tiles and Rogue-like tiles can also be found there.% The mapping between original VGLC game tiles and Rogue-like tiles is in Table~\ref{tab:zeldatiles}.

Previous work \cite{gutierrez2020zeldagan} reduced the large set of tiles inherent to Zelda to a smaller set based on functional requirements. Some Zelda tiles differ in purely aesthetic ways, and others rely on complicated mechanics not implemented in the Rogue-like. 
The reduced set of tile types is as follows: regular floor tiles, impassable tiles, and tiles that enemies can pass, but Link requires a special item to pass (raft item to cross water tiles).
%The Rogue-like accommodates three tile types:
%a floor tile which directly corresponds to the Zelda floor tile, an impassable tile that corresponds to all impassable tiles in Zelda, and a water tile that corresponds to an obstacle that enemies can pass, but Link requires a special raft item to pass.
%In Zelda, water is not passable by Link until he has obtained the raft item, and even then he can only pass over a single water tile. This item is implemented in the Rogue-like, and is in each level. Zelda also has some enemies that can pass over water, and others that cannot. In contrast, all Rogue-like enemies can pass over water.
Enemies are not represented in the VGLC data because its authors did not include them.\footnote{Pseudo-randomly placed enemies appear in visualizations as a red `e'}%\footnote{VGLC erroneously refers to statues that occupy some rooms as enemies} %, but they are simply impassable objects. Other enemies are absent from VGLC} % However, enemy data is not needed to run A* search to find paths through generated dungeons, which is the approach used for evaluation in this paper.
%VGLC does include information about doors linking rooms, but that information is excluded from the encoding because door placement is handled not by the GAN, but by CPPNs, as described in Section \ref{sec:cppn2gan}.

%\todo[inline]{Maybe mention soft-lock, locks/keys, and bombs, since they are part of the encoding below}

\section{Approach}
\label{sec:approach}

The novel approach introduced in this paper 
is a hybrid of CPPN2GAN and Direct2GAN, so each is explained in detail before explaining CPPNThenDirect2GAN.
All approaches depend on a GAN trained on data for the target game.

%%%% From previous paper %%%%%%
%employs a two-stage indirect encoding to evolve video game levels. Individual level segments are created by sending latent vectors to a Generative Adversarial Network (GAN) trained on data for a target game. To create whole levels, locations of individual segments are used as input to a Compositional Pattern Producing Network (CPPN), which outputs latent vectors for each segment. An overview of the complete CPPN2GAN approach below is shown in Fig.~\ref{fig:overview}. For comparison, the Direct2GAN control approach that generates levels from a genome consisting of many separate latent vectors is also described below. 

%%%%% Was not even present in previous paper %%%%%%%%%%%%
%\todo[inline]{This next paragraph will need to be rearranged (maybe even dropped?) after the upcoming sections are fully fleshed out}
%The next section first described how the game level are processed (Section~\ref{sec:processing}) before the training of the GAN is described in more detail (Section~\ref{sec:training}). The baseline approach, in which a separate latent vector is evolved for each level segment is described in Section~\ref{sec:direct2gan}, followed by the novel CPPN2GAN approach introduced in this paper in Section~\ref{sec:cppn2gan}. 

\subsection{GAN Training Details}
\label{sec:processing}

%The Mario model used was taken from a publicly available repository associated with previous research \cite{volz:gecco19}, %\todo{This is not technically true. The GECCO 19 proposal used the old encoding. The new models are only used in my thesis and the games benchmark},

%The Zelda model is new, but is based on other recent research \cite{gutierrez2020zeldagan}.

The Mario and Zelda models are the same as used in our previous work \cite{schrum:gecco2020cppn2gan}
but details of their training are repeated here.
Both models are Wasserstein GANs \cite{arjovsky2017wasserstein} differing
only in the size of their latent vector inputs (10 for Zelda, 30 for Mario), and the depth of the final output layer (3 for Zelda, 13 for Mario).
Their architecture otherwise matches that shown in Fig.~\ref{fig:overview}.
Output depth corresponds to the number of possible tiles for the game. The other output dimensions are $32 \times 32$, which is larger than the 2D region needed to render a level segment. The upper left corner of the output region is treated as a generated level, and the rest is ignored.

%Output dimensions can match for both games because regions
%outside the border of expected output dimensions are ignored.\todo{This is unclear. Also: The number of layers is not necessarily the same, correct? Also, my new GAN approach has a different number of layers. Could we not just use variables in the figure to avoid confusion?}

%Furthermore, the system described in this paper is compatible with all
%previously published GAN models associated with these papers.
%\todo{From Vanessa: Should we mention training details here? that we use wgan? Answer from Jacob: We can spare those details, but we can add them if there is still room. Depends on whether we need space for JSD}

To encode levels for training,
each tile type is %represented by a distinct integer, which is 
converted to a one-hot
encoded vector before being input into the
discriminator. 
The generator also outputs
levels \hl{in} %represented using 
the one-hot encoded format. %, which is then converted back to integers.
Mario or Zelda levels in this %integer-based 
format are sent to
the Mario AI framework or Rogue-like engine for rendering.
%(Tables~\ref{tab:mariotiles} and~\ref{tab:zeldatiles}).

%The mapping from VGLC tile types and symbols, to GAN training number codes, and finally to Mario AI/Rogue-like tile visualizations is detailed in Tables~\ref{tab:mariotiles} and~\ref{tab:zeldatiles}. 

GAN input files for Mario were created by processing all 12 \emph{overworld} level files
from VGLC for \emph{Super Mario Bros.} %Each level is 14 tiles high. 
The GAN expects to always see a rectangular
input of the same size, so each input was generated
by sliding a 28 (wide) $\times$ 14 (high)
window over the level from left to right, one tile at a time, where 28 tiles is the width of the screen in Mario. 

%%%% I think this following fact has been stated enough already
%In the input files each tile type is
%represented by a specific character, which was then mapped to a specific integer in the training images, as listed in Table~\ref{tab:mariotiles}.  

%While we could have used a larger dataset instead of this relatively small one, its use allows us to test the GAN's ability to learn from relatively little data, which could be especially important for games that do not offer such a large training corpus as Mario. Additionally, because of the smaller  training set it is possible to manually inspect if the LVE approach is able to generate levels with properties not directly  found in the training set itself. 

GAN input for Zelda was created from the 18 dungeons in VGLC for \emph{The Legend of Zelda}, but the training samples are the individual rooms in the dungeons, which are 16 (wide) $\times$ 11 (high) tiles. Many rooms are repeated within and across dungeons, so only unique rooms were included in the training set (only 38 samples). Training samples are simpler than the raw VGLC rooms because the various tile types are reduced to a set of just three as described in Section \ref{sec:zelda}. % in Table~\ref{tab:zeldatiles}. 
Doors were transformed into walls because door placement is not handled by the GAN, but rather by evolution, as described next.

\subsection{Level Generation: CPPN2GAN}
\label{sec:cppn2gan}

%CPPNs easily generate symmetrical and repeating patterns with and without variation. 

To generate a level using \hl{CPPN2GAN}, the CPPN is given responsibility for generating latent vector inputs for the GAN as a function of segment position within the larger level.

For Mario, the only input is the x-coordinate of the segment, scaled to $[-1,1]$. For a level of three segments, the CPPN inputs would be $-1$, $0$, and $1$. CPPN output is an entire latent vector. %(30 variables for Mario).
Each latent vector is fed to the GAN to generate the segment at that position in the level.

Zelda's 2D %arrangement of rooms is 
\hl{dungeons are} more complicated. %, as its dungeons is a 2D arrangement of rooms. %Interesting Zelda dungeons need to have some high-level shape. In fact, dungeons in the original game are known by the shape they represent, with names such as \emph{Eagle} and \emph{Snake}. 
For the overall %dungeon 
shape to be interesting, some rooms %in the 2D grid 
need to be missing. 
Also, dungeons are typically more interesting if they are maze-like, so simply connecting all adjacent rooms would be boring. How maze-like a dungeon is also depends on its start and end points.
%The GAN produces a single room for each latent vector
These additional issues are global design issues, and \hl{so} %therefore 
are handled by the CPPN (Fig.~\ref{fig:overview}), which defines global patterns, rather than the GAN, which generates individual rooms.

Thus, CPPNs for Zelda generate latent vector inputs and additional values that determine the layout and connectivity of the rooms.
Zelda CPPNs take inputs of x and y coordinates scaled to $[-1,1]$. A radial distance input is also included to encourage radial patterns, which is common in CPPNs \cite{secretan:ecj2011}. For each set of CPPN inputs, the output is a latent vector % (size 10 for Zelda) 
along with seven additional numbers: room presence, right door presence, down door presence, right door type, down door type, raft preference, and start/end preference.

Room presence determines the presence/absence of a room based on whether the number is positive. %Similarly, 
\hl{If} a room is present and has a neighboring room in the given direction, 
then positive right/down door presence values place a door in the wall heading right/down. % based on whether the corresponding value is positive. 
Whenever a door is placed, a door is also placed in the opposite direction within the connecting room %, which is why 
\hl{(so top/left door outputs are not needed)}. For variety, the right/down door type determines the types of doors, based on different number ranges for each door type: $[-1,0]$ for plain, $(0,0.25]$ for puzzle-locked, $(0.25,0.5]$ for soft-locked, $(0.5,0.75]$ for bomb-able passage, and $(0.75,1.0]$ for locked. 
Puzzle-locked doors are a new addition to this paper which were absent in the initial paper on CPPN2GAN \cite{schrum:gecco2020cppn2gan}. A puzzle-locked door \hl{is} %can only be 
opened by pushing a special block in the room in a certain direction. The remaining door types were present in previous work \cite{schrum:gecco2020cppn2gan}:
Soft-locked doors only open when all enemies in the room are killed, bomb-able passages are secret walls that can be bombed to create a door, and locked doors need a key. Enough keys to pair with all locked doors are placed at random locations in %rooms of 
the dungeon. However, to assure that the genotype-to-phenotype mapping is deterministic, the pseudo-random generator responsible for placing keys is initialized using the bit representation of the corresponding right or down door type output as a seed. %When present, 
\hl{Puzzle} blocks are placed pseudo-randomly in the same manner.

Raft preference is another addition absent in the previous CPPN2GAN paper \cite{schrum:gecco2020cppn2gan}, though rafts were included in the Graph GAN paper \cite{gutierrez2020zeldagan} that introduced this Rogue-like Zelda domain. \hl{The raft item allows Link} to traverse a single water tile, \hl{but} % . This item 
is only available in one room in each dungeon. %, and must be retrieved before it can be used. 
The %generated 
room with the highest raft preference value is the one \hl{where} %in which 
the raft is pseudo-randomly placed. The %addition of the 
raft %item 
can greatly increase the amount of back-tracking required in some dungeons.

%A raft is a special item that allows Link to traverse a single water tile.

The final output for start/end preference determines which rooms are the start/end rooms of the dungeon. Across all rooms, % in the dungeon,
the one whose start/end preference is smallest is the %player's 
starting room, and the one with the largest output is the final goal room, designated by the presence of a Triforce item. % (triangle). %% None of the figures clearly display the triangle

This approach to generating complete levels is compared with the control approach described next.

\subsection{Level Generation: Direct2GAN}
\label{sec:direct2gan}

%To have a meaningful comparison with the CPPN2GAN approach, an approach that directly evolves genomes consisting of multiple latent vectors is needed. 

Direct2GAN evolves levels consisting of $S$ segments for a GAN expecting latent inputs of size $Z$ by evolving real-valued genome vectors of length $S \times Z$. Each genome is chopped into individual GAN inputs at level generation.

This approach requires a convention as to how different segments are combined into one level. For Mario's linear levels, %it is straightforward to make 
adjacent GAN inputs from the combined vector correspond to adjacent segments in the generated level. The combined vector is processed left to right to produce segments left to right.

To generate 2D Zelda dungeons, individual segments of the linear genome are 
mapped to a 2D grid in row-major order: processing genome from left to right generates top row from left to right, then moves to next row down and so on. For fair comparison with CPPN2GAN, each portion of a genome corresponding to a single room contains not only the latent vector inputs, but the seven additional numbers for controlling global structure and connectivity: room presence, right door presence, down door presence, right door type, down door type, raft preference, and start/end preference. Therefore, a $M \times N$ room grid requires genomes of length $M \times N \times (Z + 7)$. % from Direct2GAN. 

Such massive genomes induce large search spaces that are difficult to search, but they have the benefit of easily allowing arbitrary variation in any area of the genome. Therefore, CPPNThenDirect2GAN was developed to take advantage of the benefits of both CPPN2GAN and Direct2GAN.

\subsection{Level Generation: CPPNThenDirect2GAN}

This hybrid approach is inspired by the HybrID algorithm \cite{CluneBPO09:HybrID}. HybrID was used with HyperNEAT \cite{stanley:alife2009}, an indirect encoding that evolves CPPNs, which in turn define the weights of a pre-defined neural network architecture. The benefit of evolving with CPPNs is that they easily impose global patterns of symmetry and repetition. However, localized variation is harder for a CPPN to represent. In contrast, localized variation is easy to produce given a directly encoded genome that simply represents each network weight individually. HybrID begins evolution using CPPNs, so that useful global patterns can be easily found. However, at some point during evolution the CPPNs are discarded, leaving only the directly encoded collection of weights they produced to evolve further.

CPPNThenDirect2GAN works in a similar fashion. At initialization, all individuals in the population are CPPNs evolved \hl{with} CPPN2GAN. An additional mutation operator is introduced that switches a \hl{CPPN} to the Direct2GAN representation, which is the output of the CPPN queried at all coordinates (as described above). This individual is then further evolved \hl{by} the Direct2GAN approach. This operation has a low probability, since genomes can never switch back to an indirectly encoded CPPN once they switch to a direct vector format.

In HybrID \cite{CluneBPO09:HybrID}, the transition from indirect to direct encoding occurred at a particular generation. In contrast, our hybrid approach \hl{transforms offspring based on random chance and thus allows for mixed populations. The now directly encoded offspring undergoes mutation after the transition in order to differentiate the generated level from that of its CPPN parent. The mixed population assures that CPPN genotypes persist as long as they are useful. Though it is possible for directly encoded genotypes to completely take over the population, this did not occur in any of the experiments described next.}

%, which makes more sense when using MAP-Elites to evolve a diverse population.

%CPPNs are evolved as with CPPN2GAN. Whenever a CPPN needs to generate a level, it is first queried repeatedly to generate several latent vector outputs which also serve as inputs to a pre-trained GAN. However, the collection of latent vectors that the CPPN produces is equivalent to the Direct2GAN representation, and can be evolved if the CPPN is discarded.

%In HybrID \cite{CluneBPO09:HybrID}, the transition from indirect to direct encoding occurred at a particular generation. However, CPPNThenDirect2GAN does not switch the entire population at once. Instead, the transition from CPPN to Direct encoding is caused by a mutation operator. This operation has a low probability, since genomes can never switch back to an indirectly encoded CPPN once they switch to a direct vector format. 

\section{Experiments}

%Demonstrating the expressive range of new game level encodings is important.

\hl{The experiments below demonstrate the expressive range of these game level encodings using %Therefore, 
the} Quality Diversity algorithm MAP-Elites \cite{mouret:arxiv15}, which divides the search space into phenotypically distinct bins. %, is used with each approach.

\subsection{MAP-Elites}

Instead of only optimizing towards an objective, as in standard evolutionary algorithms, MAP-Elites (Multi-dimensional Archive of Phenotypic Elites \cite{mouret:arxiv15})  collects a diversity of quality artefacts that differ along $N$ predefined dimensions. MAP-Elites discretizes the space of artefacts into bins and, given some objective, maintains the highest performing individual for each bin in the $N$-dimensional behavior space. % (the elites). 

%Our implementation starts by generating

\hl{First an initial 100 random individuals are placed in bins based on their attributes.} Each bin only holds one individual, so individuals with higher fitness replace less fit individuals. Once the initial population is generated, solutions are uniformly sampled from the bins and undergo crossover and/or mutation to generate new individuals. These newly created individuals also replace less fit individuals as appropriate, or end up occupying new bins, \hl{so that a variety of niches is represented, but only by the best examples discovered so far}.
Our experiments generate 100,000 individuals per run after the initial population is generated.

% Performance is measured both in terms of achieved fitness and the number of bins that are filled as more individuals are generated. 

%It is important to define the binning scheme in a way that supports a range of meaningful variation within the target domain, so each game uses its own distinct binning scheme. Specific fitness measures are also needed for each domain.

%%%%NEW%%%%%
%The dimensions of potential variation differ in each game, and even for a specific game 

There are many ways of defining a binning scheme to characterize diversity in a domain. The initial study introducing CPPN2GAN \cite{schrum:gecco2020cppn2gan} used one binning scheme with each game, but because the performance of a given encoding depends on the binning scheme, this paper uses two different binning schemes for each game to better explore the trade-offs between the encodings being studied. A fitness measure is also required \hl{(next section)}, \hl{and} is consistent in each of the games studied.

%To support a range of meaningful variation within each domain, each game uses its own distinct binning scheme and fitness measure.

\subsection{Dimensions of Variation Within Levels}

%\todo[inline]{We need a standard naming convention for the many binning scheme throughout the paper, and also a way of abbreviating so that it takes up less space to refer to them}

%though the performance metric used in each is based on the length of the A* solution path through the produced level. 

Two binning schemes were used on Mario levels. One from previous work~\cite{schrum:gecco2020cppn2gan} and one new \hl{for} this article.
Both are based on measurements of three quantities: decoration frequency, space coverage, and leniency. These measures were inspired by a study on evaluation measures for platformer games \cite{Summerville2017}. Each measure expresses different characteristics of a level:
\begin{itemize}
    \item decoration frequency: Percentage of \emph{non-standard} tiles\footnote{breakable tiles, question blocks, pipes, all enemies}
    \item space coverage: Percentage of tiles 
    Mario can stand on\footnote{solid and breakable tiles, question blocks, pipes and bullet bills}
    \item leniency: Average of leniency values\footnote{1: question blocks; -0.5: pipes, bullet bills, gaps in ground; -1: moving enemies; 0: remaining} across all tiles. 
    Enemies/gaps are negative, power-ups are positive.
\end{itemize}
All measures focus on visual characteristics, but also relate to how a player can navigate through a level. The previous binning scheme \cite{schrum:gecco2020cppn2gan} calculated scores for individual segments (10 per level) and then summed across the segments.

%allowing each dimension to be discretized into 10 equally sized intervals.

Preliminary experiments uncovered reasonable ranges for binning, \hl{discretizing} each dimension into 10 equally sized intervals.
Leniency has negative and positive values, so its scores are divided into 5 negative bins and 5 non-negative bins. Negative bins correspond to greater challenges, and non-negative bins correspond to easier levels. This binning scheme is %referred to as 
\textbf{Sum DSL} \hl{({\bf D}ecoration, {\bf S}pace coverage, {\bf L}eniency)}.

However, one way to hit a target bin in \textbf{Sum DSL} is to repeat a segment with appropriate properties without variation 10 times. This type of level might be boring to play and CPPNs have an advantage at generating this type of level, so an alternate binning scheme encouraging more segment diversity within levels was developed: \textbf{Distinct ASAD}.

\textbf{Distinct ASAD} uses \hl{{\bf A}}lternating \hl{{\bf S}}pace coverage, \hl{{\bf A}}lter-nating \hl{{\bf D}}ecoration frequency, and the number of distinct segments, which is a count of segments that skips repeats. Distinctness directly encourages variation, though two segments are considered distinct for just a single different tile. The alternating dimensions measure how much their quantities fluctuate from segment to segment. Specifically, if $S(i)$ calculates a given score value (e.g.\ space coverage) for the $i$-th segment in the level, then the alternating version of that score is:

\begin{equation}
A_{S} = \sum_{i=1}^9 |S(i - 1) - S(i)|
\end{equation}

Using this formula to distinguish levels from each other encourages more variation in the segments within each level. Both alternating scores are discretized into 10 intervals.

For both binning schemes, fitness is the length of the shortest path to beat the level. Maximizing path length favors levels that require jumps, the main mechanic of the game. If no path can be found the level is deemed unsolvable and receives a fitness of 0. To determine the path, A* search is performed on the tile-based representation of the level, with a heuristic encouraging heading to the right.

%%%% REMOVED THIS WHOLE PARAGRAPH FOR SPACE: 2021 %%%%%
%To determine the path, A* search is performed on the tile-based representation of the level. To limit computational costs, the game physics are simplified as follows. First, Mario always occupies a single tile (small state), and can either move left, move right, or jump. Once a jump starts, Mario moves up one tile per action for the next four actions (unless there is an obstruction), and then falls down until landing on an impassible tile. Mario can still make left/right movements while airborne, but can only initiate a jump if on solid ground. Enemies are just impassible obstacles, and Mario dies if he falls in a pit. The heuristic used favors states further to the right, because the right edge of the screen is the goal.

Zelda also uses a binning scheme from previous work~\cite{schrum:gecco2020cppn2gan} and a new one. The old scheme, \textbf{WWR}, is based on \hl{{\bf W}}ater tile percentage, \hl{{\bf W}}all tile percentage, and the number of \hl{{\bf R}}eachable rooms. A room is reachable if it is the start room, or a door connects it to a reachable room. This definition is cheap to compute, but ignores how single rooms can be impassable. Water and wall tile percentages are calculated only with respect to reachable rooms, and only for the $12 \times 7$ floor regions of rooms (surrounding walls ignored). Bins for these dimensions are divided into 10\% ranges (10 bins per dimension). Some bins are impossible to fill, because the sum of water and wall percentages must be less than 100\%. Floor tiles occupy additional space. For the number of reachable rooms, there is a bin for each number out of 25 (maximum possible number in a $5 \times 5$ grid). Note that in the previous paper \cite{schrum:gecco2020cppn2gan}, experiments were conducted with 100-room dungeons in a $10 \times 10$ grid. The experiments in this paper were repeated due to changes in the Zelda level-generation approach mentioned in Section~\ref{sec:cppn2gan}, but were done with smaller dungeons to reduce computational cost. However, all results presented here are consistent with those from the previous paper.

The new binning scheme for Zelda is \textbf{Distinct BTR}. It encourages more variety in the rooms within each dungeon, and encourages different types of paths through dungeons. The specific bin dimensions are the number of \hl{{\bf D}}istinct rooms, the number of \hl{{\bf B}}ack-\hl{{\bf T}}racked rooms in the A* solution path, and the number of \hl{{\bf R}}eachable rooms (as before). The backtracking dimension deserves some elaboration.

%For example, the player may need to find a key before backtracking to a room with a locked door, or find a special item (like the raft) before traversing a particular obstacle.

In commercial games, some dungeons can be traversed in a single pass, but others require the player to return to previously visited areas multiple times, \hl{e.g.~to} find a key before backtracking to a room with a locked door, or \hl{to} find a special item (like the raft) before traversing a particular obstacle.
Some players enjoy this type of backtracking, while others find it frustrating, which makes it a good dimension of variation. %: dungeons requiring all different levels of backtracking can be discovered with MAP-Elites.

% Such backtracking generally rewards players for paying attention to their environment and being able to navigate back to previously observed landmarks.

\begin{figure*}[t]
\centering
\begin{subfigure}{0.24\textwidth} 
    \includegraphics[width=1.0\textwidth]{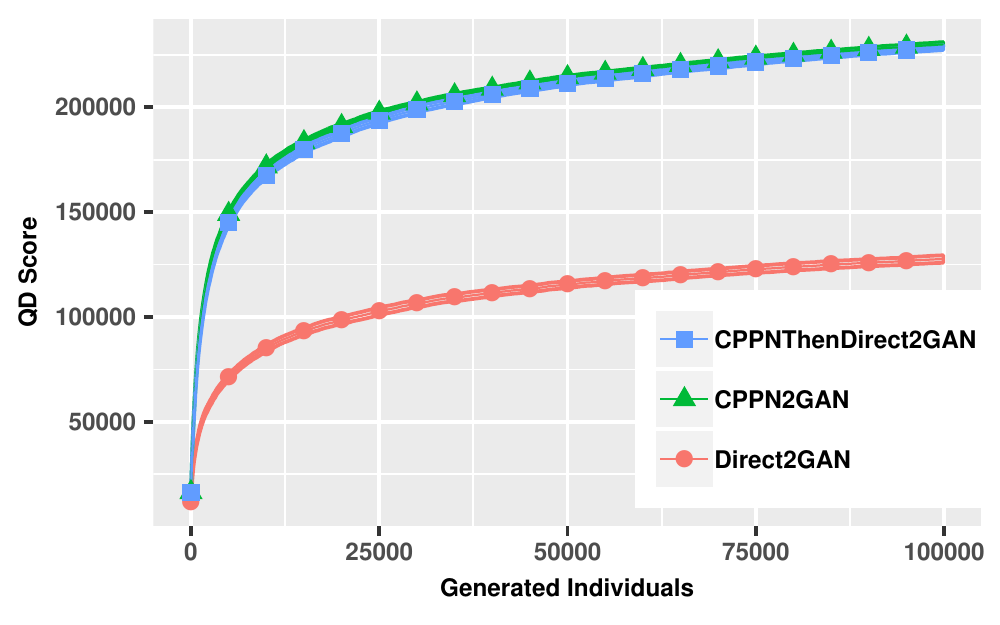}
    \caption{Mario: \textbf{Sum DSL}}
    \label{fig:marioDNSqd}
\end{subfigure}
\begin{subfigure}{0.24\textwidth} 
    \includegraphics[width=1.0\textwidth]{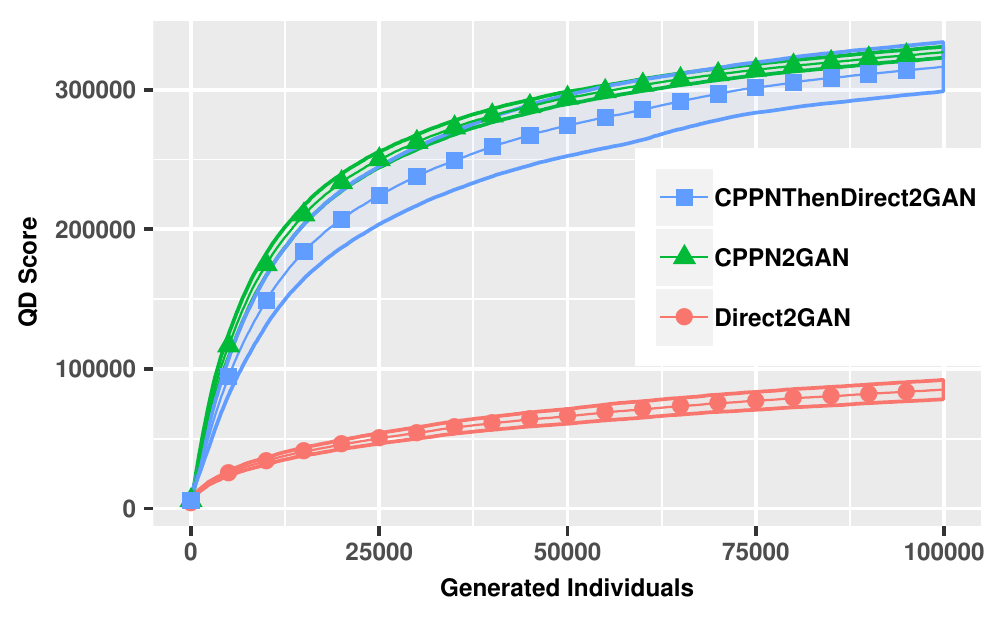}
    \caption{Mario: \textbf{Distinct ASAD}}
    \label{fig:marioDNDqd}
\end{subfigure} 
\begin{subfigure}{0.24\textwidth} 
    \includegraphics[width=1.0\textwidth]{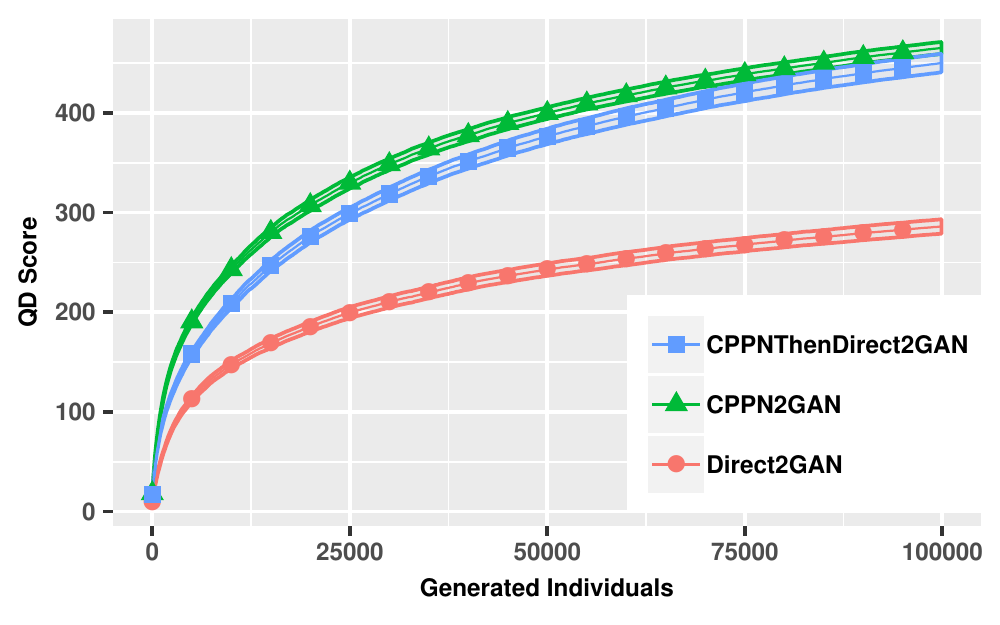}
    \caption{Zelda: \textbf{WWR}}
    \label{fig:zeldaWWRqd}
\end{subfigure}
\begin{subfigure}{0.24\textwidth} 
    \includegraphics[width=1.0\textwidth]{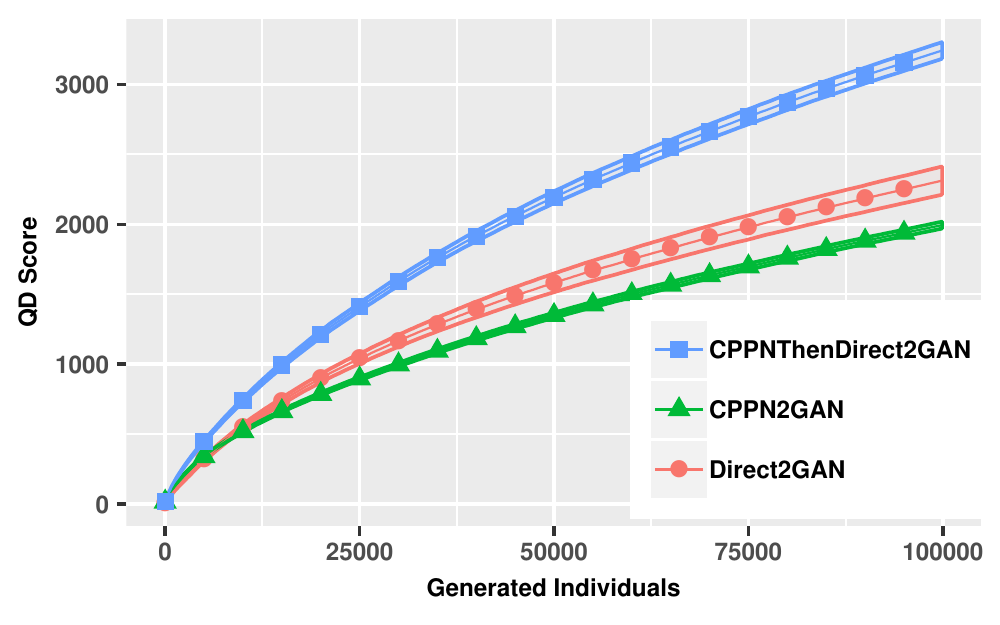}
    \caption{Zelda: \textbf{Distinct BTR}}
    \label{fig:zeldaDBRqd}
\end{subfigure}

\caption{\textbf{Average QD Score Across 30 Runs of MAP-Elites.} \normalfont For each MAP-Elites binning scheme in Mario and Zelda, plots of average QD Scores with 95\% confidence intervals demonstrate the comparative performance of the three encoding schemes. CPPNThenDirect2GAN is always the best or statistically tied for best (eventually catching up to CPPN2GAN in \textbf{WWR}). Direct2GAN is always worst, except with \textbf{Distinct BTR}, where it beats CPPN2GAN, but is inferior to CPPNThenDirect2GAN.}
\label{fig:averageQD}
\end{figure*}

The backtracking score is calculated by following the A* path and marking each room the player exits. Exiting a room means another room is being entered. Whenever a newly entered room exists in the set of previously exited rooms, a counter is incremented to measure the amount of backtracking required by the optimal path. If the path revisits a room multiple times, each revisit increments the backtracking count.

The fitness for Zelda dungeons is the percentage of reachable rooms traversed by the A* path from start to goal. The objective is to maximize the number of rooms visited, as exploring is one of the main mechanics of the game. If no path can be found, the dungeon is deemed unsolvable and receives 0 fitness. The A* heuristic used is Manhattan distance to the goal. Since the inclusion of keys makes the state space very large, there is a computation budget of 100,000 states.

%%%% REMOVED 2021 %%%%%%%%%%%%%%
%A* is again used to determine the path, now using Manhattan distance to the goal as a heuristic. We take locked doors and keys into account to only generate paths with  rooms that are \emph{actually} reachable. Since the inclusion of keys makes the state space very large, there is a computation budget of 100,000 states. 

%\todo{transition}

\subsection{Evolution Details}

CPPN2GAN levels are evolved with a variant of NEAT \cite{stanley:ec02} (Section \ref{sec:cppnsAndNEAT}), specifically
MM-NEAT.\footnote{\url{https://github.com/schrum2/MM-NEAT}}
Because CPPNs are being evolved, every neuron in each network can have a different activation function from the following list: 
sawtooth wave, 
linear piecewise, 
id,
square wave, 
cosine,
sine, 
sigmoid,
Gaussian, 
triangle wave, and 
absolute value. 

Whenever a new network is generated, is has a 50\% chance of being the offspring of two parents rather than a clone. The resulting network then has a 20\% chance of having a new node spliced in, 40\% chance of creating a new link, and a 30\% chance of randomly replacing one neuron's activation function. There is a per-link perturbation rate of 5\%.

For Direct2GAN, real-valued vectors are initialized with random values in the range $[-1,1]$. When offspring are produced, there is a 50\% chance of single-point crossover. Otherwise, the offspring is a clone of one parent. Either way, each real number in the vector then has an independent 30\% chance of polynomial mutation \cite{deb1:cs95:polynomial}.

When using CPPNThenDirect2GAN, all genomes start as CPPNs. When bins are randomly sampled to generate offspring, a CPPN or directly encoded vector could be selected. CPPNs have a 30\% chance of being converted into directly encoded vectors. This procedure generates a directly-encoded genome that represents the exact same level previously encoded by the CPPN. However, the newly generated vector genome immediately undergoes the mutation for real-valued vectors described above, so it will only persist in the archive if the resulting level is an elite. Genomes that are not converted are exposed to the standard mutation probabilities for their encoding as described above. Whenever crossover occurs, parents mate in the usual fashion if they are of the same type, but if two parents of different types are selected, then the crossover operation is cancelled and the first parent is simply mutated to create a new offspring.

\section{Results}

This section highlights our most relevant results, but an online appendix contains additional result figures and sample evolved levels: \url{https://southwestern.edu/~schrum2/SCOPE/cppn-then-direct-to-gan.php}

Fig.~\ref{fig:averageQD} shows the average QD score across 30 runs of each genome encoding for each binning scheme in the two games. QD score \cite{pugh:gecco2015qd} is the sum of the fitness scores of all elites in the archive, and gives an indication of both the coverage and quality of solutions.
Fig.~A.1 (appendix) %\ref{fig:averageBins}
shows the average number of bins filled by each encoding, and is qualitatively similar. In Mario, CPPN2GAN and CPPNThenDirect2GAN are statistically tied in terms of filled bins and QD score, but are significantly better than Direct2GAN ($p < 0.05$) for both binning schemes. Performance in Zelda depends more on the binning scheme. In the \textbf{WWR} scheme from previous work, CPPN2GAN is slightly better than CPPNThenDirect2GAN in terms of filled bins and QD score ($p < 0.05$), though CPPNThenDirect2GAN almost catches up in terms of QD score. Both are far better than Direct2GAN in both metrics ($p < 0.05$). For the new \textbf{Distinct BTR} scheme, CPPNThenDirect2GAN is significantly better than both CPPN2GAN and Direct2GAN in terms of filled bins and QD score ($p < 0.05$). CPPN2GAN and Direct2GAN are statistically tied in terms of filled bins, but Direct2GAN actually has the better QD score ($p < 0.05$).

\begin{figure*}[t]
\centering
\begin{subfigure}{0.48\textwidth} 
    \includegraphics[width=1.0\textwidth]{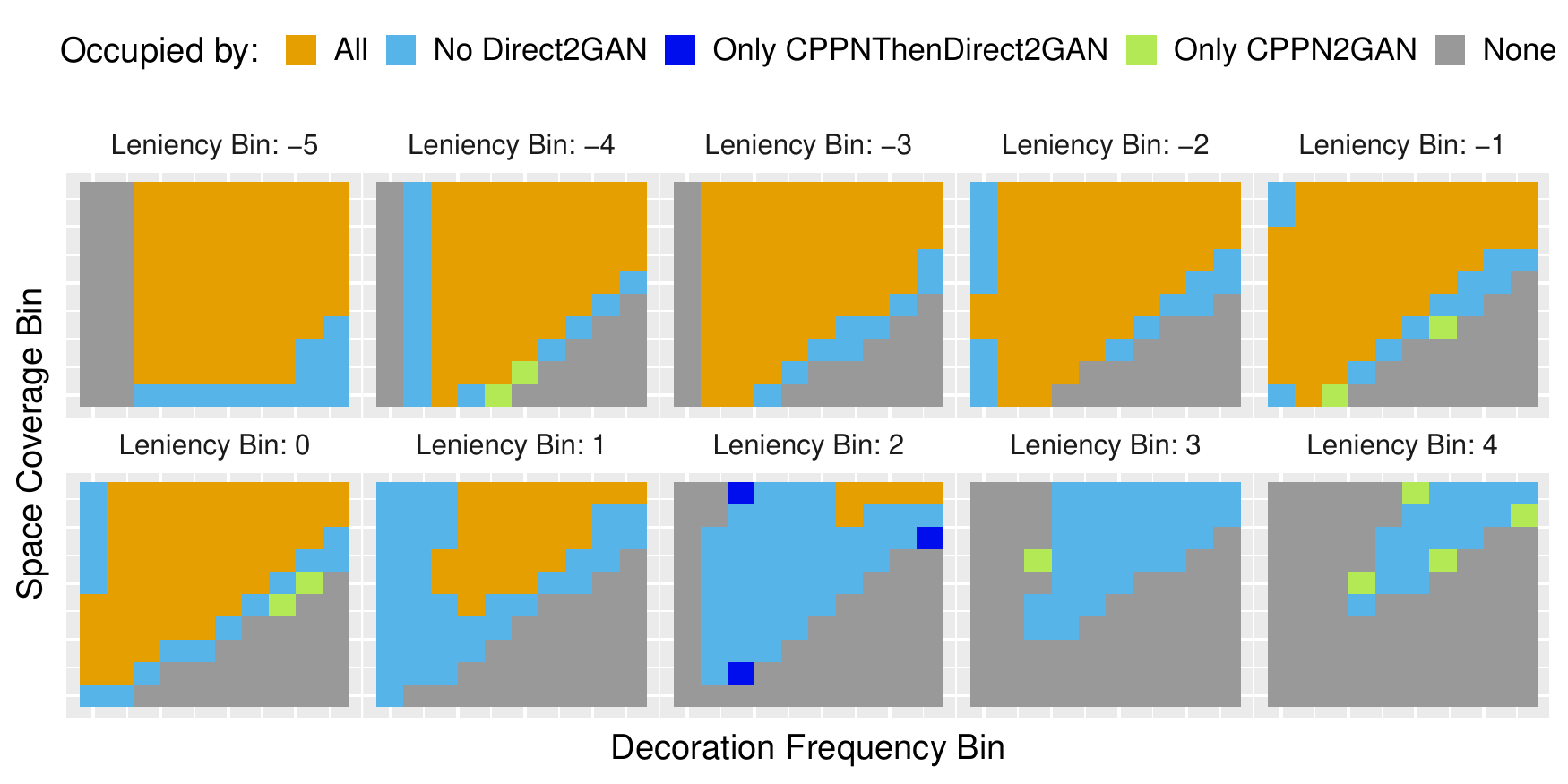}
    \caption{Difference in Bin Occupancy}
    \label{fig:marioDNSdiff}
\end{subfigure}
\begin{subfigure}{0.48\textwidth} 
    \includegraphics[width=1.0\textwidth]{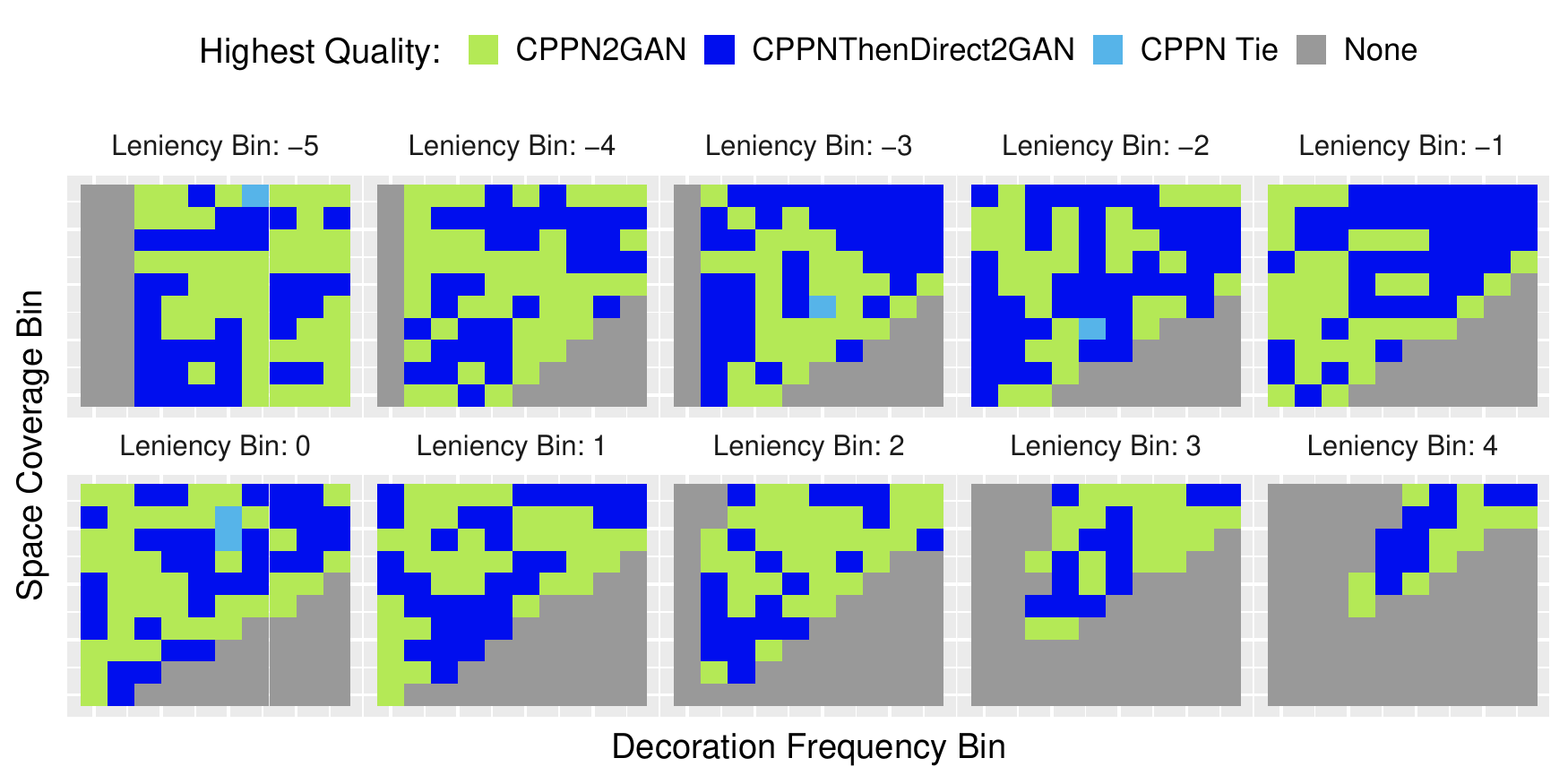}
    \caption{Best Occupant}
    \label{fig:marioDNSBest}
\end{subfigure}
\caption{\textbf{MAP-Elites Archive Comparisons Across 30 Runs of Evolution in Mario Using \textbf{Sum DSL}.} Each sub-grid represents a different leniency score. Within each sub-grid, summed decoration frequency increases to the right, and summed space coverage increases moving up. (\subref{fig:marioDNSdiff}) Color coding shows whether any of the 30 runs of each method produced an occupant for each bin. Direct2GAN leaves many bins completely absent, but CPPN2GAN and CPPNThenDirect2GAN have similar coverage. (\subref{fig:marioDNSBest}) The average bin \hl{fitness} scores across 30 runs of each method were calculated, and the method with the best average \hl{fitness} is indicated for each bin. Direct2GAN was never the best, though CPPN2GAN and CPPNThenDirect2GAN have a comparable number and spread of best bins, and tie for best in some cases (indicated by ``CPPN Tie'').}
\label{fig:marioSumDSL}
\end{figure*}

\begin{figure*}[t]
\centering
\begin{subfigure}{0.48\textwidth} 
    \includegraphics[width=1.0\textwidth]{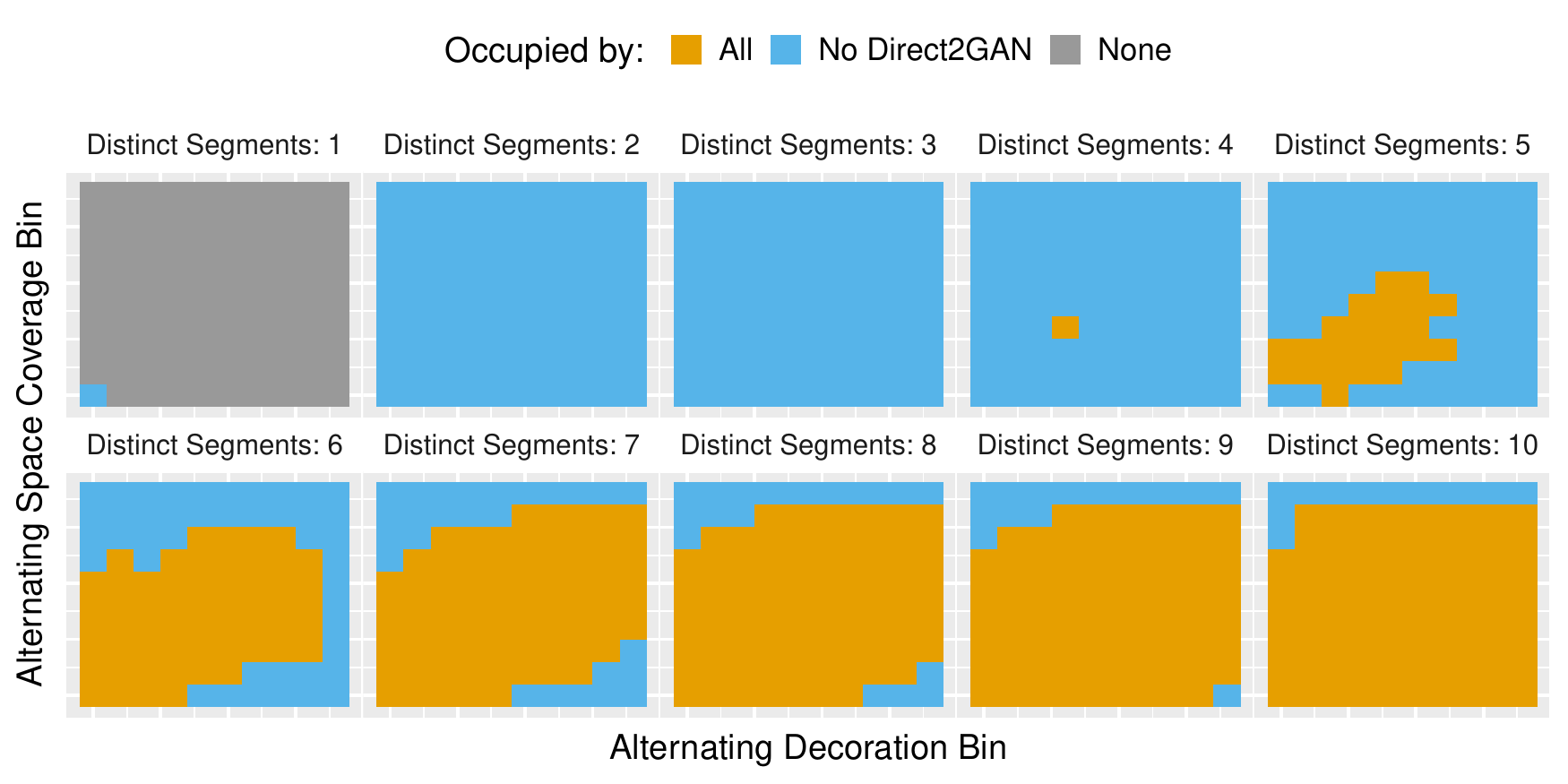}
    \caption{Difference in Bin Occupancy}
    \label{fig:marioDNDdiff}
\end{subfigure} 
\begin{subfigure}{0.48\textwidth} 
    \includegraphics[width=1.0\textwidth]{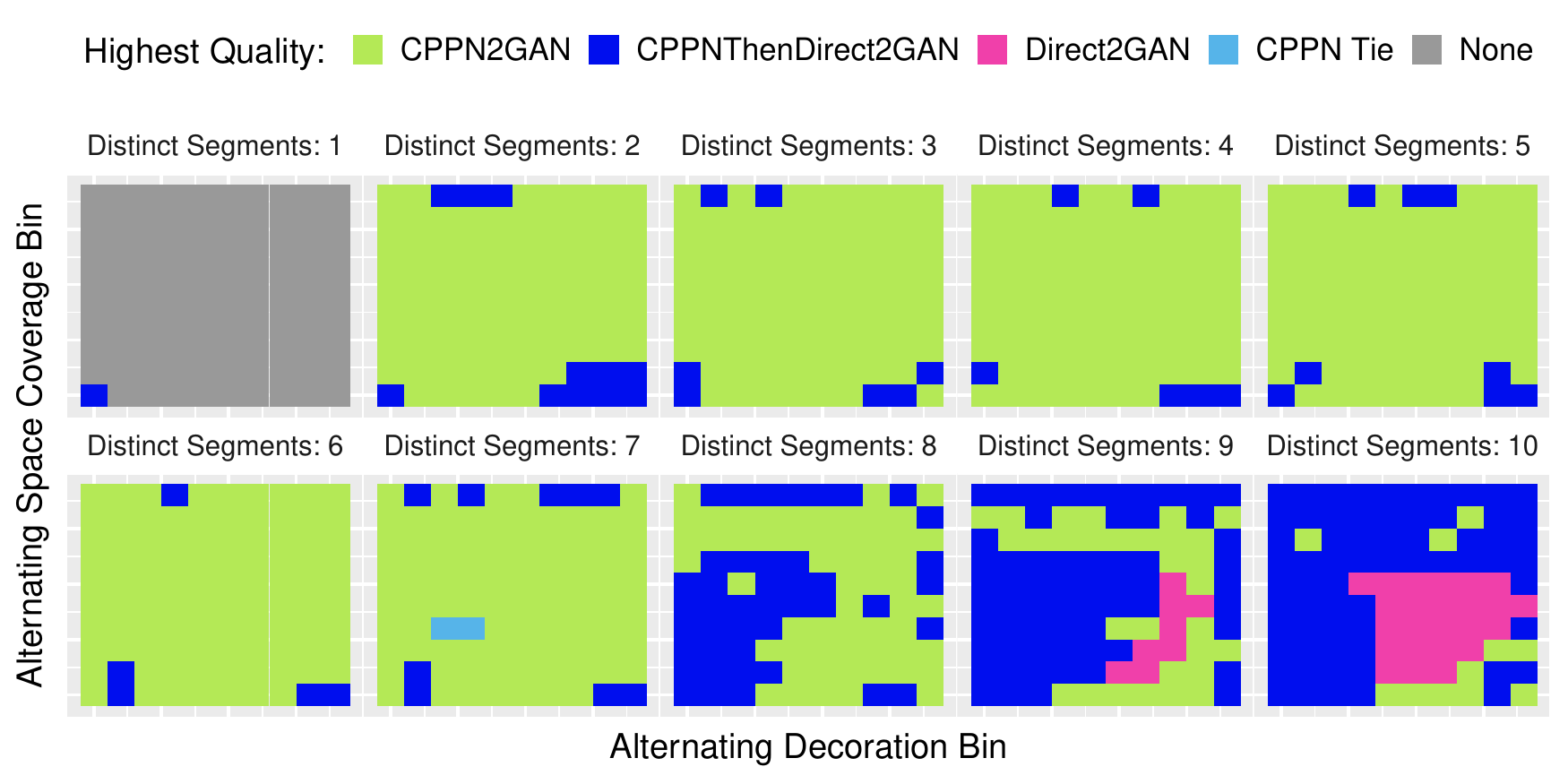}
    \caption{Best Occupant}
    \label{fig:marioDNDBest}
\end{subfigure} 

\caption{\textbf{MAP-Elites Archive Comparisons Across 30 Runs of Evolution in Mario Using \textbf{Distinct ASAD}.} Methods are compared as in Fig.~\ref{fig:marioSumDSL}, but with \textbf{Distinct ASAD}. Each sub-grid now represents the count of distinct segments. In each sub-grid, alternating decoration score increases to the right, and alternating space coverage score increases moving up. The upper-left grid is mostly empty, since it \hl{is for} %corresponds to 
levels with only one repeated segment, %. When the same segment is repeating, 
\hl{meaning} decoration and space coverage scores cannot alternate.
(\subref{fig:marioDNDdiff}) Direct2GAN is missing from many bins, but CPPN2GAN and CPPNThenDirect2GAN have identical coverage.
(\subref{fig:marioDNDBest}) The highest scoring levels for fewer distinct
segments mostly come from CPPN2GAN, though some CPPNThenDirect2GAN results are mixed in.
As number of distinct segments increases, CPPNThenDirect2GAN and Direct2GAN \hl{are} more prominent.}
\label{fig:marioDistinctASAD}
% and for 9 or 10 distinct segments Direct2GAN sometimes has the highest scores.
\end{figure*}

Fig.~\ref{fig:marioSumDSL} analyzes the final archives for \textbf{Sum DSL}. These results reaffirm the dominance of CPPN2GAN over Direct2GAN \cite{schrum:gecco2020cppn2gan}, and demonstrate the qualitative similarity between CPPN2GAN and the new CPPNThenDirect2GAN. For \hl{\textbf{Sum DSL}}, CPPNs are generally superior, but the new %hybrid 
CPPNThenDirect2GAN approach is not hindered by %also 
producing directly encoded genomes. Average archive heat maps for each method are in Fig.~A.2 (appendix). % \ref{fig:marioDNSaverageHeat} 
\hl{For each approach, the percentage of beatable levels averages to 97\%, though the actual number of beatable levels is much higher for CPPN-based approaches because of the higher number of filled bins.}

Fig.~\ref{fig:marioDistinctASAD} shows \textbf{Distinct ASAD} results. Direct2GAN performs even worse in terms of coverage, \hl{but} there is no difference %in bin occupancy 
between CPPN2GAN and CPPNThenDirect2GAN. However, CPPNThenDirect2GAN and %plain 
Direct2GAN more frequently produce the \hl{most fit} level as the number of distinct segments increases. Average archive heat maps are in Fig.~A.3 (appendix). % \ref{fig:marioDNDaverageHeat} 
\hl{Direct2GAN averages 97\% beatable levels, whereas both CPPN-based approaches average 99\% beatable levels.}

% but there are bins with many reachable rooms where CPPN2GAN is represented and CPPNThenDirect2GAN is not.

% Furthermore, in some bins with few reachable rooms, CPPNThenDirect2GAN is not represented. 

\begin{figure*}[t!]
\centering
\begin{subfigure}{0.48\textwidth} 
    \includegraphics[width=1.0\textwidth]{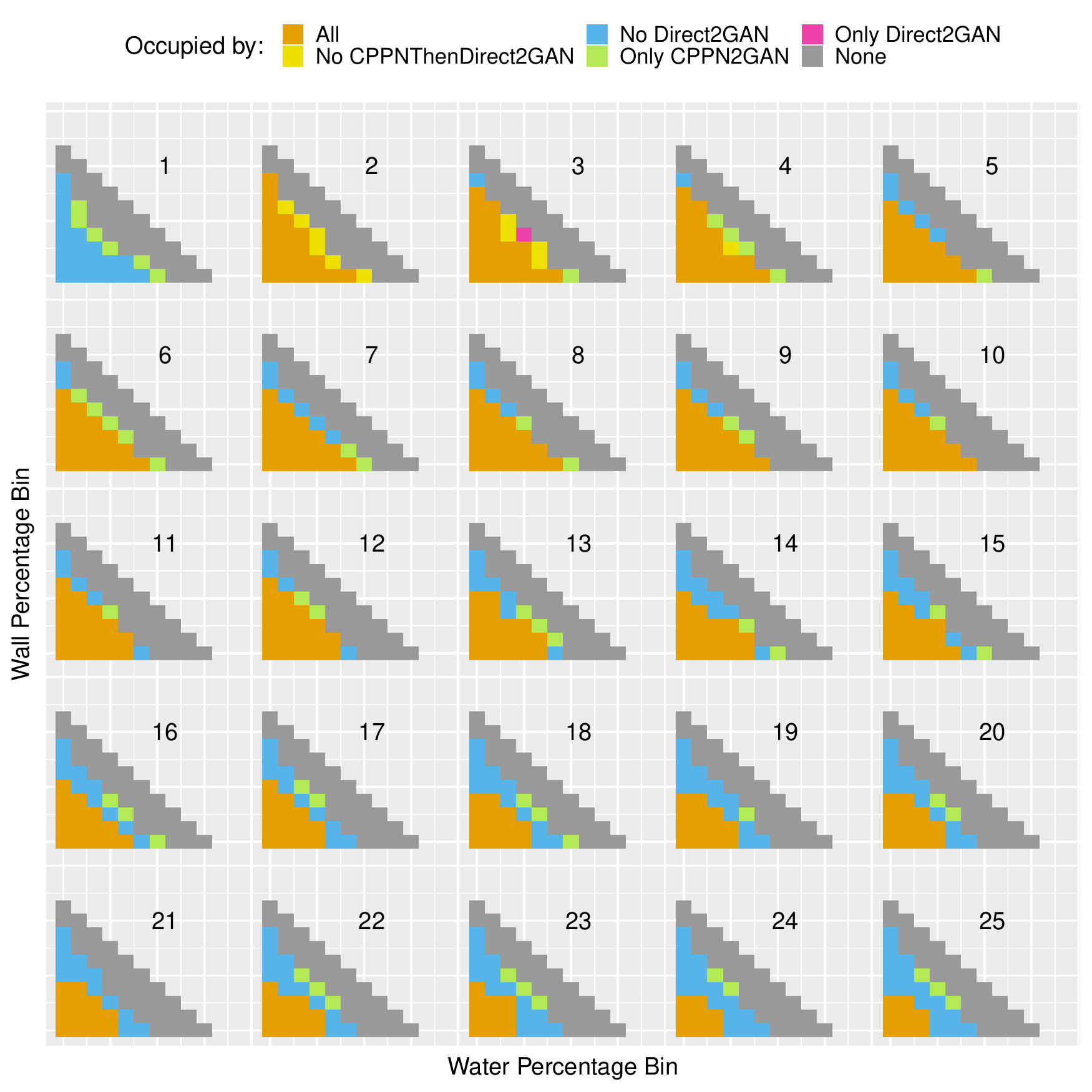}
    \caption{Difference in Bin Occupancy}
    \label{fig:zeldaWWRdiff}
\end{subfigure}
\begin{subfigure}{0.48\textwidth} 
    \includegraphics[width=1.0\textwidth]{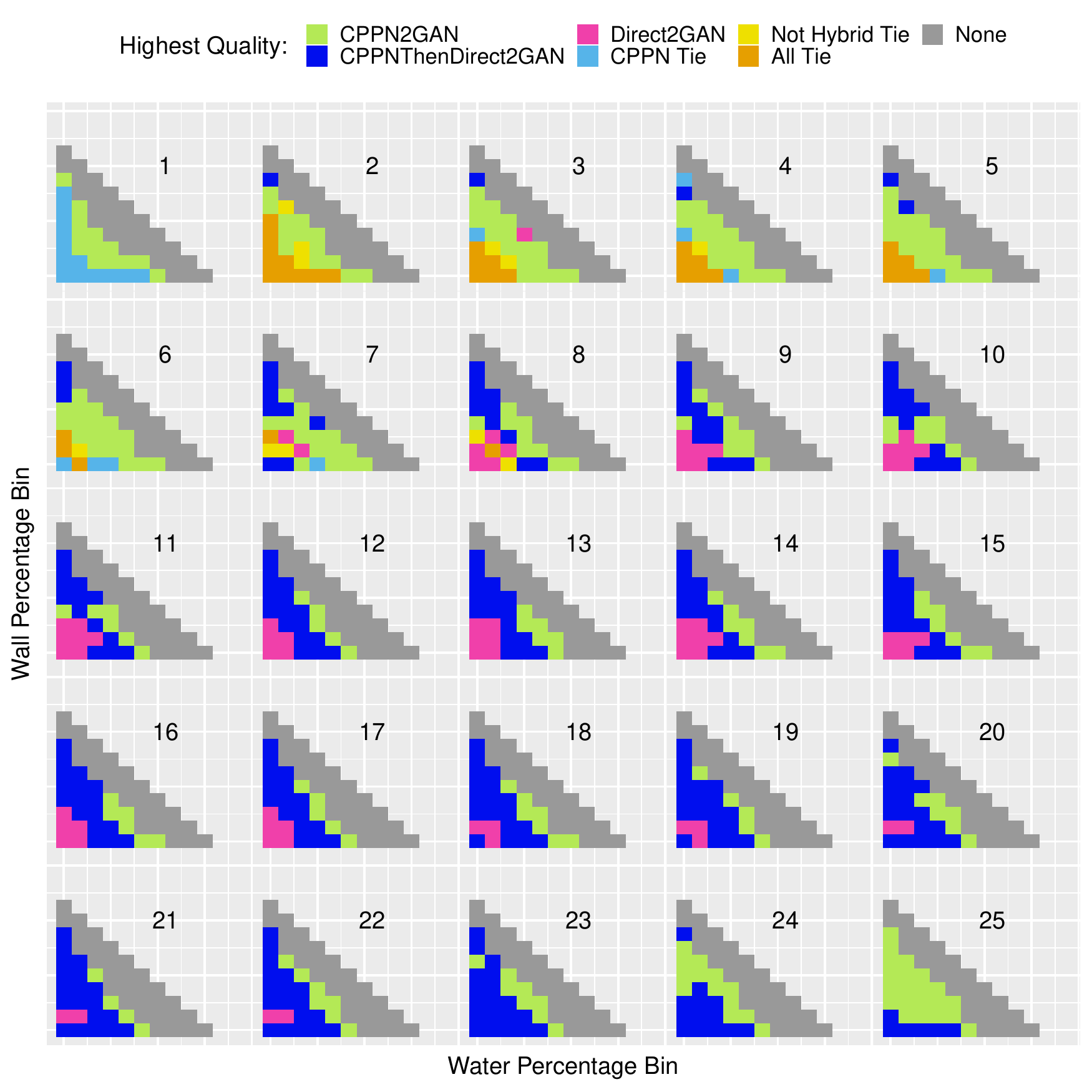}
    \caption{Best Occupant}
    \label{fig:zeldaWWRBest}
\end{subfigure}

\caption{\textbf{MAP-Elites Archive Comparisons Across 30 Runs of Evolution in Zelda Using \textbf{WWR}.} The three methods are compared as in Mario. Each triangular grid corresponds to levels with a particular number of reachable rooms (top-right of grid). Wall percentage increases upwards and water percentage increases to the right. Grids are triangular \hl{due to} a trade-off between water percentage and wall percentage (sum cannot exceed 100\%). 
(\subref{fig:zeldaWWRdiff}) Direct2GAN cannot produce levels with only one reachable room, but otherwise has high representation \hl{for smaller numbers} of reachable rooms and less representation \hl{for larger numbers}. CPPN2GAN is mostly comparable to CPPNThenDirect2GAN. However, CPPNThenDirect2GAN fails to reach certain bins that CPPN2GAN reaches when the number of reachable rooms is high, and is even sometimes beaten by Direct2GAN when the number of reachable rooms is small.
(\subref{fig:zeldaWWRBest}) CPPNThenDirect2GAN makes up for less coverage with higher \hl{fitness} scores in many bins, particularly as the number of reachable rooms grows. Bins with a ``Not Hybrid Tie'' represent a tie between CPPN2GAN and Direct2GAN. For certain small numbers of reachable rooms there are bins \hl{with a three-way} %where all three methods 
tie.}
\label{fig:zeldaWWR}
\end{figure*}

%Additionally, there is an interesting comparison with CPPNThenDirect2GAN.
Fig.~\ref{fig:zeldaWWR} shows results for \textbf{WWR}. Although these dungeons are smaller than those from previous work \cite{schrum:gecco2020cppn2gan}, results are consistent with the previous paper: CPPN2GAN is better than Direct2GAN. 
CPPNThenDirect2GAN is comparable to CPPN2GAN, \hl{but is not represented in some bins with many reachable rooms where CPPN2GAN is represented. CPPNThenDirect2GAN is also absent in some bins with few reachable rooms.} Direct2GAN is the method most underrepresented across all bins, but sometimes has the \hl{fittest} solutions. However, CPPNThenDirect2GAN generally has the most best solutions in bins with a large number of reachable rooms. Average heat maps in Fig.~A.4 (appendix). % \ref{fig:zeldaWWRaverageHeat}
\hl{Interestingly, only 85\% of Direct2GAN levels are beatable, whereas 91\% of CPPN2GAN and 94\% of CPPNThenDirect2GAN levels are.}

\begin{figure*}[tp]
\centering
\begin{subfigure}{0.48\textwidth} 
    \includegraphics[width=1.0\textwidth]{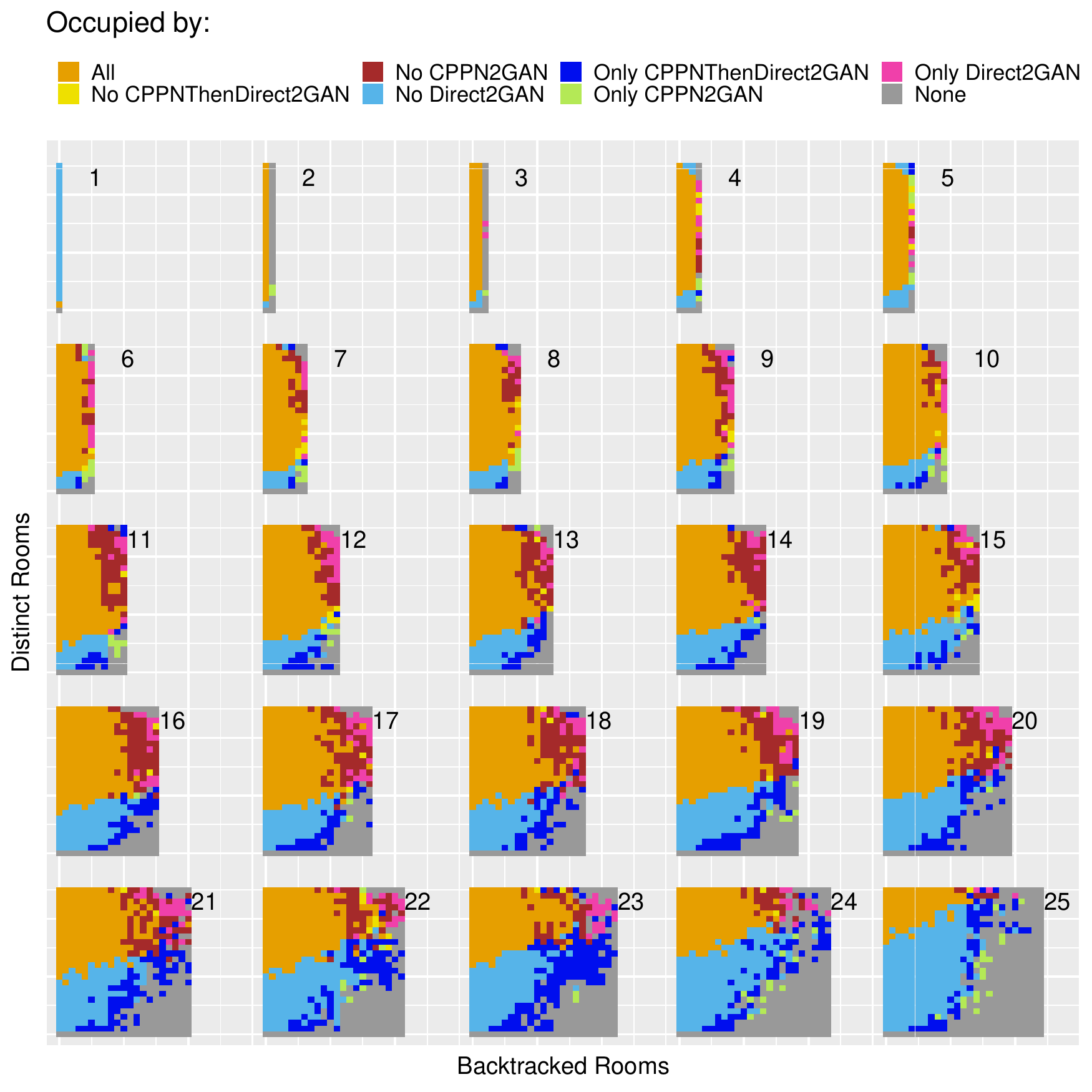}
    \caption{Difference in Bin Occupancy}
    \label{fig:zeldaDBRdiff}
\end{subfigure}
\begin{subfigure}{0.48\textwidth} 
    \includegraphics[width=1.0\textwidth]{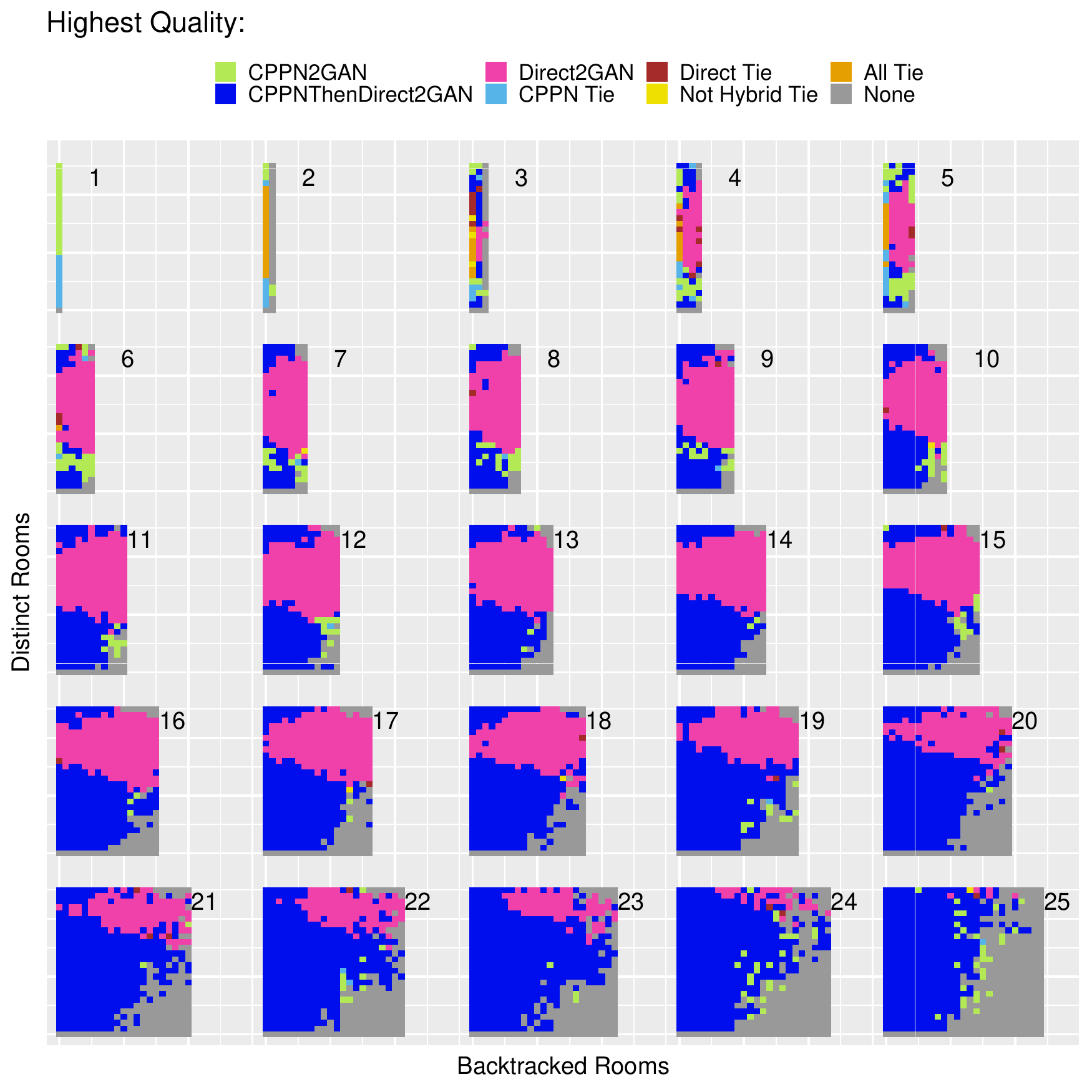}
    \caption{Best Occupant}
    \label{fig:zeldaDBRBest}
\end{subfigure}

\caption{\textbf{MAP-Elites Archive Comparisons Across 30 Runs of Evolution in Zelda Using \textbf{Distinct BTR}.} Methods are compared as in Fig.~\ref{fig:zeldaWWR}, but with \textbf{Distinct BTR}. The number by each sub-grid still represents the number of reachable rooms. Backtracking increases to the right and distinct rooms increases moving up. When there are fewer reachable rooms, less backtracking is possible, hence the changing sub-grid width. (\subref{fig:zeldaDBRdiff}) Dark blue regions show bins that only CPPNThenDirect2GAN reaches, though each method has some bins to themselves. Direct2GAN backtracks well in dungeons with many distinct rooms, but CPPN2GAN and CPPNThenDirect2GAN dominate bins for smaller numbers of distinct rooms.
(\subref{fig:zeldaDBRBest}) CPPN2GAN almost never produces the best result for a bin. There are regions where Direct2GAN does best. However, CPPNThenDirect2GAN is a clear winner in terms of the number of best bin occupants, especially as the number of reachable rooms increases. In some rare bins with a ``Direct Tie'' Direct2GAN and CPPNThenDirect2GAN are both the best.}
\label{fig:zeldaDBTR}
\end{figure*}

Fig.~\ref{fig:zeldaDBTR} \hl{has \textbf{Distinct BTR} results}. %This comparison shows that 
\hl{Although} CPPNThenDirect2GAN occupies more bins than either Direct2GAN or CPPN2GAN, each method occupies some bins that the others do not. However, CPPNThenDirect2GAN \hl{fitness} is superior to CPPN2GAN in nearly every bin where both are present, and superior %in \hl{fitness} 
to Direct2GAN in some cases \hl{too}. Average heat maps in Fig.~A.5 (appendix). %\ref{fig:zeldaDBRaverageHeat}
\hl{Direct2GAN averages 90\% beatable levels compared to 99\% for both CPPN-based methods.}

\begin{figure*}[th!]
\centering
\begin{subfigure}{0.3\textwidth} 
    \includegraphics[width=1.0\textwidth]{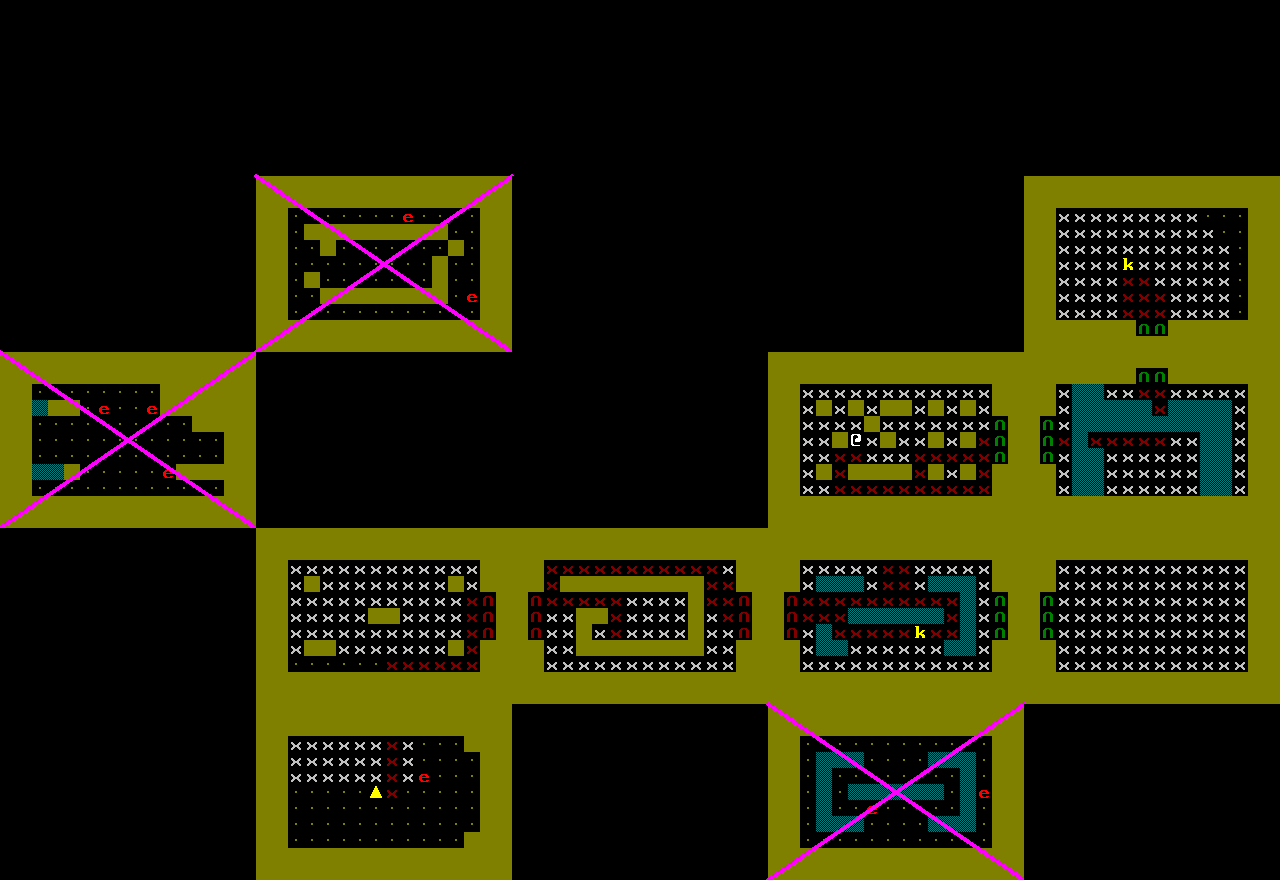}
    \caption{Direct2GAN. \hl{Distinct:~10, Backtracked:~6, Reachable:~8.}}
    \label{fig:zeldaDirect2GANChamp}
\end{subfigure} 
\begin{subfigure}{0.3\textwidth} 
    \includegraphics[width=1.0\textwidth]{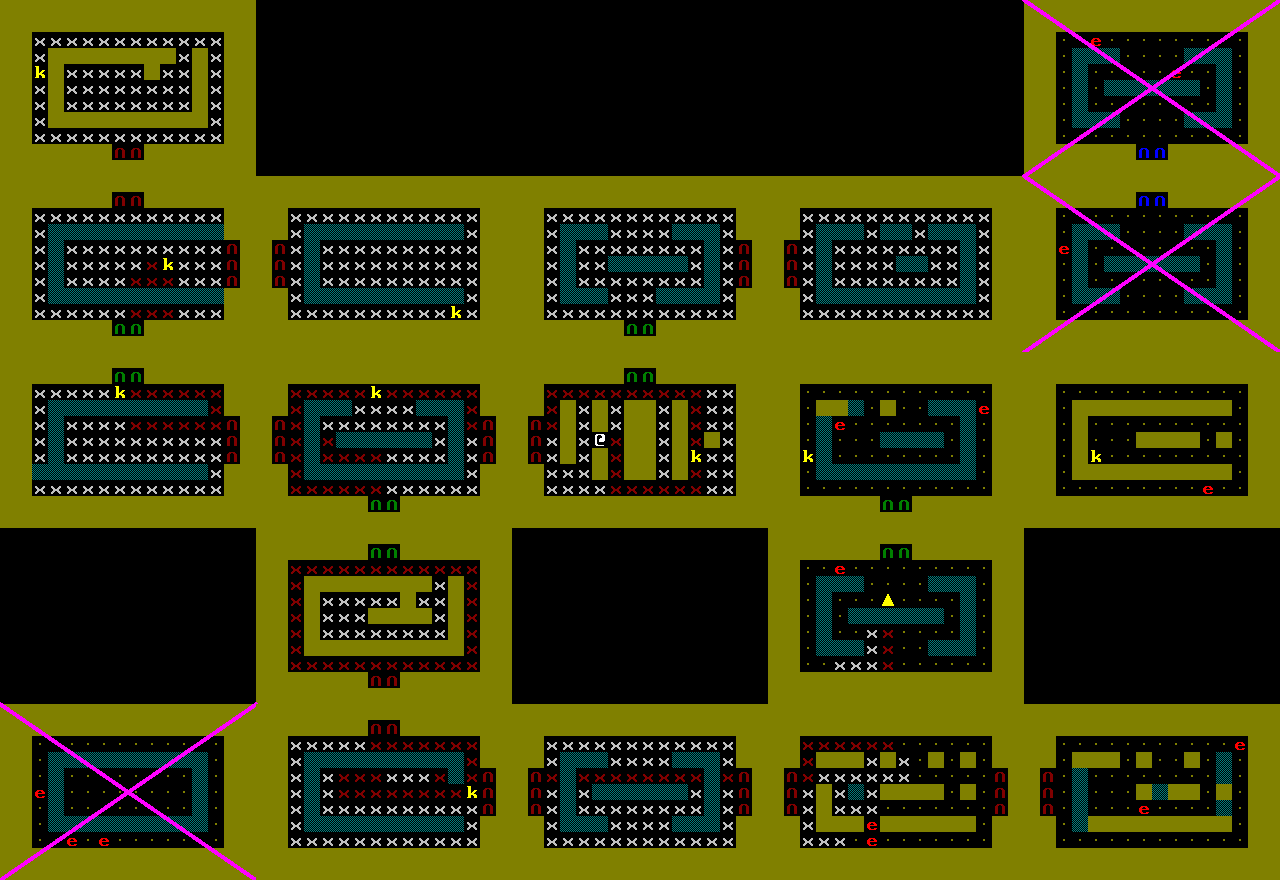}
    \caption{CPPN2GAN. \hl{Distinct:~16, Backtracked:~8, Reachable:~16.}}
    \label{fig:zeldaCPPN2GANChamp}
\end{subfigure}
\begin{subfigure}{0.3\textwidth} 
    \includegraphics[width=1.0\textwidth]{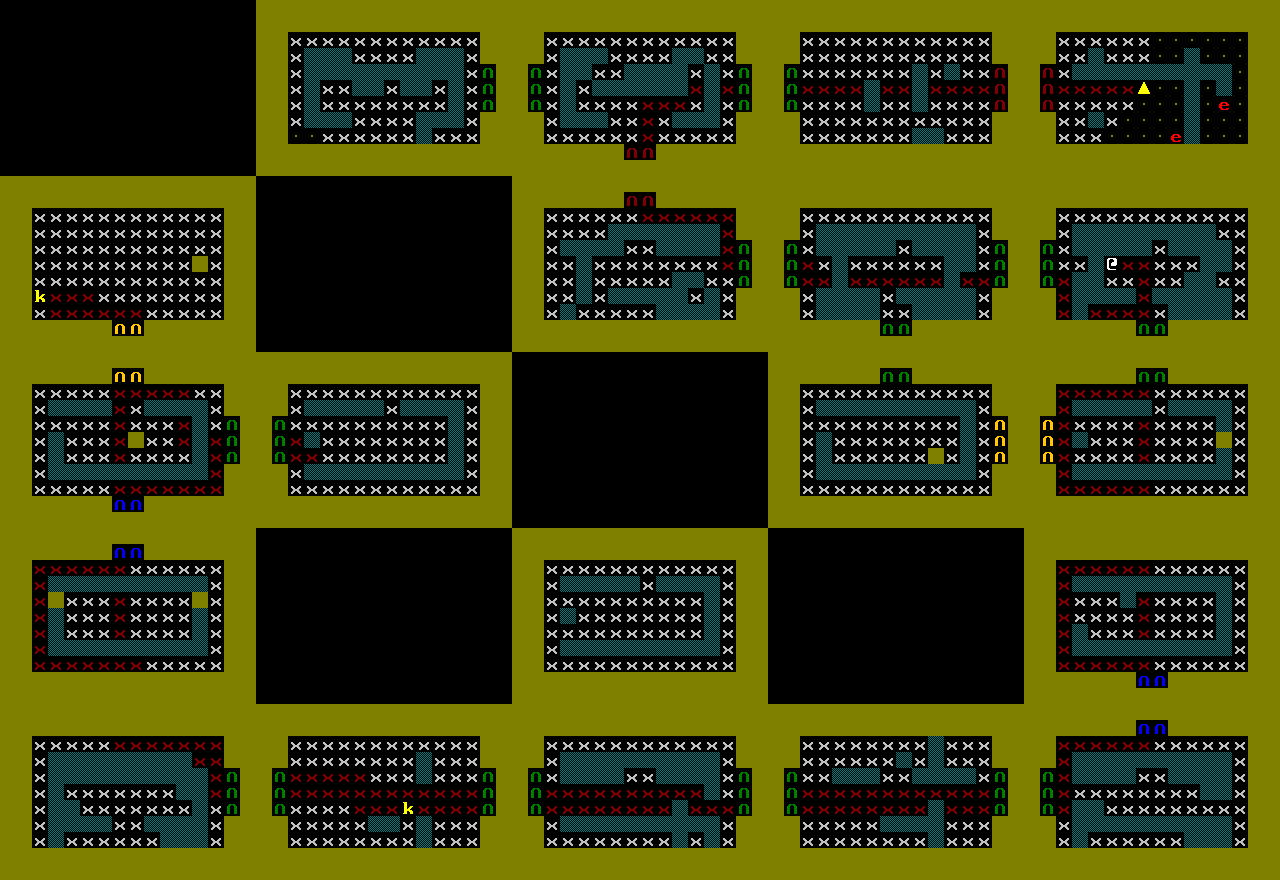}
    \caption{CPPNThenDirect2GAN \hl{(direct)}. \hl{Distinct: 18, Backtracked: 11, Reachable:~20.}}
    \label{fig:zeldaCPPNThenDirect2GANChamp}
\end{subfigure}

\caption{\textbf{Evolved Zelda Levels}. \normalfont (\subref{fig:zeldaDirect2GANChamp}) Direct2GAN generates levels with unique rooms and structure, but struggles to form a cohesive pattern. (\subref{fig:zeldaCPPN2GANChamp}) CPPN2GAN generates levels with symmetry or other global patterns. (\subref{fig:zeldaCPPNThenDirect2GANChamp}) CPPNThenDirect2GAN generates levels with interesting overall patterns like CPPN2GAN, but can tweak individual rooms like Direct2GAN to increase overall fitness.}
\label{fig:zeldaChamps}
\end{figure*}

CPPNThenDirect2GAN archives can be analyzed in terms of their genotype composition (Fig.~A.6 in appendix). The two Mario binning schemes end up with more CPPNs, and the two Zelda schemes end up with more direct vectors, but no type of genotype goes extinct with any approach. The distribution of genotypes within each archive is fairly consistent for the Mario \textbf{Sum DSL} scheme and both Zelda schemes (Fig.~A.7 in appendix). Oddly, Mario \textbf{Distinct ASAD} shows more variation, and some distinct examples are shown in Fig.~A.8 (appendix). These results show that CPPNThenDirect2GAN is highly adaptable to different diversity characterizations.

Among generated levels,
global patterns in Zelda dungeons from CPPN2GAN and CPPNThenDirect2GAN stand out immediately. There is often a symmetrical or repeating motif in these dungeons that is missing in results from Direct2GAN. Even when all rooms in a CPPN-generated dungeon are distinct, there is often a theme throughout the dungeon, such as a water-theme, or reuse of maze-like wall obstacles. Some examples are in Fig.~\ref{fig:zeldaChamps},
with more in the appendix (Fig.~A.9). % \ref{fig:zeldaBinLevels} 

Mario levels produced by CPPNs tend to have too much repetition, except when the number of distinct segments is an explicit dimension of variation. When encouraged to be distinct, CPPN levels have more variety, yet maintain a theme of similar elements. An example of a repeated theme is an arrangement of blocks that presents a particular jumping challenge, but requires a slightly different approach on each occurrence due to small variations. See appendix (Fig.~A.10). % \ref{fig:marioBinLevels}

\section{\hl{Discussion} and Future Work}

%the first method that can create large-scale game levels

%Generative Adversarial Networks (GANs) have shown impressive results as generators for high-quality images. 

%However, 

\hl{Combining} multiple GAN-generated \hl{segments} into a cohesive whole %, which is especially relevant for level generation in games, 
was %so far an unexplored 
an \hl{under explored research area}. Our previous work \cite{schrum:gecco2020cppn2gan} presented \hl{a method for creating} large %-scale 
game levels \hl{by combining CPPNs with} a GAN, %. One of the main insights is 
revealing %that there is 
a functional relationship between the latent vectors of different game segments, which the CPPN \hl{exploits}. %In 
\hl{Compared} to a direct representation of multiple latent vectors, CPPN2GAN %can 
\hl{generates} a larger variety of different complete game levels. CPPNThenDirect2GAN %goes one step further to 
\hl{incorporates} strengths of both approaches.

Our original work showed that CPPN2GAN results often contained repeated segments, and those results are repeated here. However, the additional experiments in this paper demonstrate that CPPN2GAN can generate diverse segments within a single level if distinct segments are a dimension of variation for MAP-Elites. CPPN2GAN levels can easily evolve any number of distinct segments within a level, but Direct2GAN has trouble repeating segments within a level.  

The new hybrid approach introduced in this paper is CPPNThenDirect2GAN. 
First, CPPN2GAN introduces global structure and imposes regular patterns in the level. After mutation, Direct2GAN can take over to fine-tune levels and/or introduce more local variety. For the binning schemes in Mario, this approach is comparable to CPPN2GAN, which already fills many bins in the archive. However, CPPNThenDirect2GAN produced interesting results in Zelda, particularly in the new \textbf{Distinct BTR} scheme where it outperformed both Direct2GAN and CPPN2GAN.
Specifically, CPPNThenDirect2GAN filled many bins that neither of the other methods filled, and had the \hl{most fit} levels in many bins as well.

%The conclusion of all of this is that although CPPNThenDirect2GAN can reach some areas of the archive that neither Direct2GAN nor CPPN2GAN can, it is also hindered from reaching all of the bins that each of these methods reach by themselves. 

%%%% This narrative is unnecessary when using QD score as primary measure of performance
%However, in the \textbf{WWR} scheme from the previous paper \cite{schrum:gecco2020cppn2gan}, CPPNThenDirect2GAN is slightly worse than CPPN2GAN in terms of the number of bins filled. Also, for both Zelda binning schemes, there are bins filled by CPPN2GAN or Direct2GAN that CPPNThenDirect2GAN was unable to fill. 

%However, in the \textbf{Distinct BTR} scheme, CPPNThenDirect2GAN filled some bins that neither of the other methods filled. The conclusion of all of this is that although CPPNThenDirect2GAN can reach some areas of the archive that neither Direct2GAN nor CPPN2GAN can, it is also hindered from reaching all of the bins that each of these methods reach by themselves. 

%% However, there would always be a risk of one type replacing the other in any given bin. This is problematic if it eliminates useful stepping stones. Certain bins might only be easily reachable via a Direct or CPPN genome. Our results further show that it is the hybridization that makes some levels discoverable, as they are not present in the combined archives of either CPPN2GAN or Direct2GAN.

However, Direct2GAN \hl{reaches} some bins that CPPNThenDirect2GAN could not.
These bins \hl{might be reachable} if CPPNThenDirect2GAN genomes were not forced to start as CPPNs. Starting as a CPPN may introduce a bias towards patterns that is so strong that subsequent direct vector manipulation has trouble breaking the patterns. Therefore, the initial population could simply be a combination of CPPN and direct genomes. 
\hl{In fact, the archive size could be doubled to allow separate archives for direct and CPPN genomes, with the benefit of CPPNs occasionally mutating into direct vectors.}

%This idea is similar to HybrID \cite{CluneBPO09}. % YES! \todo{Is this still correct?}

\hl{CPPNThenDirect2GAN is inspired by HybrID {\cite{CluneBPO09:HybrID}}, but later research introduced Offset-HybrID {\cite{helms:pone2017}}: instead of transitioning to direct vectors, one is evolved with each CPPN. Vector components are offsets added to CPPN output. Applied to game levels, this approach would still generate large-scale patterns while also allowing local variation.}

%% Was going to mention \cite{helms:pone2017} but too niche

%Unfortunately, there is no easy way to incorporate a Direct genome back into a CPPN genome. Furthermore, 

But direct, hybrid, and offset genomes %lose the ability to 
\hl{cannot} scale to arbitrary sizes; a major benefit of CPPNs. Plain CPPN2GAN could enable levels %in which components are 
\hl{with components} %lazily 
generated as needed: %in 
levels \hl{would} never stop growing. This would be especially useful for exploration games, as new segments %(e.g.\ dungeon rooms) 
can be served by CPPN2GAN whenever new areas of the map are discovered. % by the player. 
Evaluating this special scenario, and finding a way to incorporate the benefits of directly encoded components, is an interesting area for future work. Further work is also planned on characterizing different binning schemes and their relationship to the performance of different level generators.

%There is no need to confine a dungeon to a grid of any size. If the player starts at coordinates $(0,0)$ then CPPN2GAN can serve up a new room whenever new grid coordinates are uncovered. 

%Besides different combination approaches, 

\hl{Modifications} to the training process are also possible. % Here, the 
\hl{Our} GAN was pre-trained and only CPPNs evolved.
%However, it should be possible to make use of 
\hl{Instead,}
a discriminator %to also 
\hl{could} decide whether global patterns %(such as Zelda room structure) 
are similar to original levels %. This would mean 
\hl{via} adversarial training %of 
\hl{against} the complete CPPN2GAN network. % against a discriminator. 
Training could be end-to-end or by alternating between the CPPN and the generator. \hl{The generator could even be represented by an autoencoder or some other non-GAN approach.} The resulting samples should be able to reproduce both global \emph{and} local patterns in complete game levels rather than just in individual segments.

%%% No room for HyperNetworks
%\todo{should we mention HyperNetworks here?}

%Is it possible to train the GAN with the CPPN in mind in order to prove results? Perhaps complete end-to-end training of the CPPN and GAN combination could be used to actually learn global \emph{and} local patterns in complete game levels rather than in individual segments.

%%% See above
 %training a GAN that is particularly well suited for CPPNs? - generate not just maps but complete games? - End-to-end training of CPPN+GAN?

%\todo[inline]{Sebastian or Vanessa: Do one of you want to write this up?: other application areas - fractal levels? CPPNs can also interpolate -}

\section{\hl{Conclusion}}

\hl{CPPNThenDirect2GAN is a new hybrid approach combining the
benefits of CPPN2GAN and Direct2GAN. Whereas CPPN2GAN 
generates global patterns with GAN-generated segments,
CPPNThenDirect2GAN allows for additional variation in segments, increasing the expressive range of
generated levels in some situations, e.g.~Zelda. CPPNThenDirect2GAN
could be useful in other game domains and situations
requiring global organization of GAN-generated segments.}

\section*{Acknowledgment}

The authors thank Schloss Dagstuhl and
the organisers of Dagstuhl Seminars 17471 and 19511 for 
productive seminars. \hl{They} also thank %the %Summer Collaborative Opportunities and Experiences (SCOPE) 
\hl{Southwestern University's SCOPE program}
for supporting %continuation of
this research with undergraduate researchers.

\ifCLASSOPTIONcaptionsoff
  \newpage
\fi



%\bibliographystyle{IEEEtran}
\bibliographystyle{IEEEtran}

\onecolumn

\appendix

\renewcommand\thefigure{\thesection.\arabic{figure}}    

\section{Appendix}
\setcounter{figure}{0}  

\begin{figure*}[h!]
\centering
\begin{subfigure}{0.49\textwidth} 
    \includegraphics[width=1.0\textwidth]{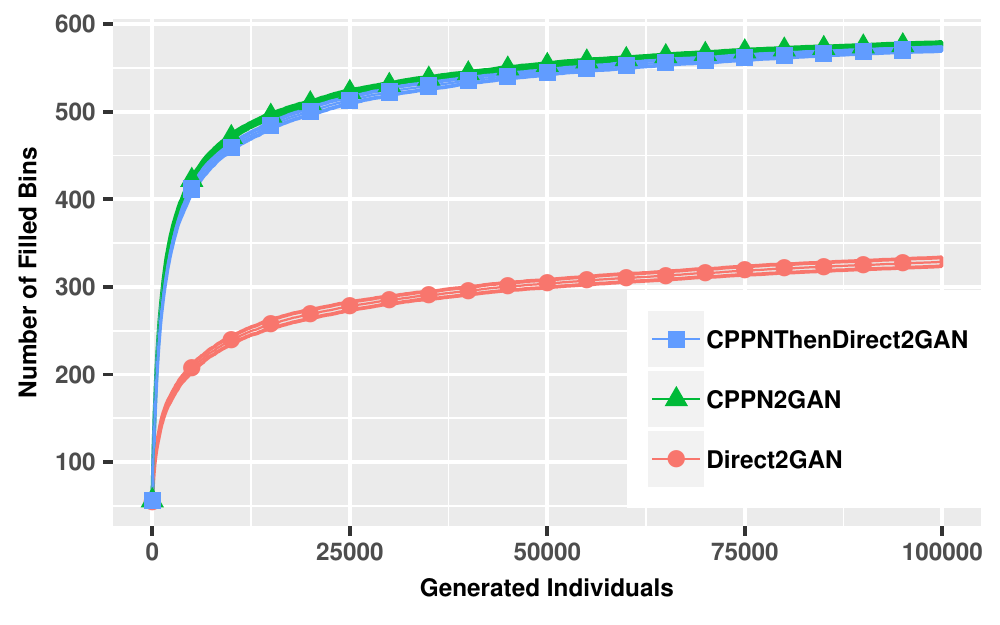}
    \caption{Mario: \textbf{Sum DSL}}
    \label{fig:marioDNS}
\end{subfigure}
\begin{subfigure}{0.49\textwidth} 
    \includegraphics[width=1.0\textwidth]{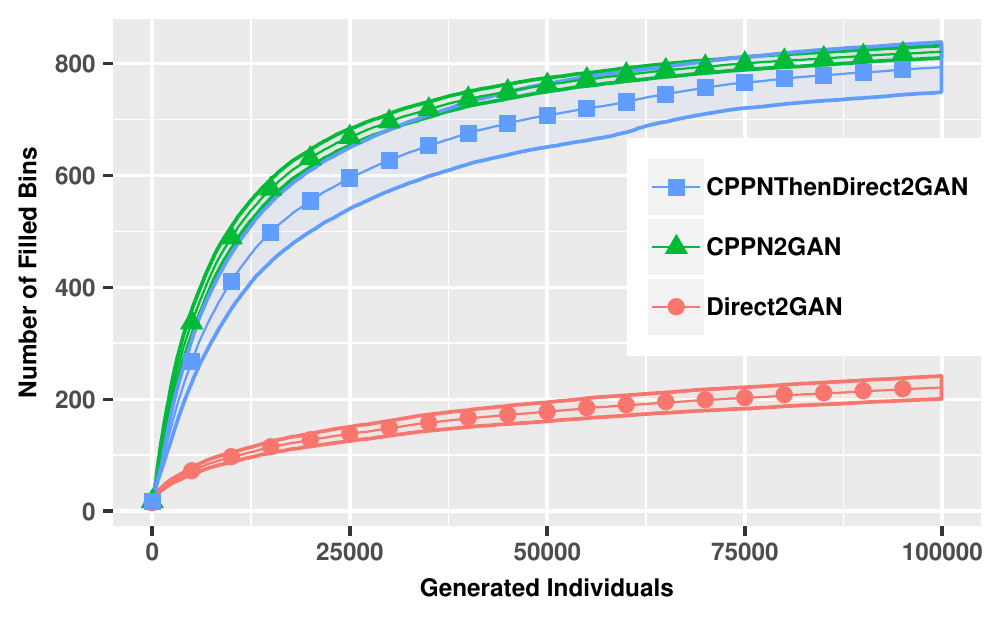}
    \caption{Mario: \textbf{Distinct ASAD}}
    \label{fig:marioDND}
\end{subfigure} \\
\begin{subfigure}{0.49\textwidth} 
    \includegraphics[width=1.0\textwidth]{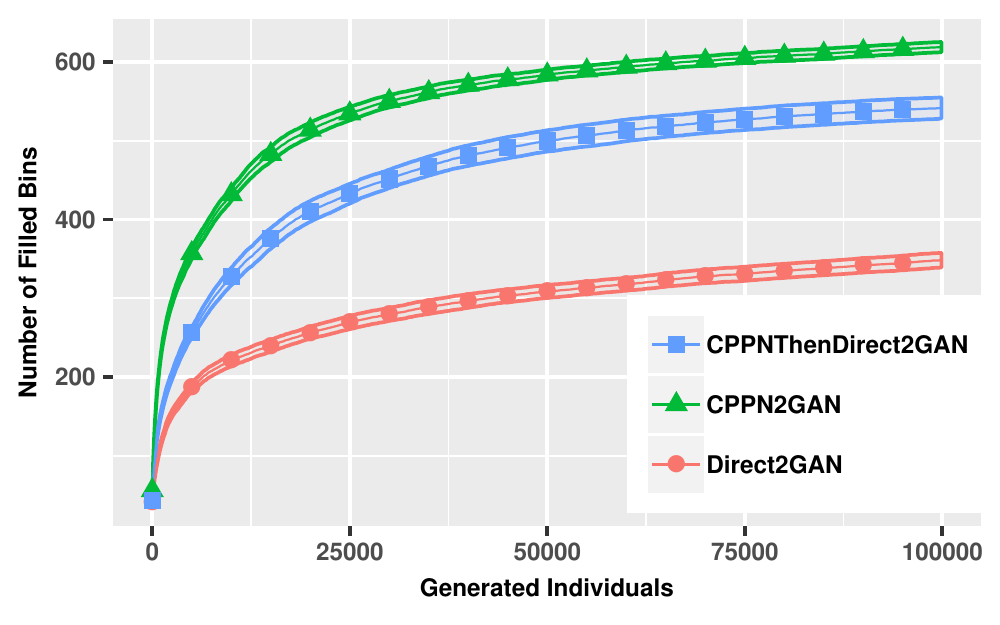}
    \caption{Zelda: \textbf{WWR}}
    \label{fig:zeldaWWRfill}
\end{subfigure}
\begin{subfigure}{0.49\textwidth} 
    \includegraphics[width=1.0\textwidth]{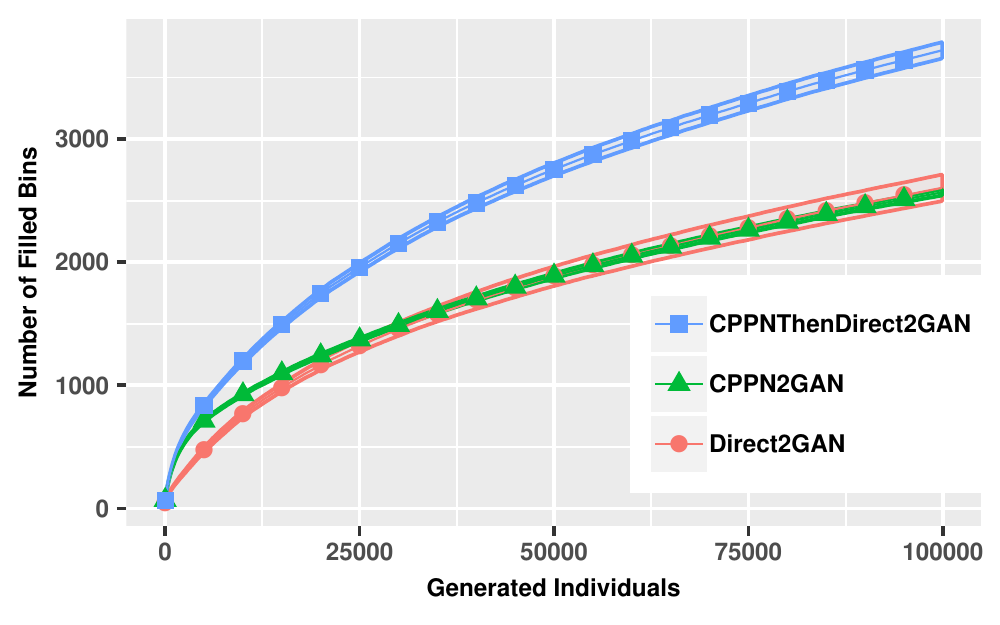}
    \caption{Zelda: \textbf{Distinct BTR}}
    \label{fig:zeldaDBRfill}
\end{subfigure}

\caption{\textbf{Average Filled Bins Across 30 Runs of MAP-Elites.} \normalfont For the two distinct MAP-Elites binning schemes used in both Mario and Zelda, plots of the average number of bins filled with 95\% confidence intervals demonstrate the comparative performance of the three encoding schemes. For both binning schemes used in Mario, CPPN-based approaches are vastly superior to Direct2GAN. CPPN2GAN and CPPNThenDirect2GAN are statistically tied, though CPPNThenDirect2GAN has larger confidence intervals for the \textbf{Distinct ASAD} binning scheme. There is a starker contrast in the Zelda results. Plain CPPN2GAN is slightly superior to CPPNThenDirect2GAN for the \textbf{WWR} binning scheme, though both approaches are vastly superior to Direct2GAN. For the \textbf{Distinct BTR} scheme, CPPNThenDirect2GAN is the best, with plain CPPN2GAN and Direct2GAN tied for worst. Although the different approaches have their strengths and weaknesses, which are dependent on the binning scheme, it is clear that using some form of CPPN-based encoding leads to the best results.}
\label{fig:averageBins}
\end{figure*}

\begin{figure*}[h!]
\centering
\begin{subfigure}{0.49\textwidth} 
    \includegraphics[width=1.0\textwidth]{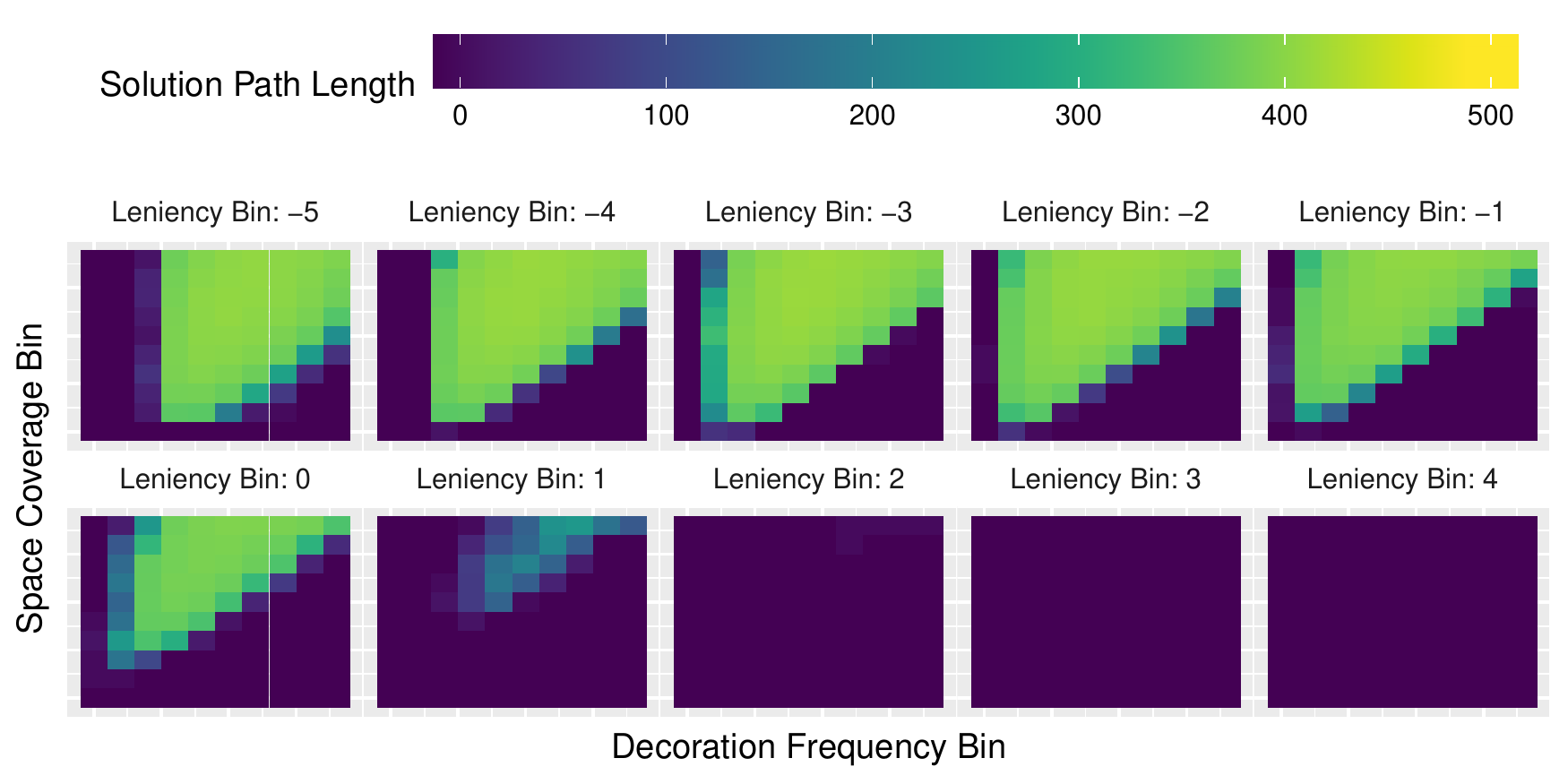}
    \caption{Direct2GAN}
    \label{fig:marioDNSheatDiredt2GAN}
\end{subfigure}
\begin{subfigure}{0.49\textwidth} 
    \includegraphics[width=1.0\textwidth]{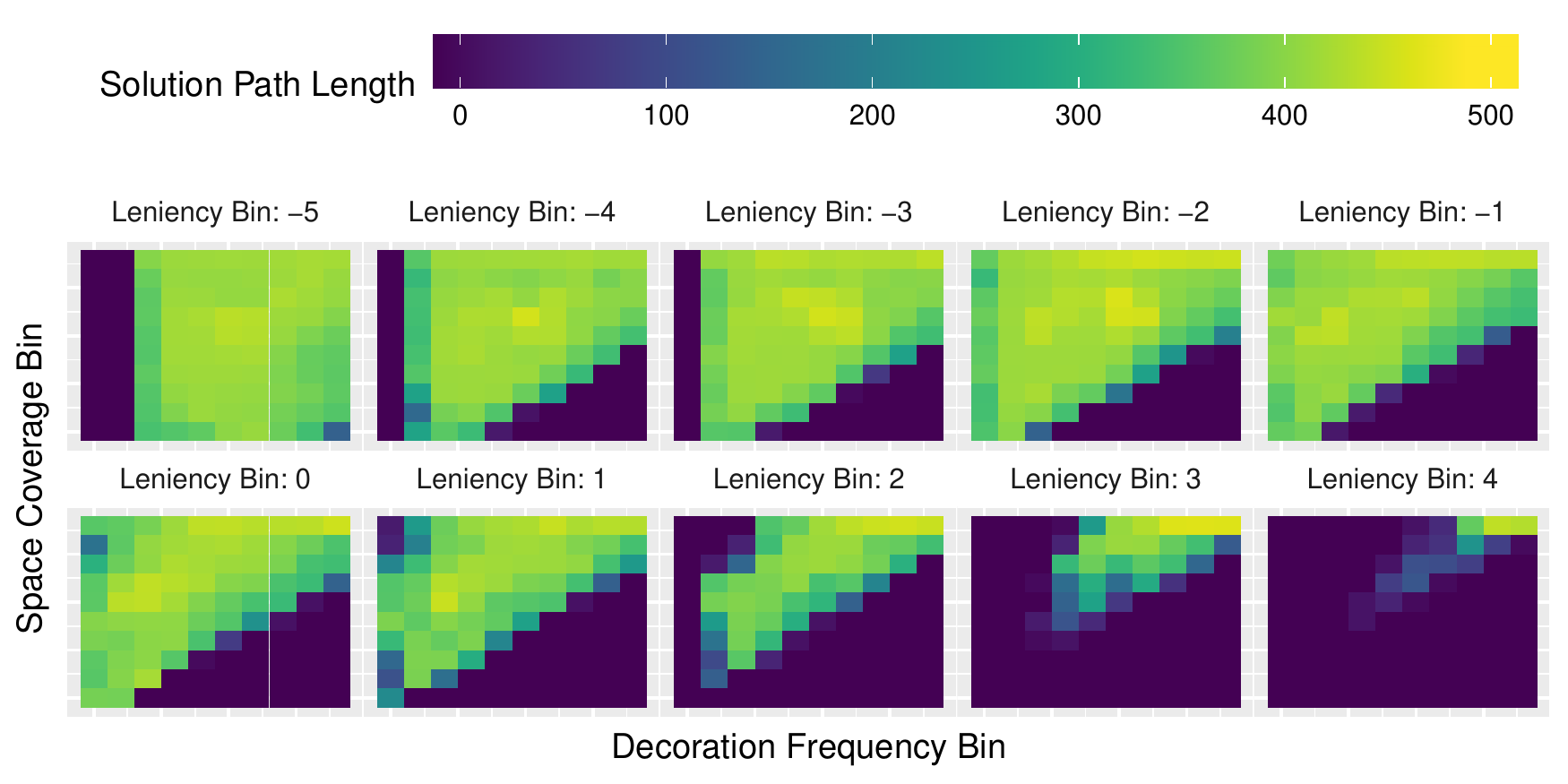}
    \caption{CPPN2GAN}
    \label{fig:marioDNSheatCPPN2GAN}
\end{subfigure} \\

\begin{subfigure}{0.49\textwidth} 
    \includegraphics[width=1.0\textwidth]{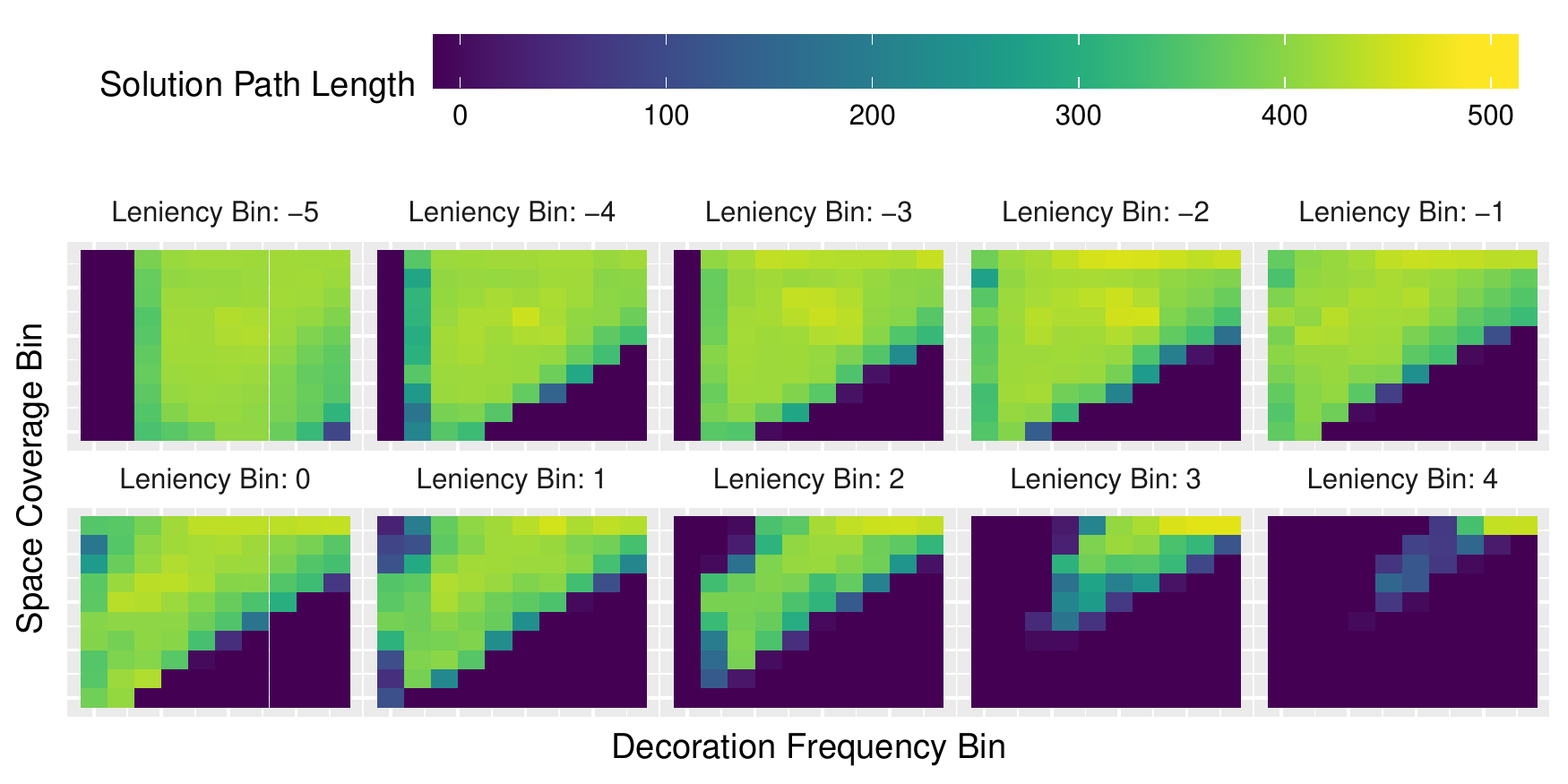}
    \caption{CPPNThenDirect2GAN}
    \label{fig:marioDNSheatCPPNThenDirect2GAN}
\end{subfigure}
\begin{subfigure}{0.49\textwidth} 
    \includegraphics[width=1.0\textwidth]{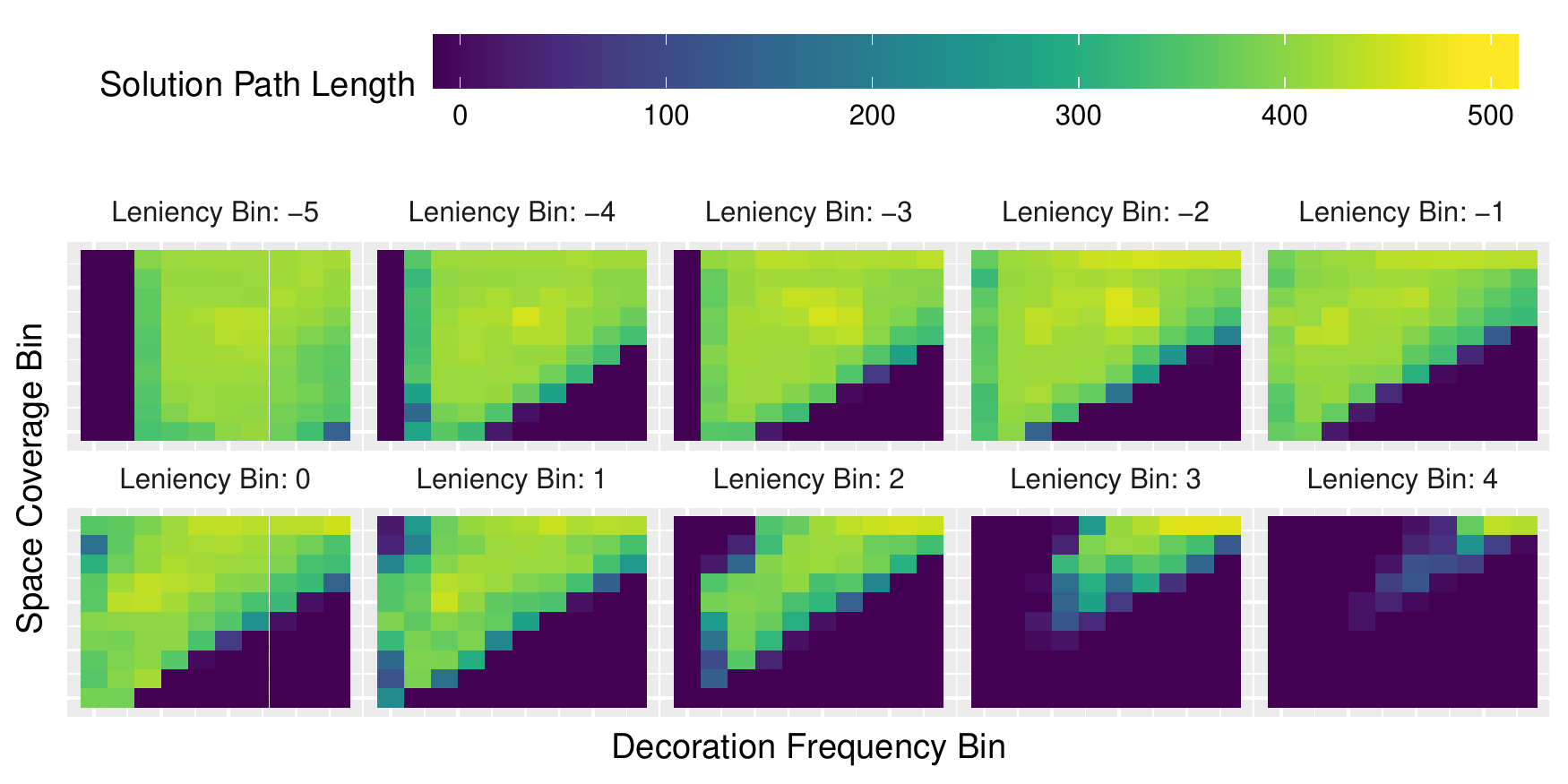}
    \caption{Max of Direct2GAN and CPPN2GAN}
    \label{fig:marioDNSheatCombined}
\end{subfigure}

\caption{\textbf{Average MAP-Elites Bin Fitness Across 30 Runs of Evolution in Mario Using \textbf{Sum DSL}.} The first three figures average fitness scores across 30 runs within each bin. Each sub-grid represents a different leniency score. Within each sub-grid, summed decoration frequency increases to the right, and summed space coverage increases moving up. (\subref{fig:marioDNSheatDiredt2GAN}) Direct2GAN leaves many bins completely absent. (\subref{fig:marioDNSheatCPPN2GAN}) CPPN2GAN and (\subref{fig:marioDNSheatCPPNThenDirect2GAN}) CPPNThenDirect2GAN are comparable to each other, and fill in many of the bins left empty by Direct2GAN. 
(\subref{fig:marioDNSheatCombined}) The final figure shows the result of combining Direct2GAN with CPPN2GAN and taking the best average result for each bin. The combined results mostly depend on CPPN2GAN, which is very similar to CPPNThenDirect2GAN.}
\label{fig:marioDNSaverageHeat}
\end{figure*}

\begin{figure*}[h!]
\centering
\begin{subfigure}{0.49\textwidth} 
    \includegraphics[width=1.0\textwidth]{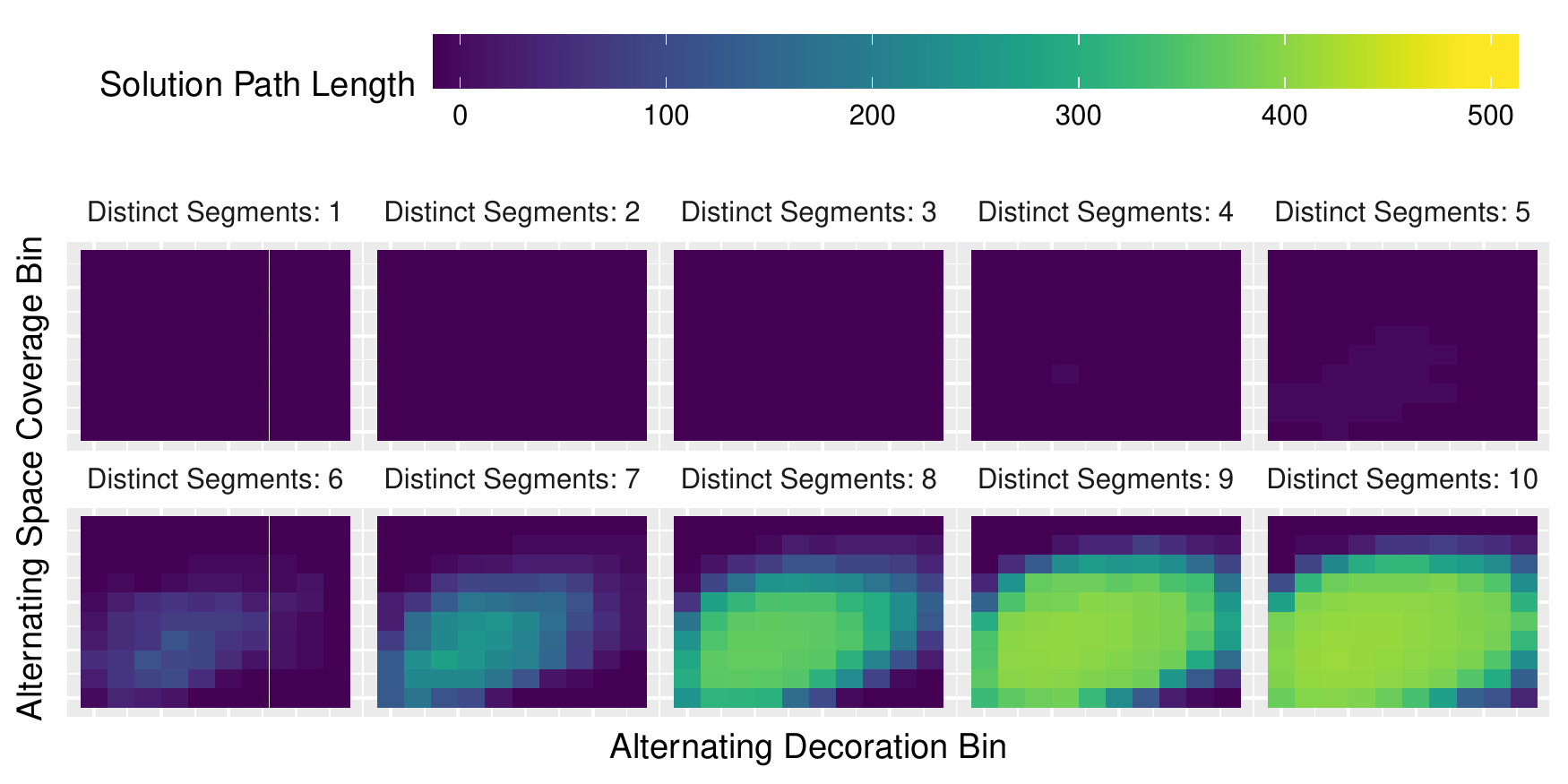}
    \caption{Direct2GAN}
    \label{fig:marioDNDheatDiredt2GAN}
\end{subfigure}
\begin{subfigure}{0.49\textwidth} 
    \includegraphics[width=1.0\textwidth]{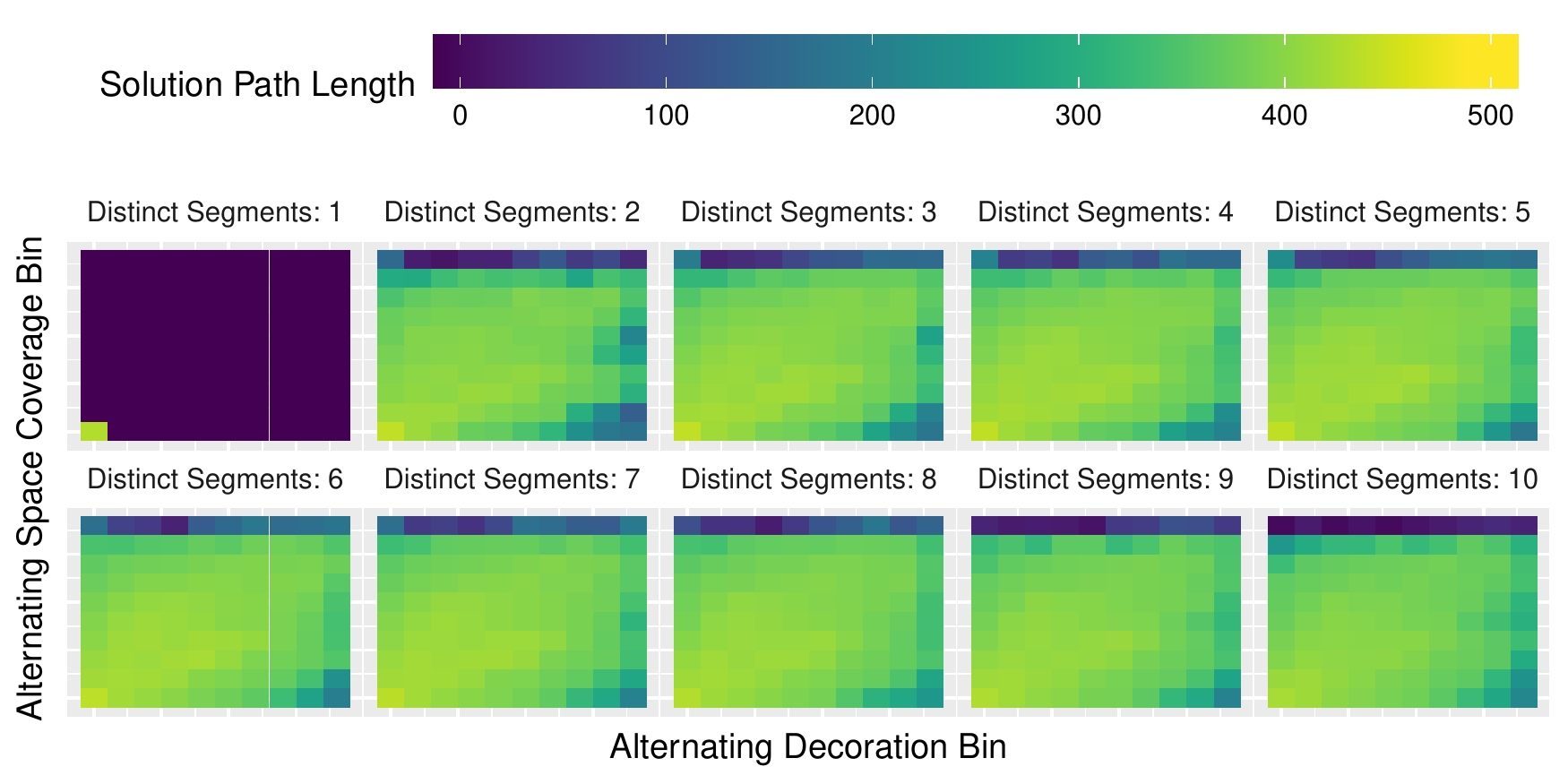}
    \caption{CPPN2GAN}
    \label{fig:marioDNDheatCPPN2GAN}
\end{subfigure} \\

\begin{subfigure}{0.49\textwidth} 
    \includegraphics[width=1.0\textwidth]{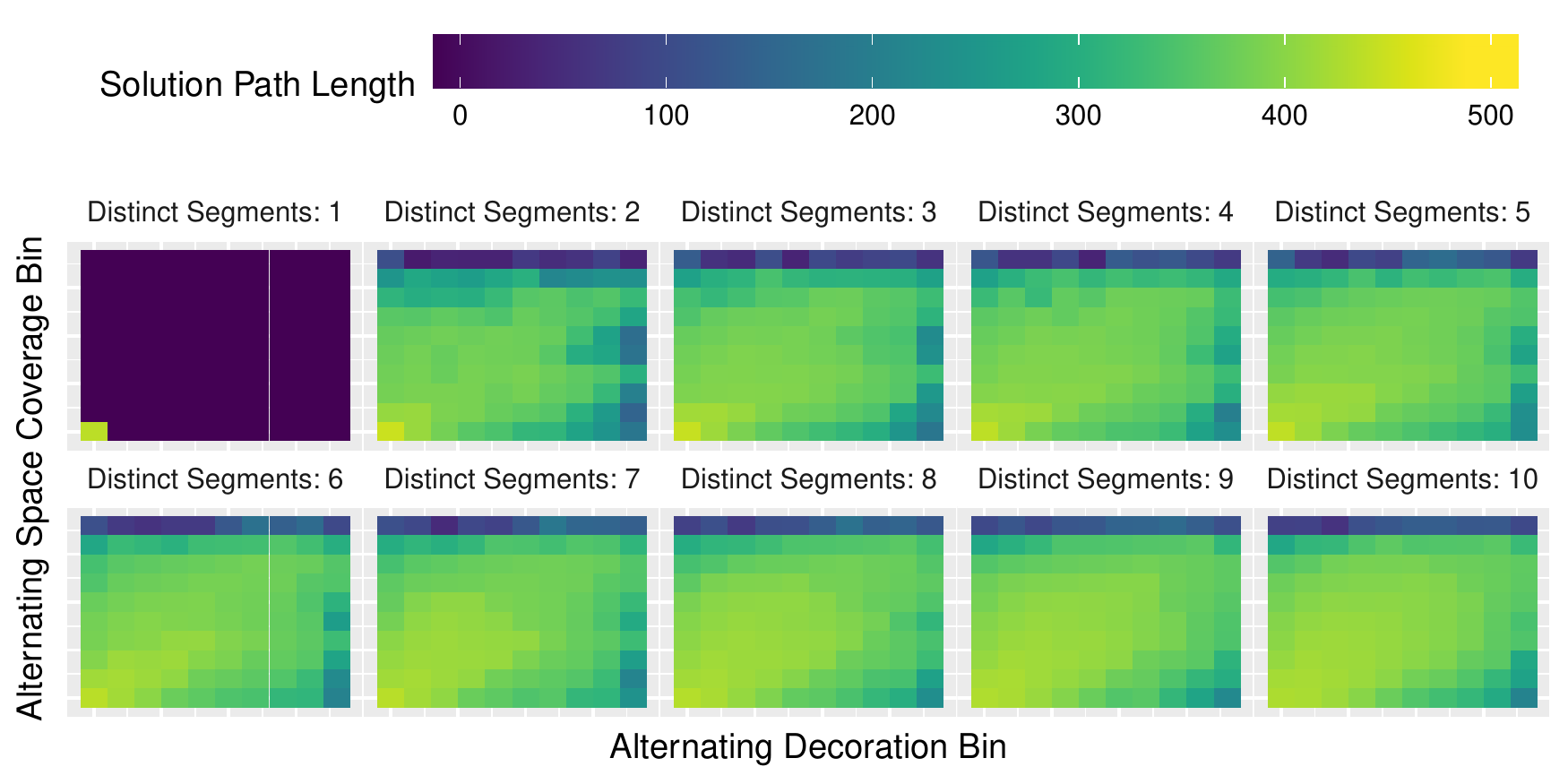}
    \caption{CPPNThenDirect2GAN}
    \label{fig:marioDNDheatCPPNThenDirect2GAN}
\end{subfigure}
\begin{subfigure}{0.49\textwidth} 
    \includegraphics[width=1.0\textwidth]{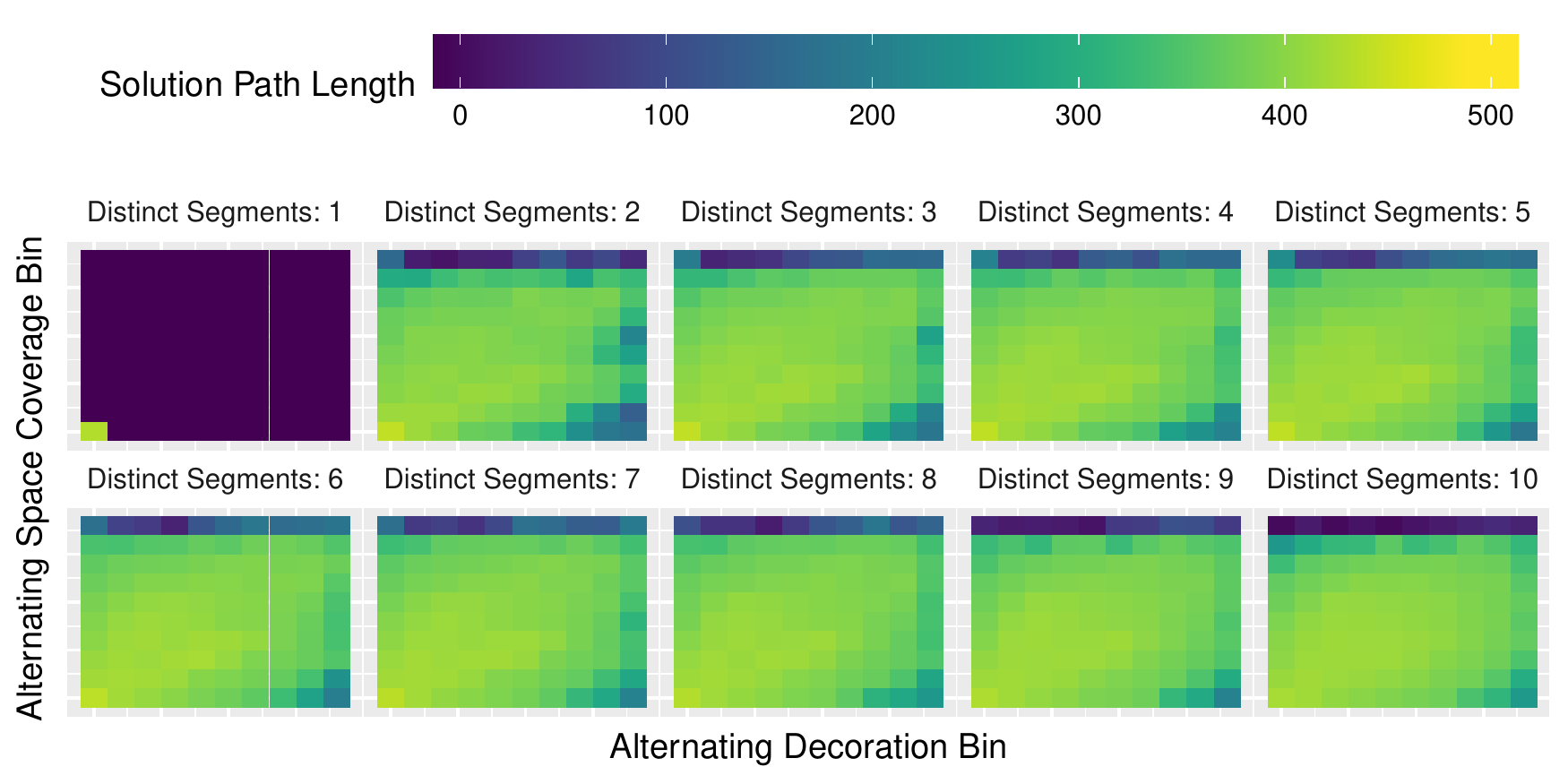}
    \caption{Max of Direct2GAN and CPPN2GAN}
    \label{fig:marioDNDheatCombined}
\end{subfigure}

\caption{\textbf{Average MAP-Elites Bin Fitness Across 30 Runs of Evolution in Mario Using \textbf{Distinct ASAD}.} All three methods are compared as in Fig.~\ref{fig:marioDNSaverageHeat}, but using \textbf{Distinct ASAD}. Each sub-grid now represents the count of distinct segments in the level. Within each sub-grid, alternating decoration score increases to the right, and alternating space coverage score increases moving up. The upper-left grid is mostly empty, since it corresponds to levels with only one repeated segment. When the exact same segment is repeating, it is impossible for decoration or space coverage scores to alternate from segment to segment. (\subref{fig:marioDNDheatDiredt2GAN}) Direct2GAN leaves even more bins unoccupied with this scheme, particularly those with a small number of distinct segments. \hl{Fitness} scores also decrease as the number of distinct segments decreases. (\subref{fig:marioDNDheatCPPN2GAN}) CPPN2GAN and (\subref{fig:marioDNDheatCPPNThenDirect2GAN}) CPPNThenDirect2GAN are once again nearly identical, but CPPNThenDirect2GAN has some brighter areas for larger numbers of distinct segments. (\subref{fig:marioDNDheatCombined}) The combination of Direct2GAN and CPPN2GAN has the coverage of CPPN2GAN, but includes the brighter cells from Direct2GAN when there are 9 or 10 distinct segments.}
\label{fig:marioDNDaverageHeat}
\end{figure*}

\begin{figure*}[t!]
\centering
\begin{subfigure}{0.49\textwidth} 
    \includegraphics[width=1.0\textwidth]{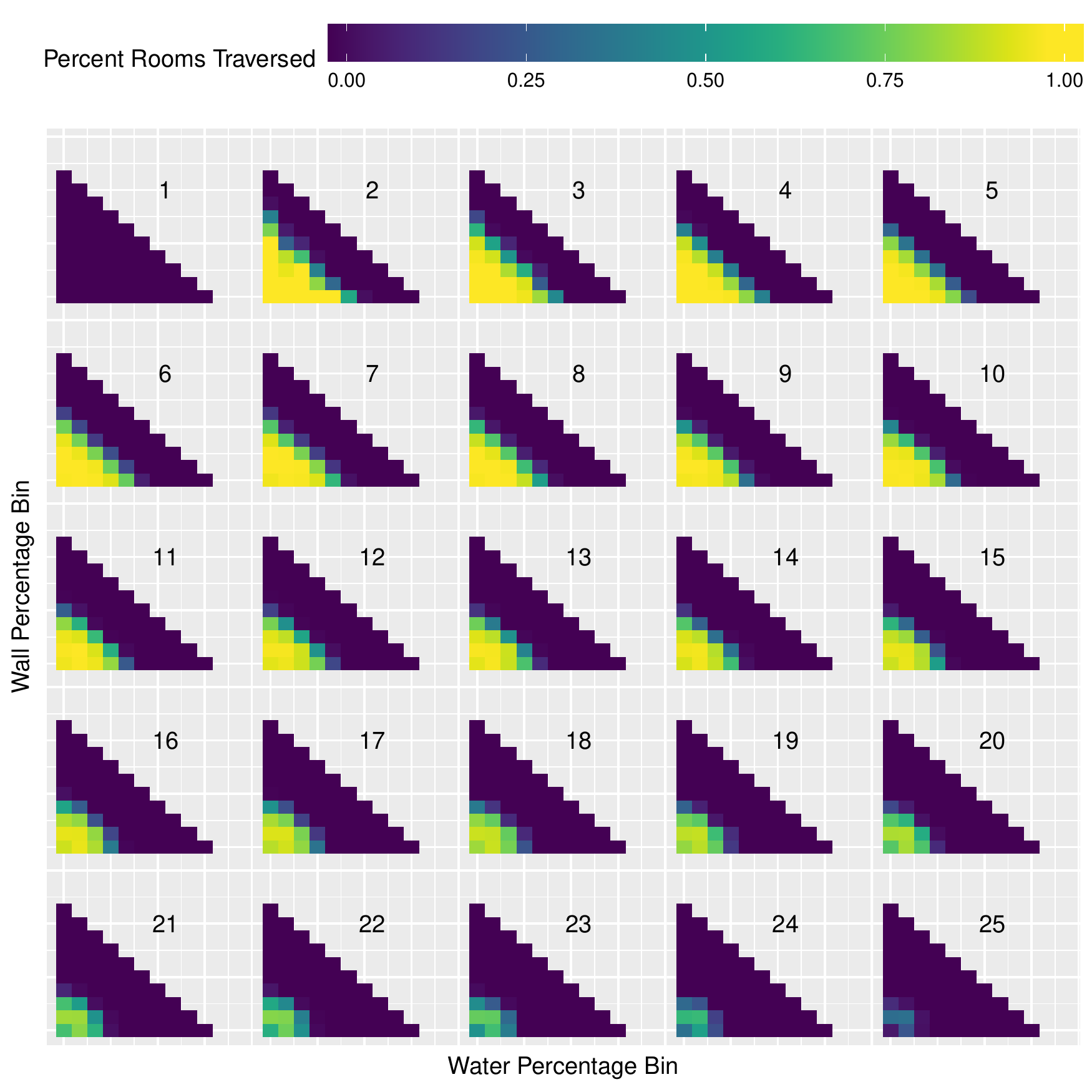}
    \caption{Direct2GAN}
    \label{fig:zeldaWWRheatDirect2GAN}
\end{subfigure}
\begin{subfigure}{0.49\textwidth} 
    \includegraphics[width=1.0\textwidth]{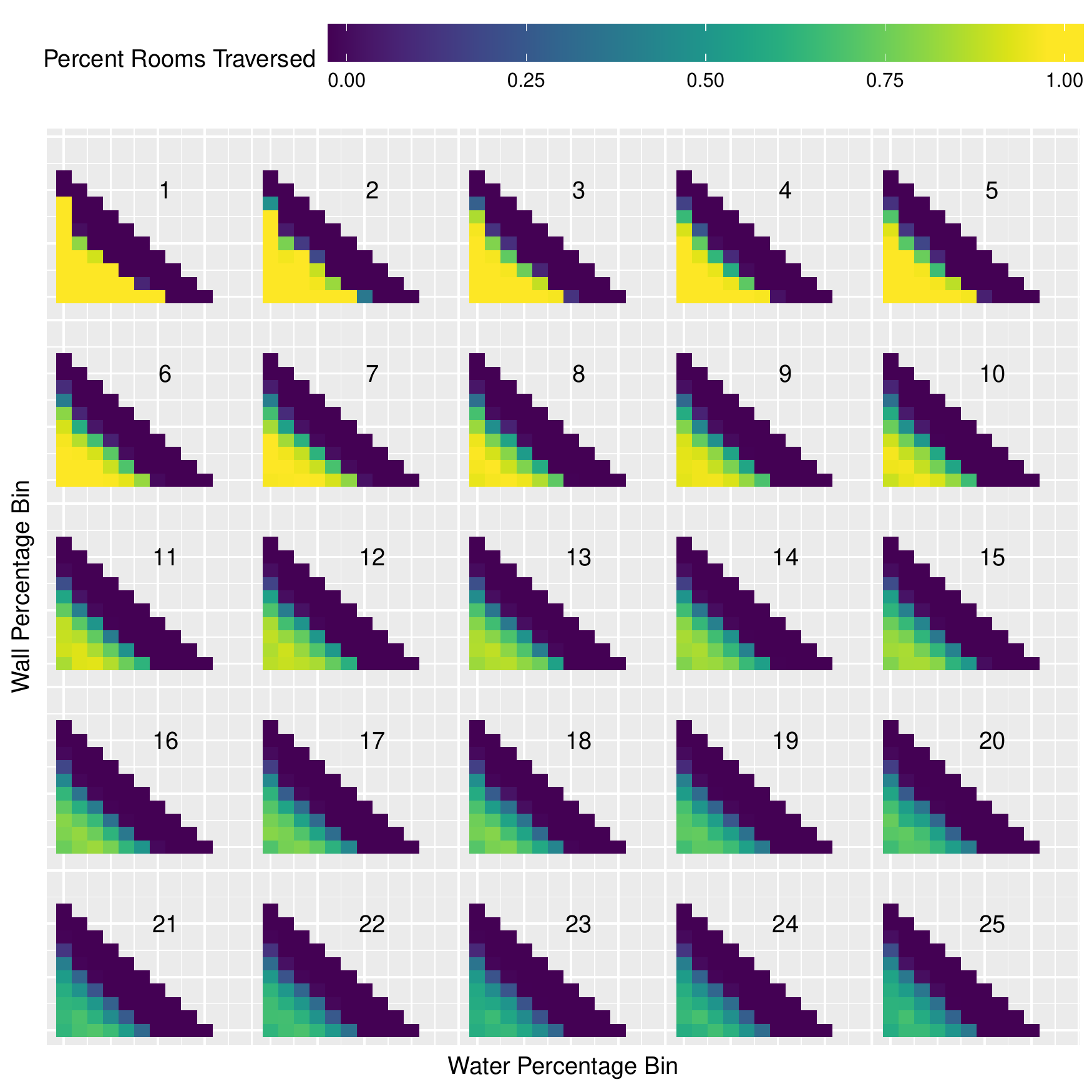}
    \caption{CPPN2GAN}
    \label{fig:zeldaWWRheatCPPN2GAN}
\end{subfigure} \\

\begin{subfigure}{0.49\textwidth} 
    \includegraphics[width=1.0\textwidth]{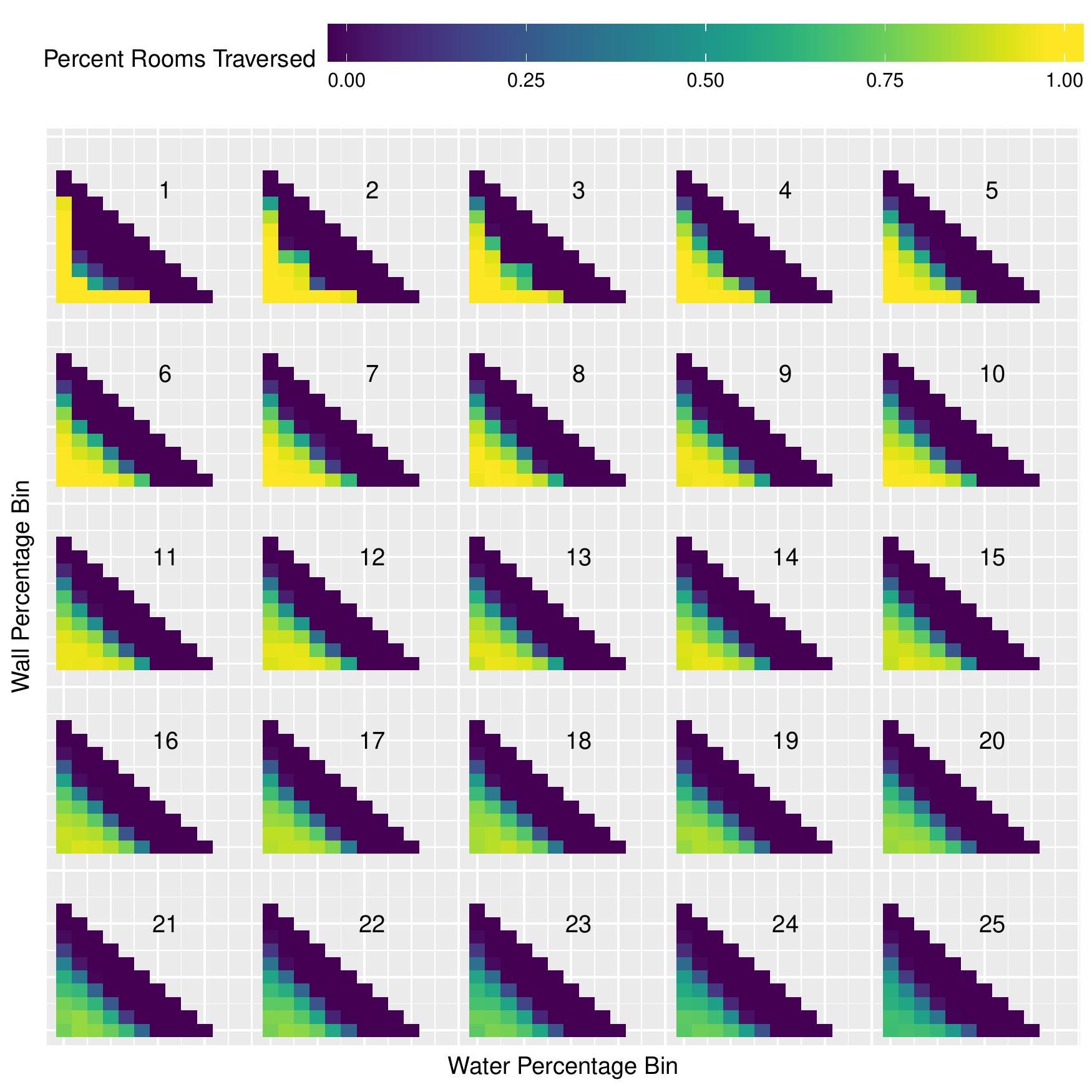}
    \caption{CPPNThenDirect2GAN}
    \label{fig:zeldaWWRheatCPPNThenDirect2GAN}
\end{subfigure}
\begin{subfigure}{0.49\textwidth} 
    \includegraphics[width=1.0\textwidth]{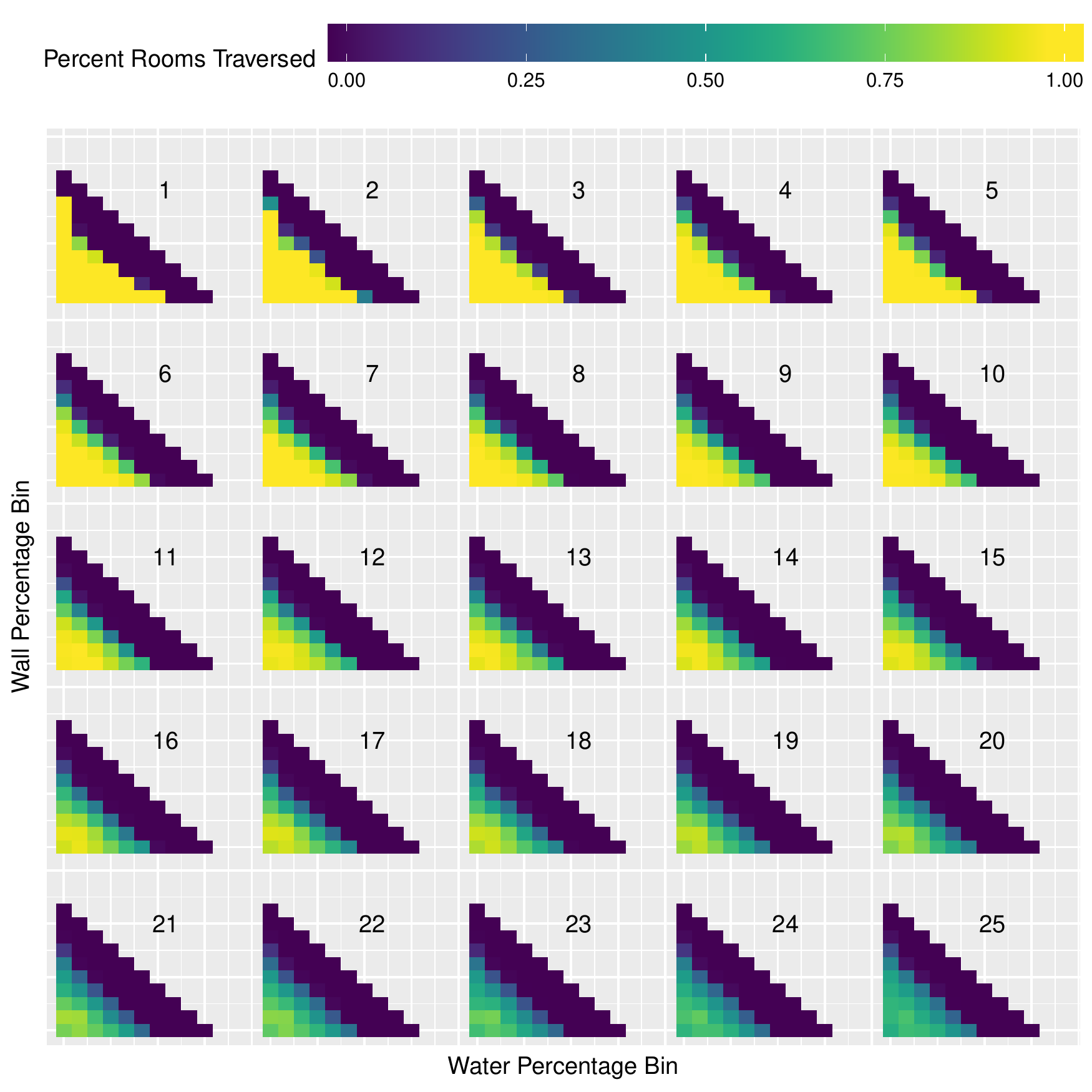}
    \caption{Max of Direct2GAN and CPPN2GAN}
    \label{fig:zeldaWWRheatCombined}
\end{subfigure}

\caption{\textbf{Average MAP-Elites Bin Fitness Across 30 Runs of Evolution in Zelda Using \textbf{WWR}.} The three methods are compared the same way as in Mario. Each distinct triangular grid corresponds to levels with a particular number of reachable rooms (top-right of the grid). Grids are triangular because there is a trade-off between water percentage and wall percentage. It is impossible for this sum to exceed 100\%, and likely for it to be much less because some percentage of each room must be dedicated to empty floor tiles. (\subref{fig:zeldaWWRheatDirect2GAN}) Direct2GAN has trouble filling bins with just one reachable room, and generally has trouble discovering levels in bins near the upper-right edge of each triangle, especially as the number of reachable rooms increases. (\subref{fig:zeldaWWRheatCPPN2GAN}) CPPN2GAN and (\subref{fig:zeldaWWRheatCPPNThenDirect2GAN}) CPPNThenDirect2GAN are both better in this regard, though even for these methods it becomes harder to reach all rooms as the number of reachable rooms increases (bin brightness fades). (\subref{fig:zeldaWWRheatCombined}) The combined archive is closer to CPPN2GAN than Direct2GAN, and has a little bit more coverage than CPPNThenDirect2GAN.}
\label{fig:zeldaWWRaverageHeat}
\end{figure*}

\begin{figure*}[tp]
\centering
\begin{subfigure}{0.49\textwidth} 
    \includegraphics[width=1.0\textwidth]{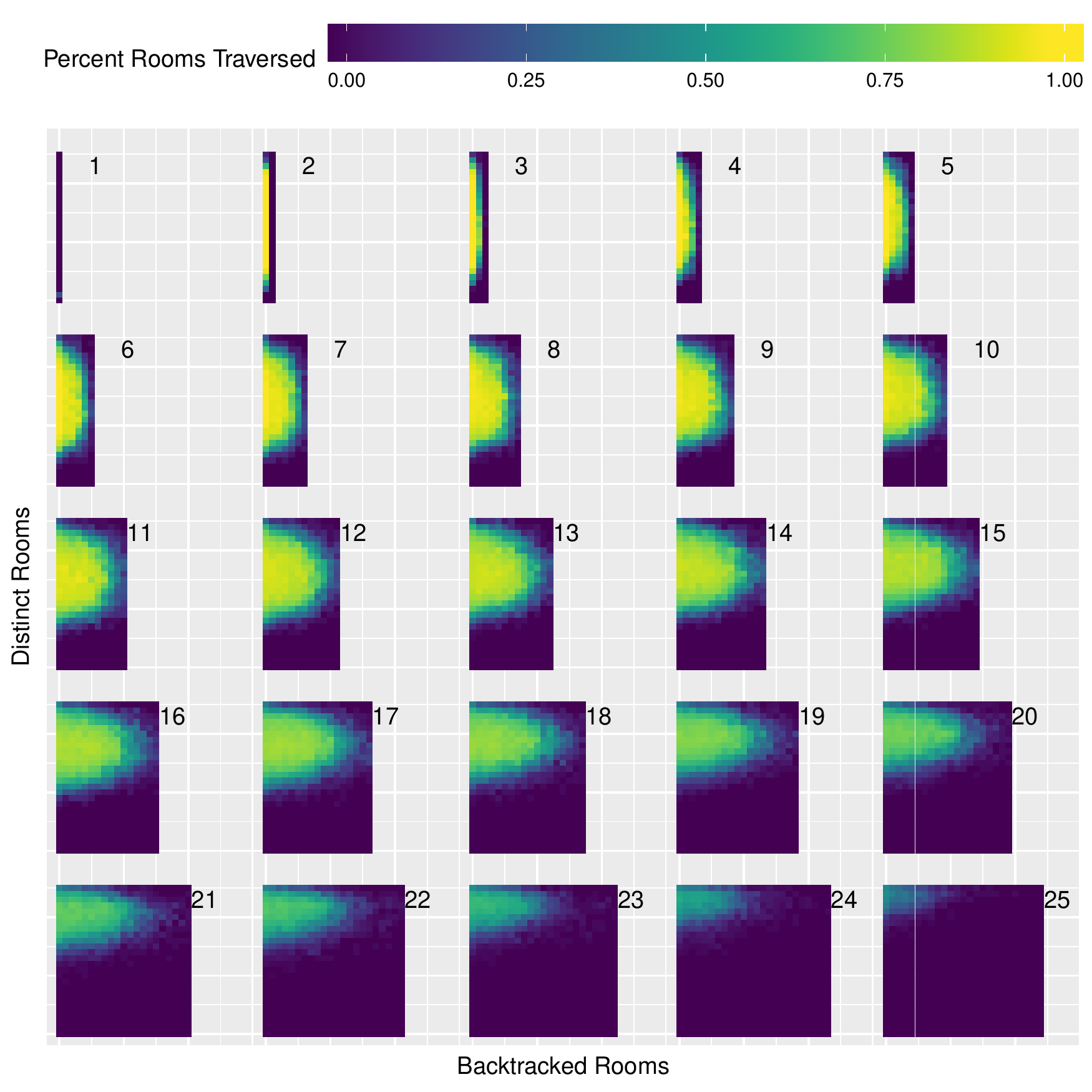}
    \caption{Direct2GAN}
    \label{fig:zeldaDBRheatDirect2GAN}
\end{subfigure}
\begin{subfigure}{0.49\textwidth} 
    \includegraphics[width=1.0\textwidth]{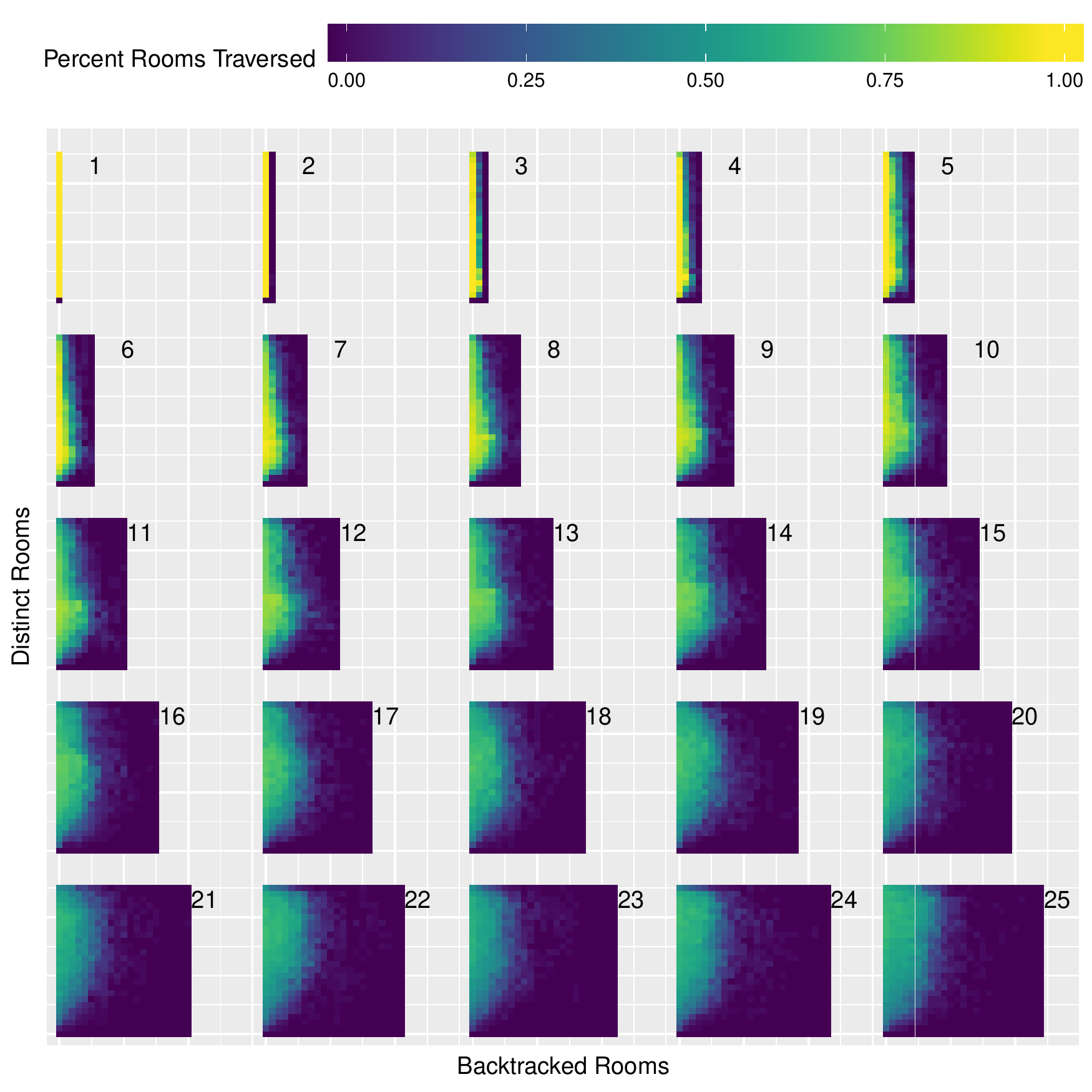}
    \caption{CPPN2GAN}
    \label{fig:zeldaDBRheatCPPN2GAN}
\end{subfigure} \\

\begin{subfigure}{0.49\textwidth} 
    \includegraphics[width=1.0\textwidth]{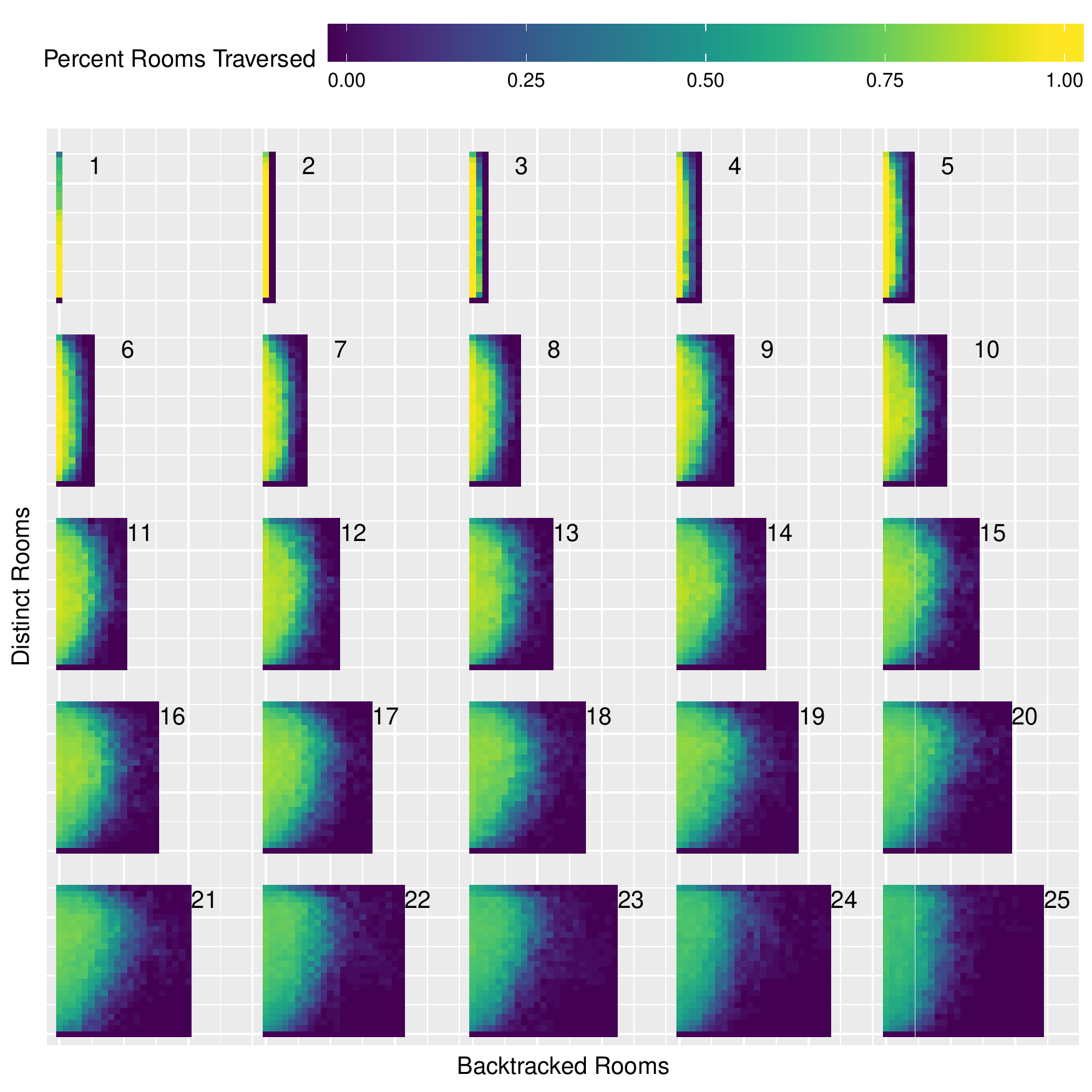}
    \caption{CPPNThenDirect2GAN}
    \label{fig:zeldaDBRheatCPPNThenDirect2GAN}
\end{subfigure}
\begin{subfigure}{0.49\textwidth} 
    \includegraphics[width=1.0\textwidth]{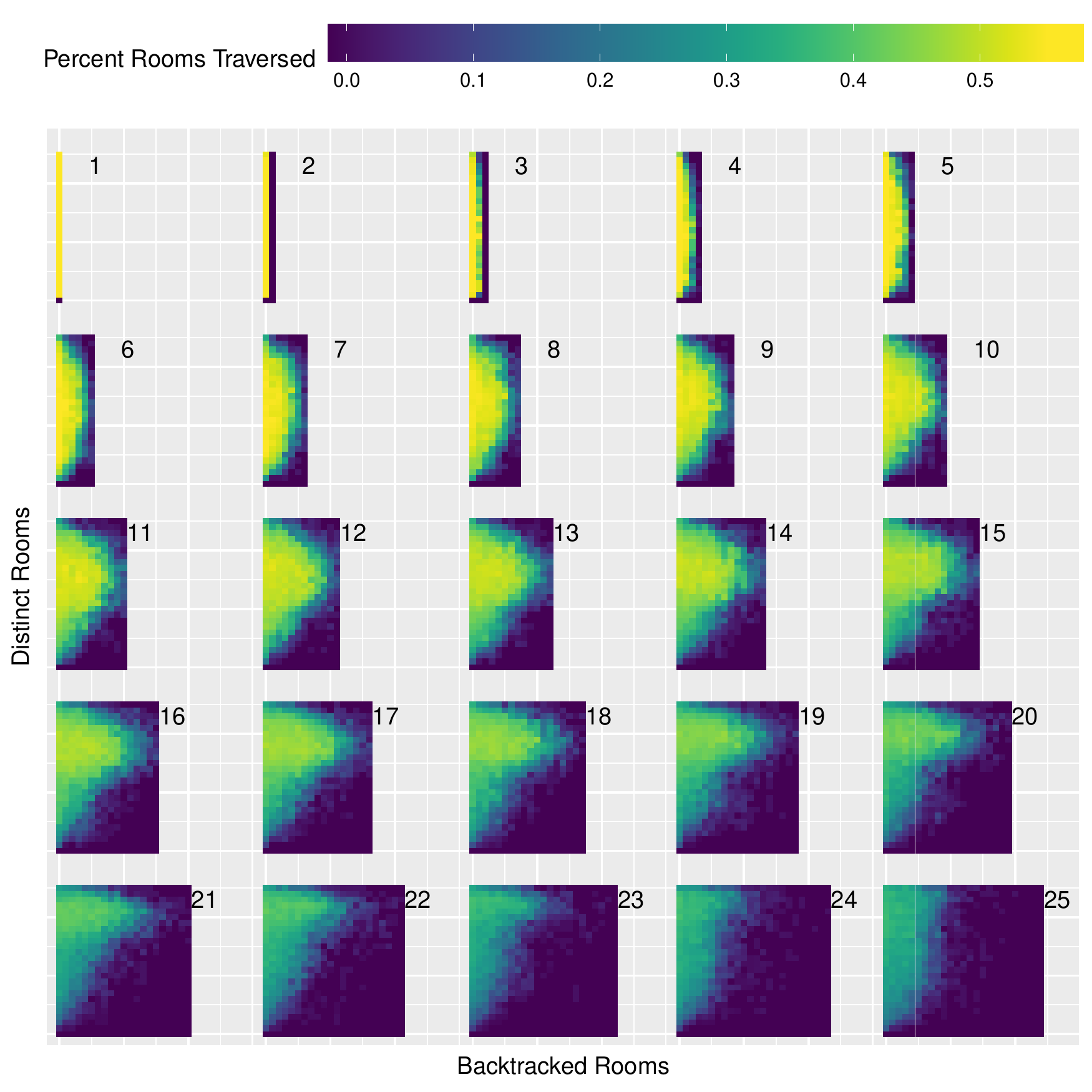}
    \caption{Max of Direct2GAN and CPPN2GAN}
    \label{fig:zeldaDBRheatCombined}
\end{subfigure}

\caption{\textbf{Average MAP-Elites Bin Fitness Across 30 Runs of Evolution in Zelda Using \textbf{Distinct BTR}.} Methods are compared as in Fig.~\ref{fig:zeldaWWRaverageHeat}, but with \textbf{Distinct BTR}. The number beside each sub-grid still represents the number of reachable rooms. Backtracking increases to the right and distinct rooms increases moving up. When there are fewer reachable rooms, the number of rooms through which one can backtrack is reduced, hence the changing width of each sub-grid. (\subref{fig:zeldaDBRheatDirect2GAN}) Direct2GAN has trouble creating levels with a low percentage of distinct rooms, especially as the number of rooms increases. (\subref{fig:zeldaDBRheatCPPN2GAN}) CPPN2GAN has trouble creating lots of backtracking. (\subref{fig:zeldaDBRheatCPPNThenDirect2GAN}) CPPNThenDirect2GAN is better than CPPN2GAN at backtracking for all numbers of distinct rooms, though Direct2GAN is still best at backtracking for dungeons with many distinct rooms. (\subref{fig:zeldaDBRheatCombined}) Combining Direct2GAN and CPPN2GAN fills bins in the upper-right of some sub-grids that CPPNThenDirect2GAN does not reach, but CPPNThenDirect2GAN generally fills out more bins in the lower-right of sub-grids in the bottom two rows.}
\label{fig:zeldaDBRaverageHeat}
\end{figure*}

\begin{figure*}[h!]
\centering
\begin{subfigure}{0.49\textwidth} 
    \includegraphics[width=1.0\textwidth]{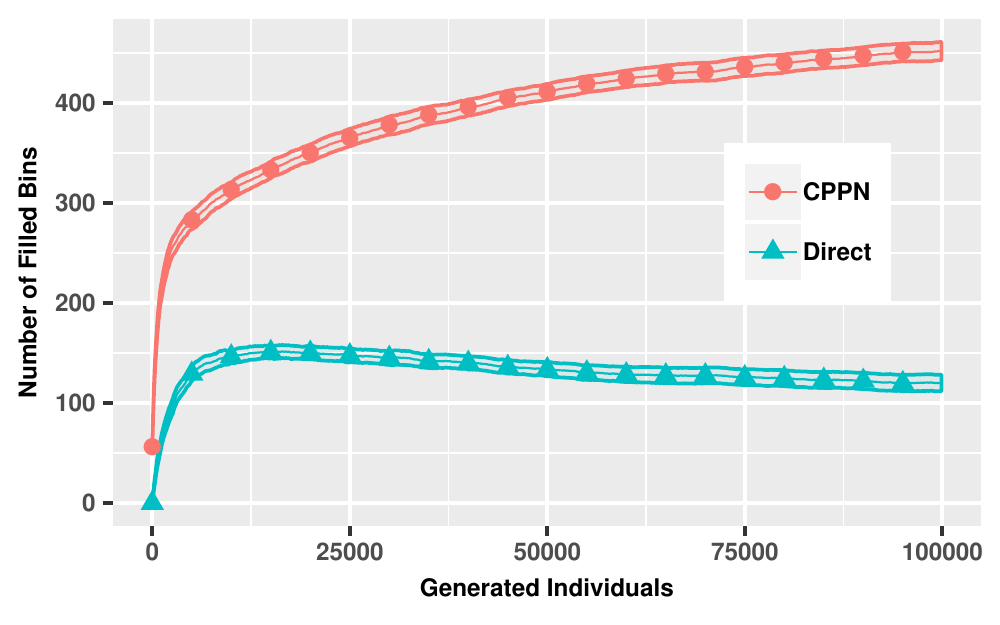}
    \caption{Mario: \textbf{Sum DSL}}
    \label{fig:marioDNSdirectvscppn}
\end{subfigure}
\begin{subfigure}{0.49\textwidth} 
    \includegraphics[width=1.0\textwidth]{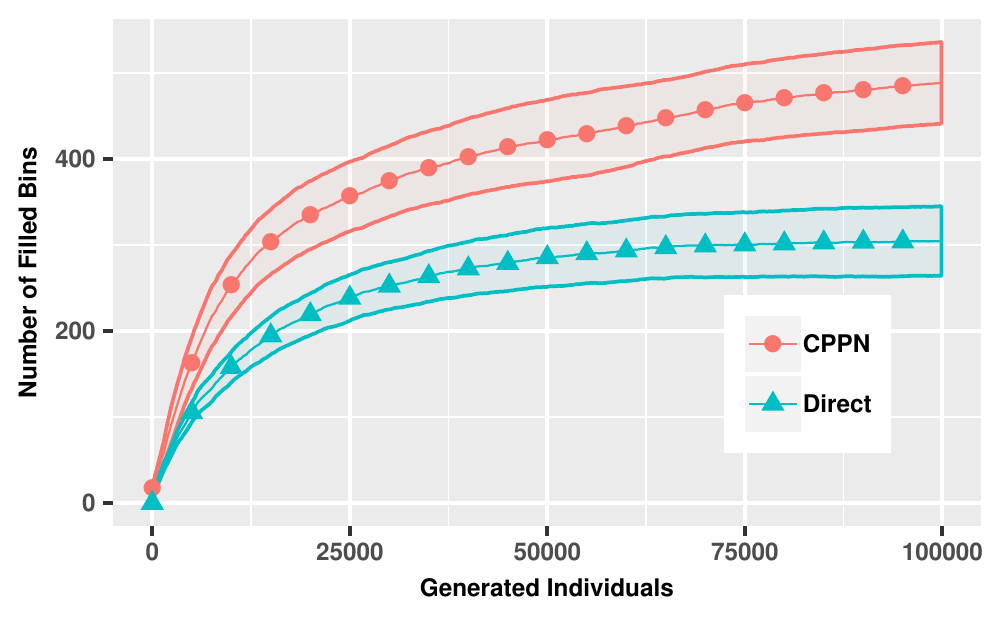}
    \caption{Mario: \textbf{Distinct ASAD}}
    \label{fig:marioDNDdirectvscppn}
\end{subfigure} \\
\begin{subfigure}{0.49\textwidth} 
    \includegraphics[width=1.0\textwidth]{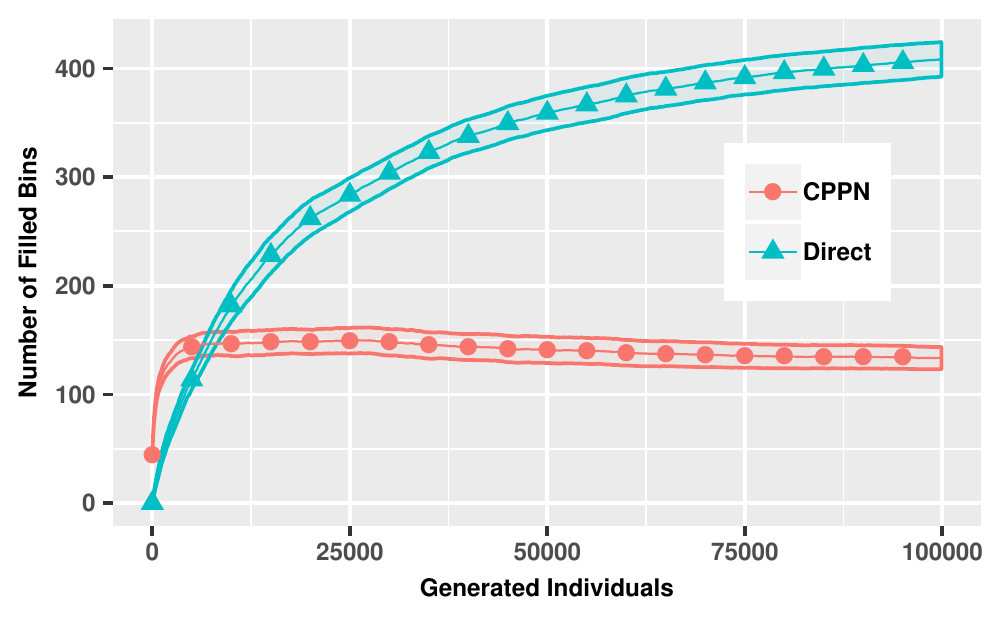}
    \caption{Zelda: \textbf{WWR}}
    \label{fig:zeldaWWRfilldirectvscppn}
\end{subfigure}
\begin{subfigure}{0.49\textwidth} 
    \includegraphics[width=1.0\textwidth]{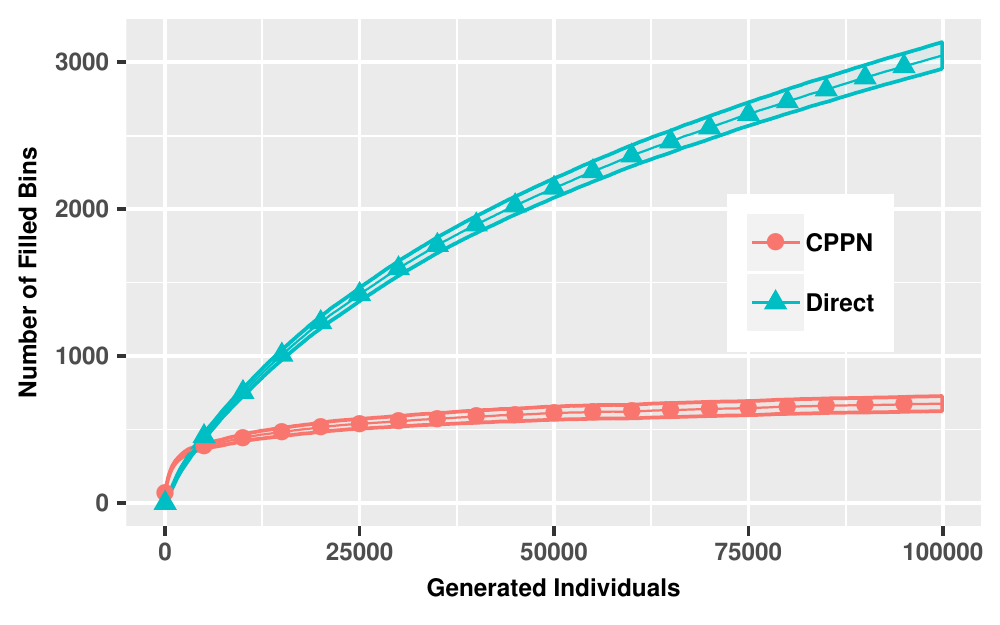}
    \caption{Zelda: \textbf{Distinct BTR}}
    \label{fig:zeldaDBRfilldirectvscppn}
\end{subfigure}

\caption{\textbf{Average CPPN and Direct Elites Across 30 Runs of CPPNThenDirect2GAN.} \normalfont For the CPPNThenDirect2GAN runs of each domain, the average count of CPPN and direct vector genotypes in the archive across evolution are shown with 95\% confidence intervals. (\subref{fig:marioDNSdirectvscppn}) The Mario \textbf{Sum DSL} scheme succeeds primarily with CPPN genotypes, explaining the equal performance with CPPN2GAN. (\subref{fig:marioDNDdirectvscppn}) Mario \textbf{Distinct ASAD} also relies primarily on CPPN genotypes, but there is more variation between runs, hence the larger confidence intervals. (\subref{fig:zeldaWWRfilldirectvscppn}) In Zelda \textbf{WWR}, CPPNs are slightly more prevalent very early in evolution, but the archive is quickly filled with direct vectors, albeit with a reasonably steady portion of CPPNs. (\subref{fig:zeldaDBRfilldirectvscppn}) Zelda \textbf{Distinct BTR} fills with direct vectors even quicker. Overall, Mario binning schemes settle more on CPPN genotypes, but Zelda binning schemes fill more with direct vectors. However, note that these direct vectors are still derived from CPPNs and thus the levels can still contain many of the benefits of CPPN generated levels.}
\label{fig:averageDirectVsCPPN}
\end{figure*}

\begin{figure*}[th!]
\centering
\begin{subfigure}{0.9\textwidth} 
    \includegraphics[width=1.0\textwidth]{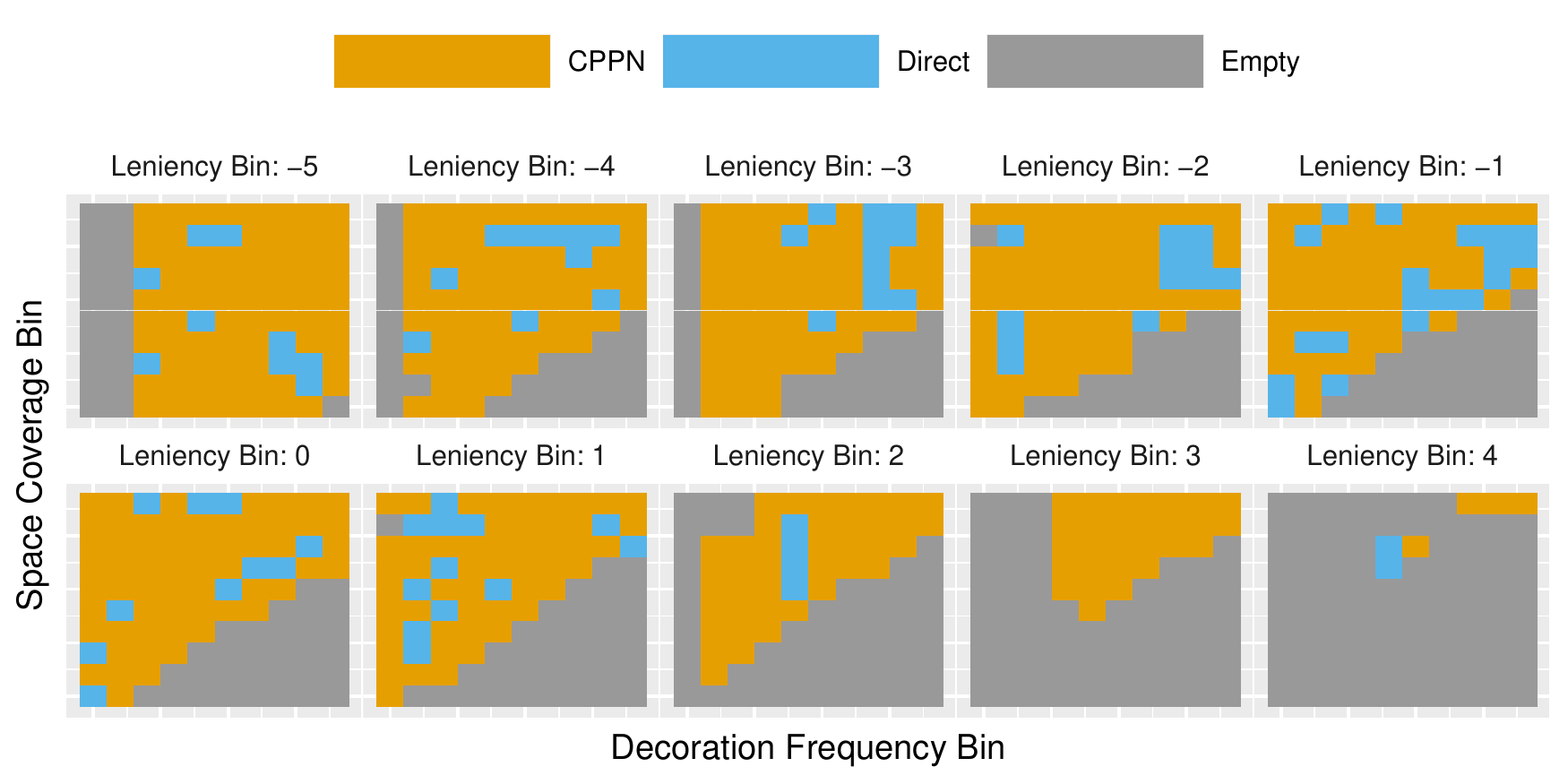}
    \caption{Mario \textbf{Sum DSL}}
    \label{fig:marioDumDSLCPPNVsDirect}
\end{subfigure} \\
\begin{subfigure}{0.49\textwidth} 
    \includegraphics[width=1.0\textwidth]{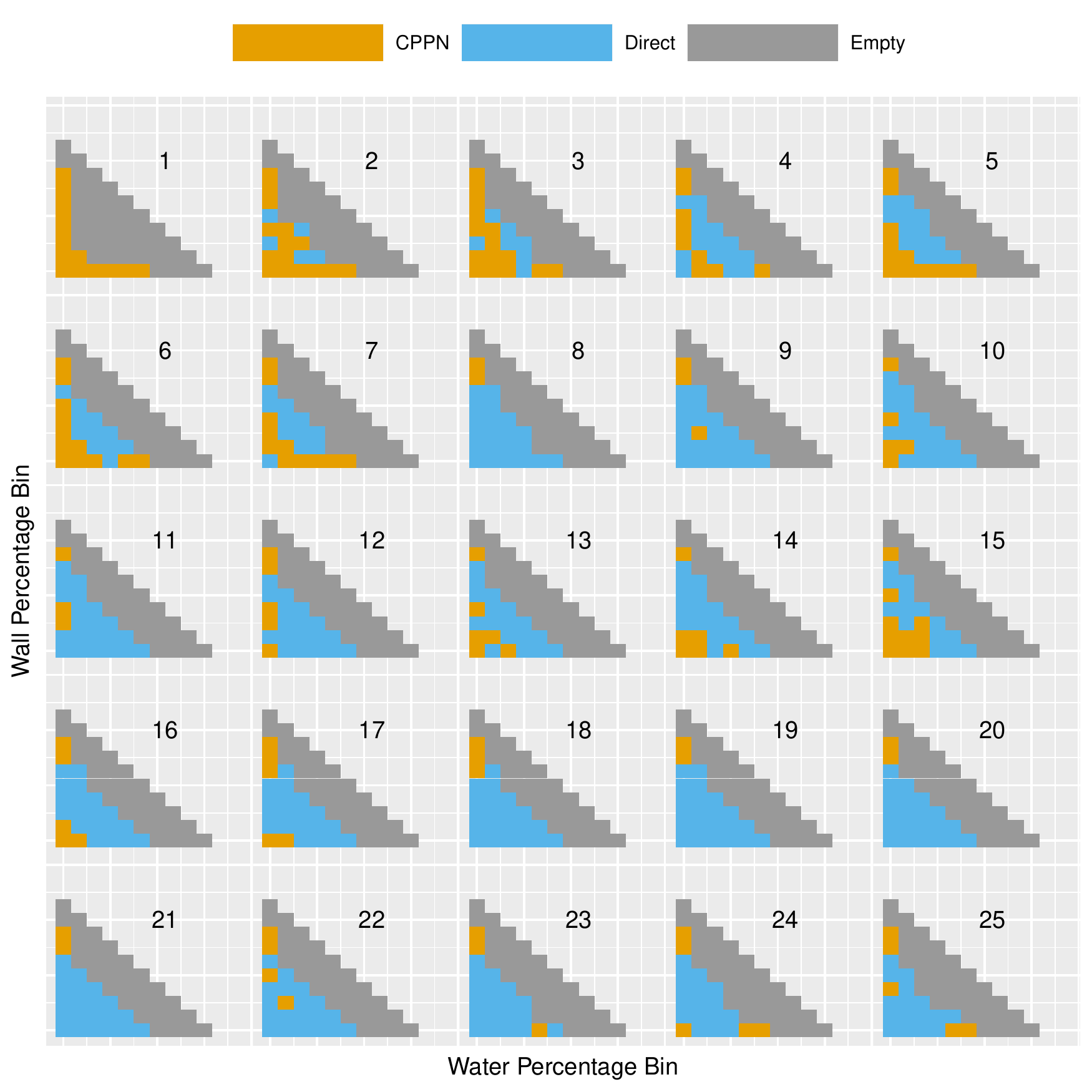}
    \caption{Zelda \textbf{WWR}}
    \label{fig:zeldaWWRCPPNVsDirect}
\end{subfigure}
\begin{subfigure}{0.49\textwidth} 
    \includegraphics[width=1.0\textwidth]{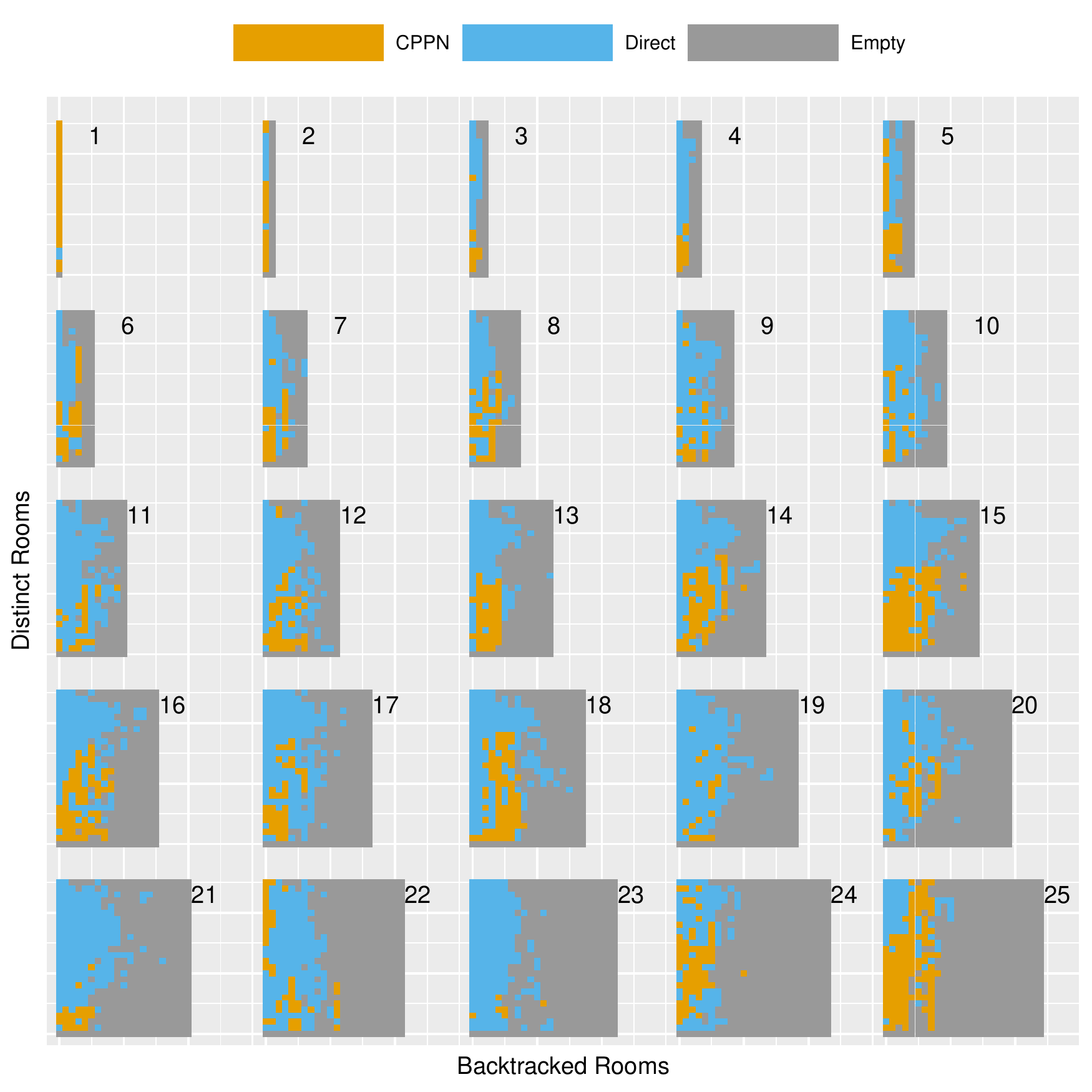}
    \caption{Zelda \textbf{Distinct BTR}}
    \label{fig:zeldaDBTRCPPNVsDirect}
\end{subfigure}

\caption{\textbf{Example Final Distributions of CPPN and Direct Vector Genotypes for Consistent Approaches}. \normalfont For the binning schemes shown, there are only slight variations in the final distribution of CPPN and direct vector genotypes across the archives of different runs. Therefore, an individual example is shown for each scheme. (\subref{fig:marioDumDSLCPPNVsDirect}) Mario \textbf{Sum DSL} consistently fills the archive with CPPN genotypes, and only a few direct vector genotypes are sprinkled throughout the archive in seemingly random places. (\subref{fig:zeldaWWRCPPNVsDirect}) In contrast, Zelda \textbf{WWR} consistently fills archives with direct vectors. CPPNs tend to cluster in bins with fewer reachable rooms, though CPPNs are sprinkled elsewhere throughout the archive too. (\subref{fig:zeldaDBTRCPPNVsDirect}) Zelda \textbf{Distinct BTR} archives also fill up with direct vectors, but CPPNs tend to appear in bins with fewer distinct rooms.}
\label{fig:consistentCPPNVsDirect}
\end{figure*}

\begin{figure*}[th!]
\centering
\begin{subfigure}{0.49\textwidth} 
    \includegraphics[width=1.0\textwidth]{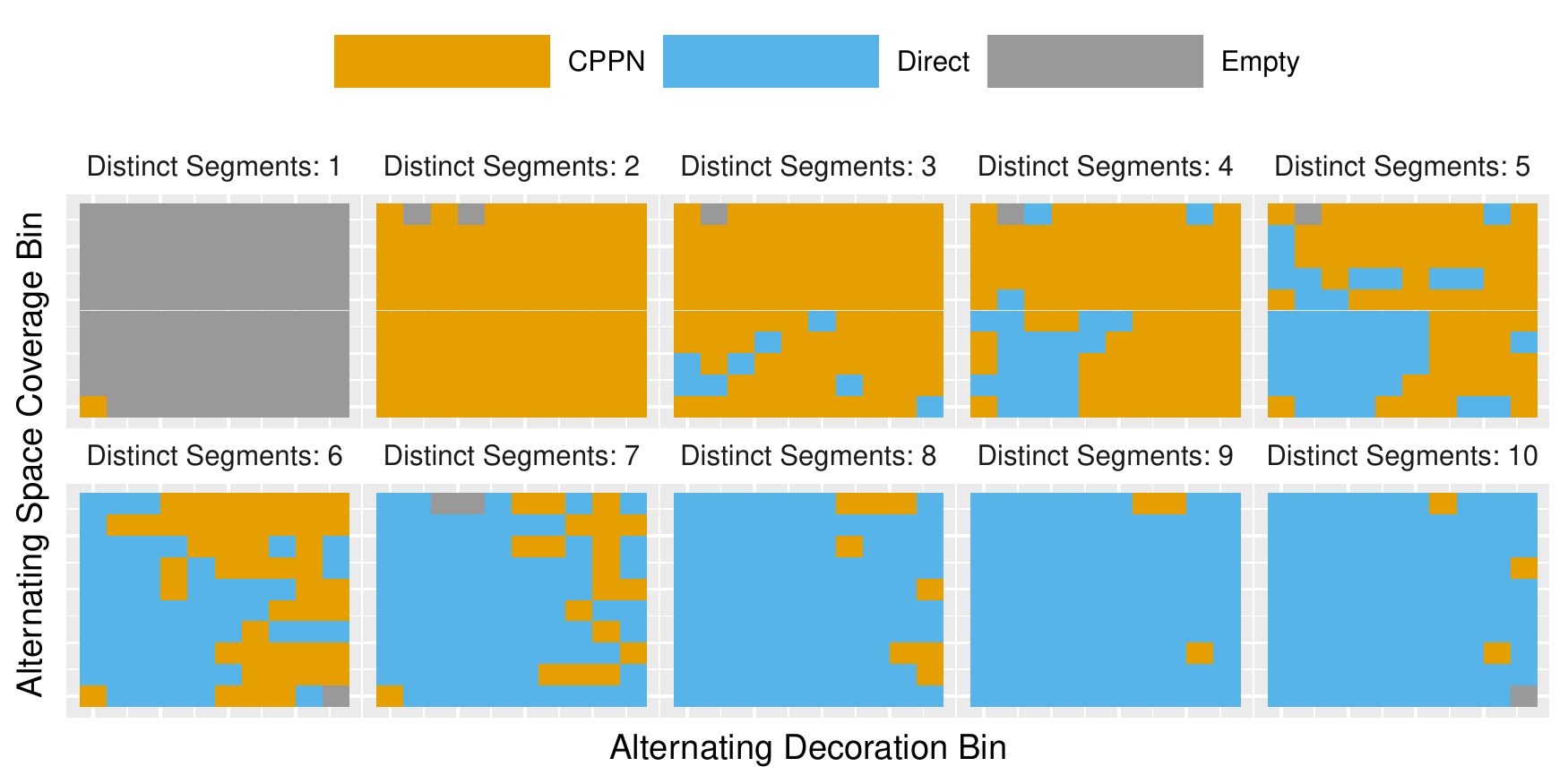}
    \caption{More Direct Vectors Than CPPNs}
    \label{fig:moreDirect}
\end{subfigure} 
\begin{subfigure}{0.49\textwidth} 
    \includegraphics[width=1.0\textwidth]{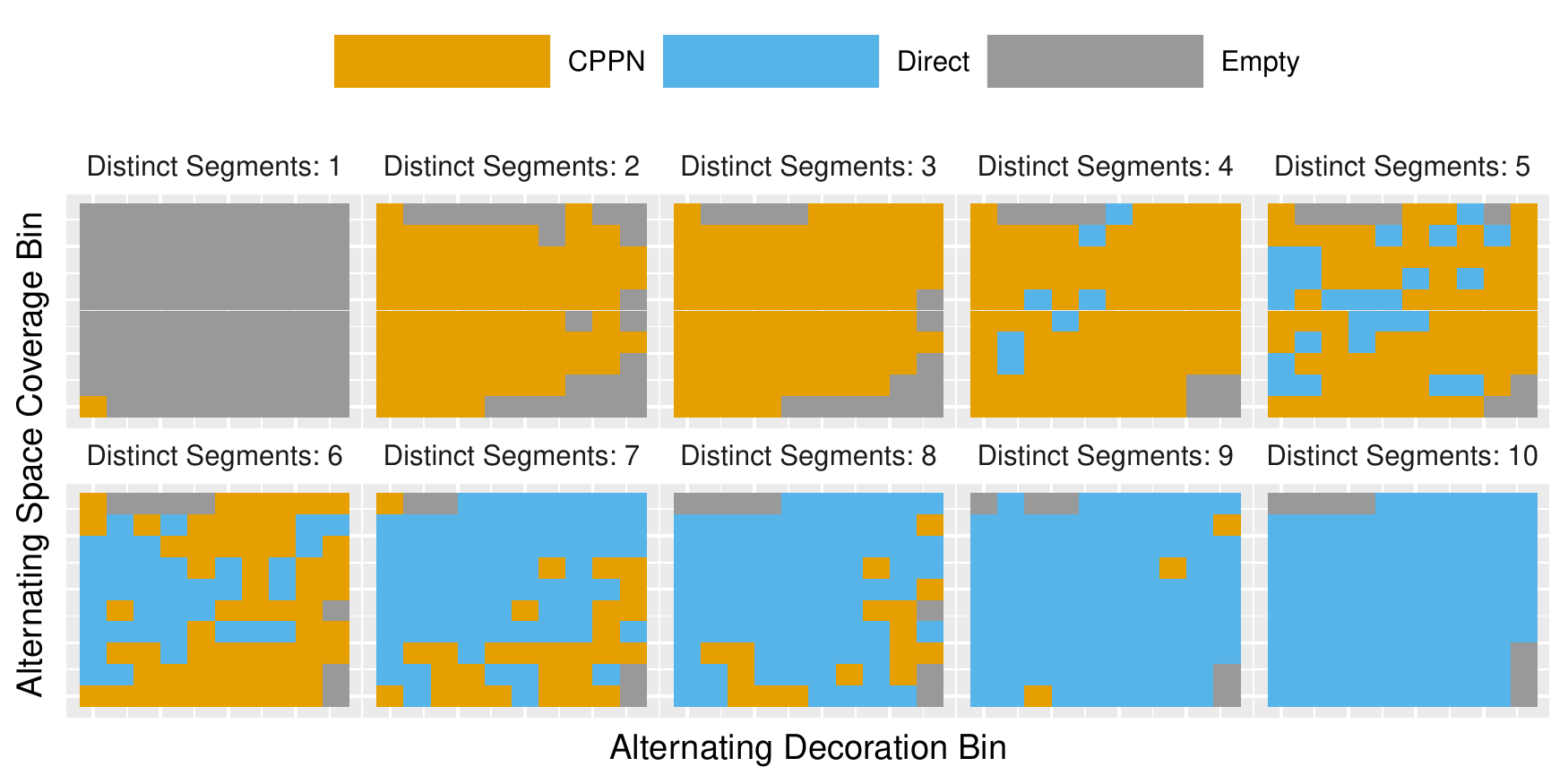}
    \caption{Roughly Even Split of CPPNs and Direct Vectors}
    \label{fig:evenSplit}
\end{subfigure} \\
\begin{subfigure}{0.49\textwidth} 
    \includegraphics[width=1.0\textwidth]{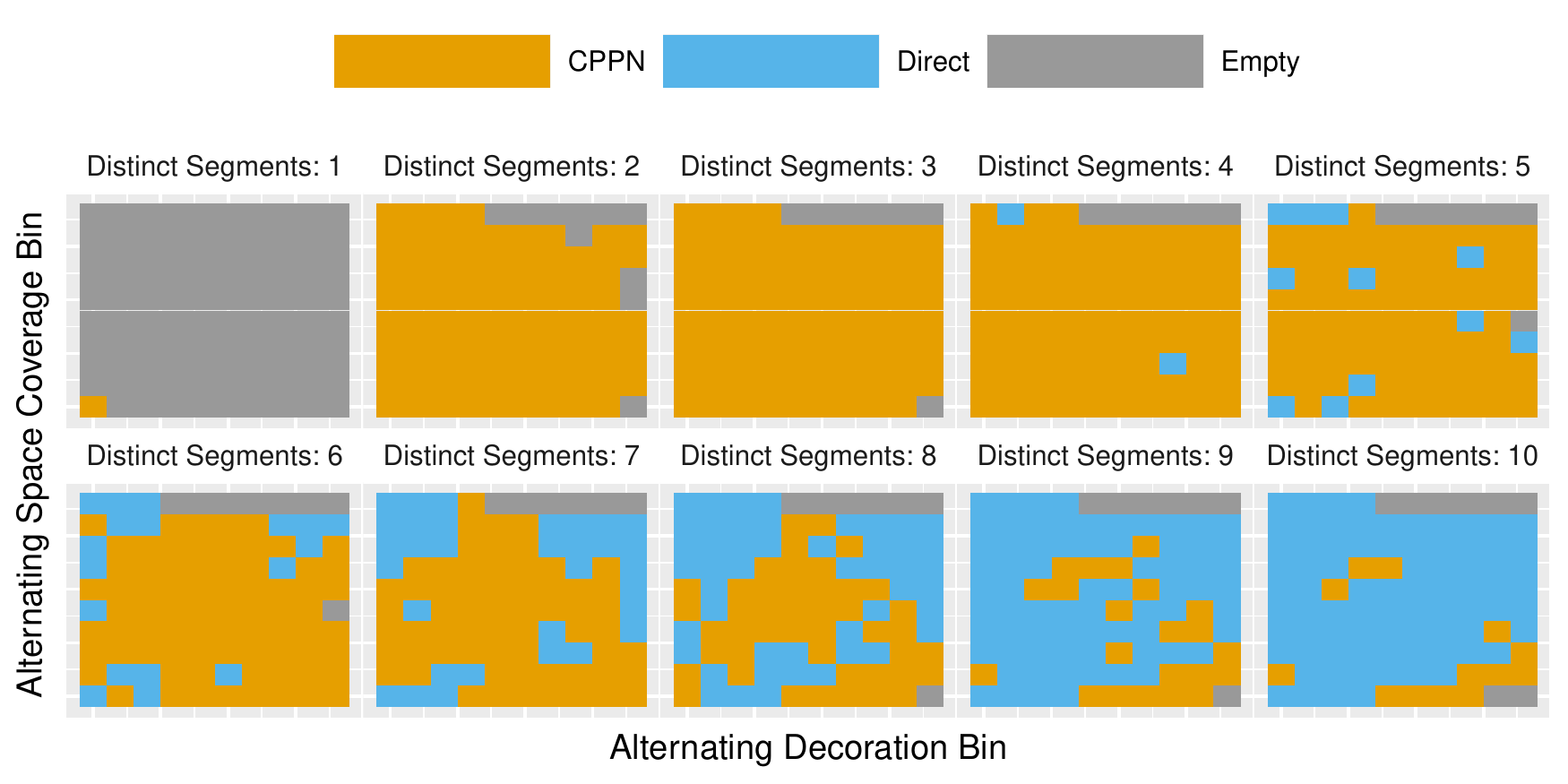}
    \caption{More CPPNs Than Direct Vectors}
    \label{fig:moreCPPN}
\end{subfigure} 
\begin{subfigure}{0.49\textwidth} 
    \includegraphics[width=1.0\textwidth]{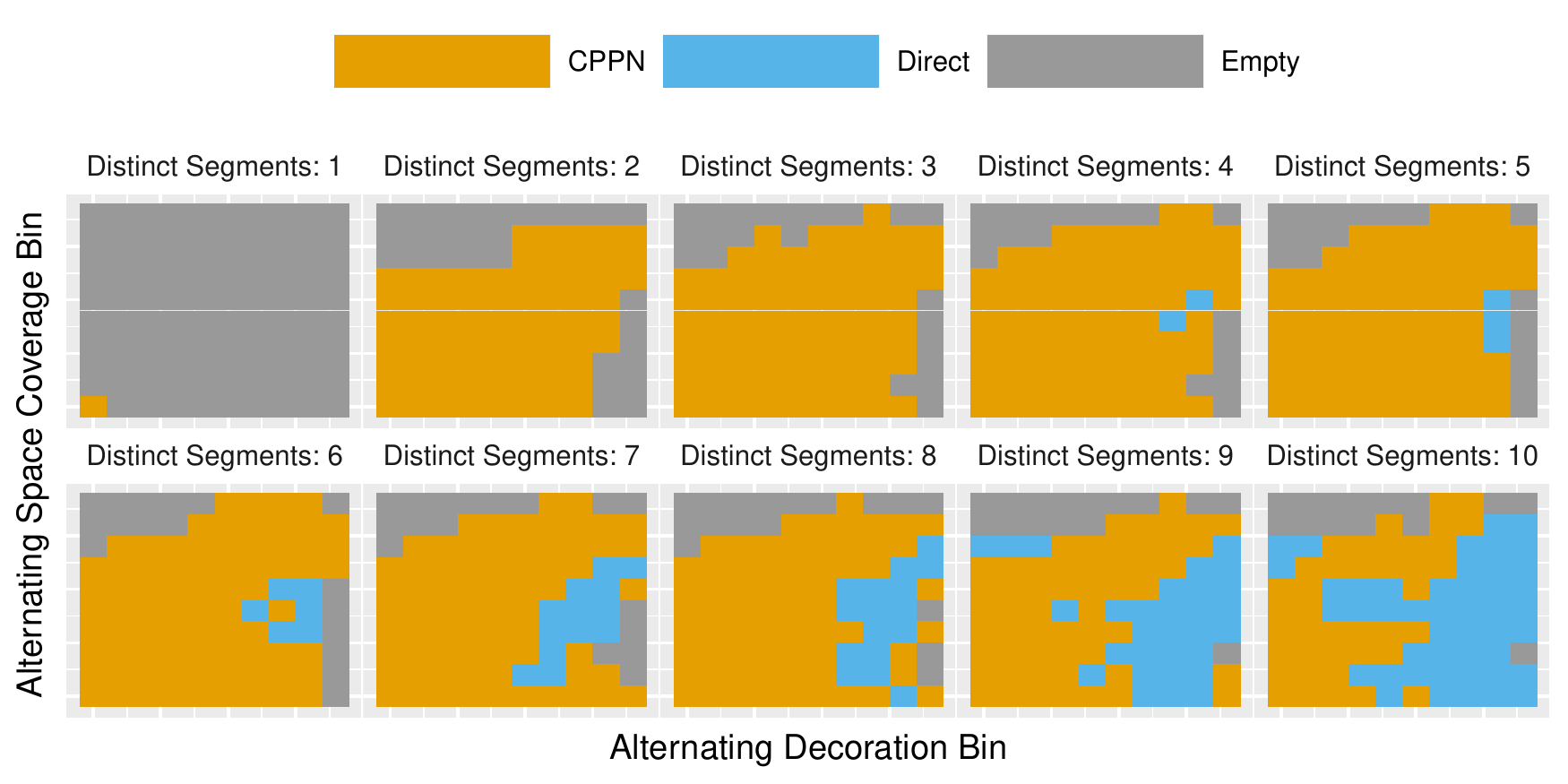}
    \caption{Mostly CPPNs}
    \label{fig:lotsCPPN}
\end{subfigure}

\caption{\textbf{Final Distributions of CPPN and Direct Vector Genotypes for Different Mario Distinct ASAD Runs}. \normalfont Mario \textbf{Distinct ASAD} runs show lots of variation. (\subref{fig:moreDirect}) This run ended with the largest number of direct vectors in the archive. (\subref{fig:evenSplit}) This run and several others had a roughly even split between CPPNs and direct vectors. (\subref{fig:moreCPPN}) The majority of runs have more CPPNs than direct vectors, like this run. (\subref{fig:lotsCPPN}) This run has the largest number of CPPNs in the archive at the end of evolution.}
\label{fig:marioASADCPPNVsDirect}
\end{figure*}

\begin{figure*}[tp]
\centering

\begin{subfigure}{0.49\textwidth} 
    \includegraphics[width=1.0\textwidth]{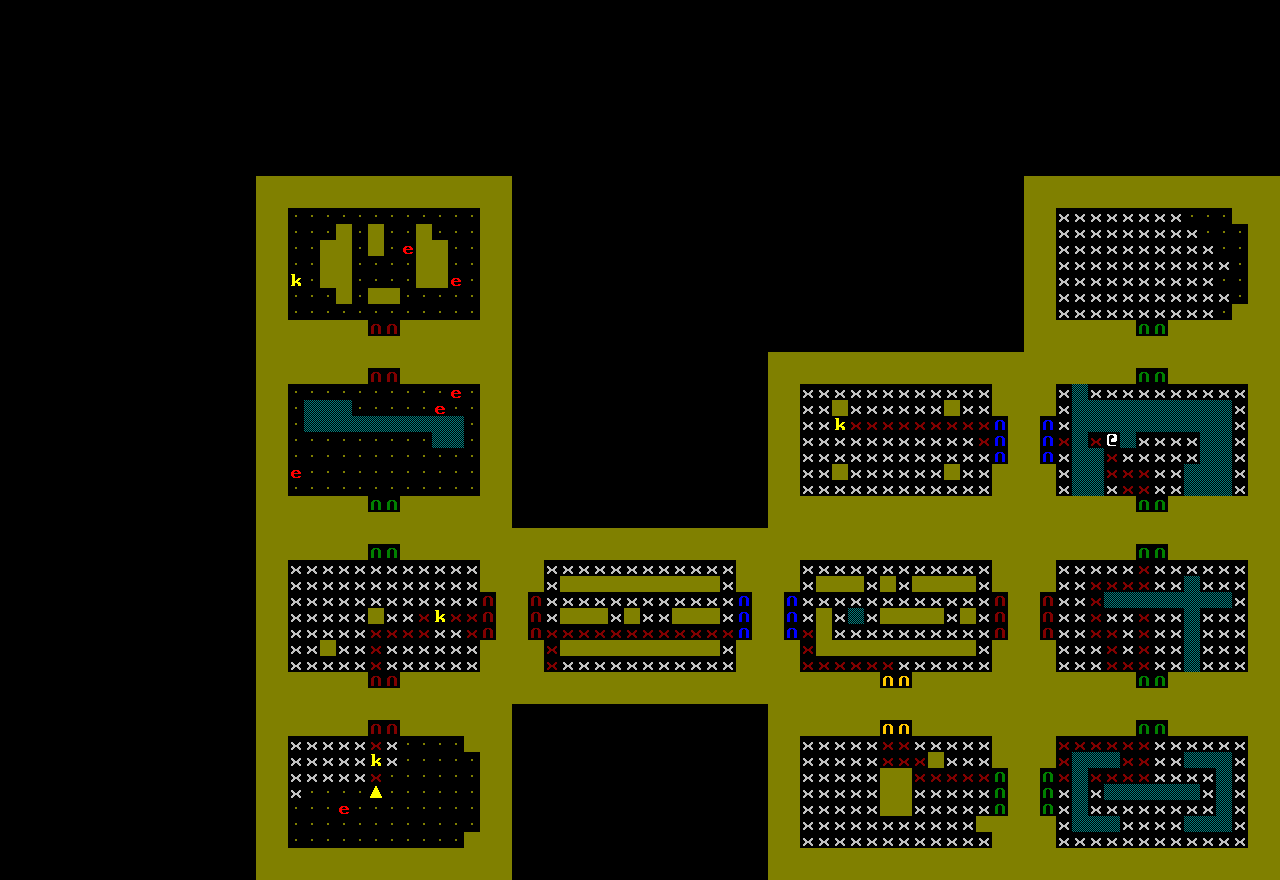}
    \caption{Direct2GAN with \textbf{Distinct BTR}. \hl{Distinct: 11, Backtracked: 11, Reachable:~12.}}
    \label{fig:zeldaDBRDirect}
\end{subfigure}
\begin{subfigure}{0.49\textwidth} 
    \includegraphics[width=1.0\textwidth]{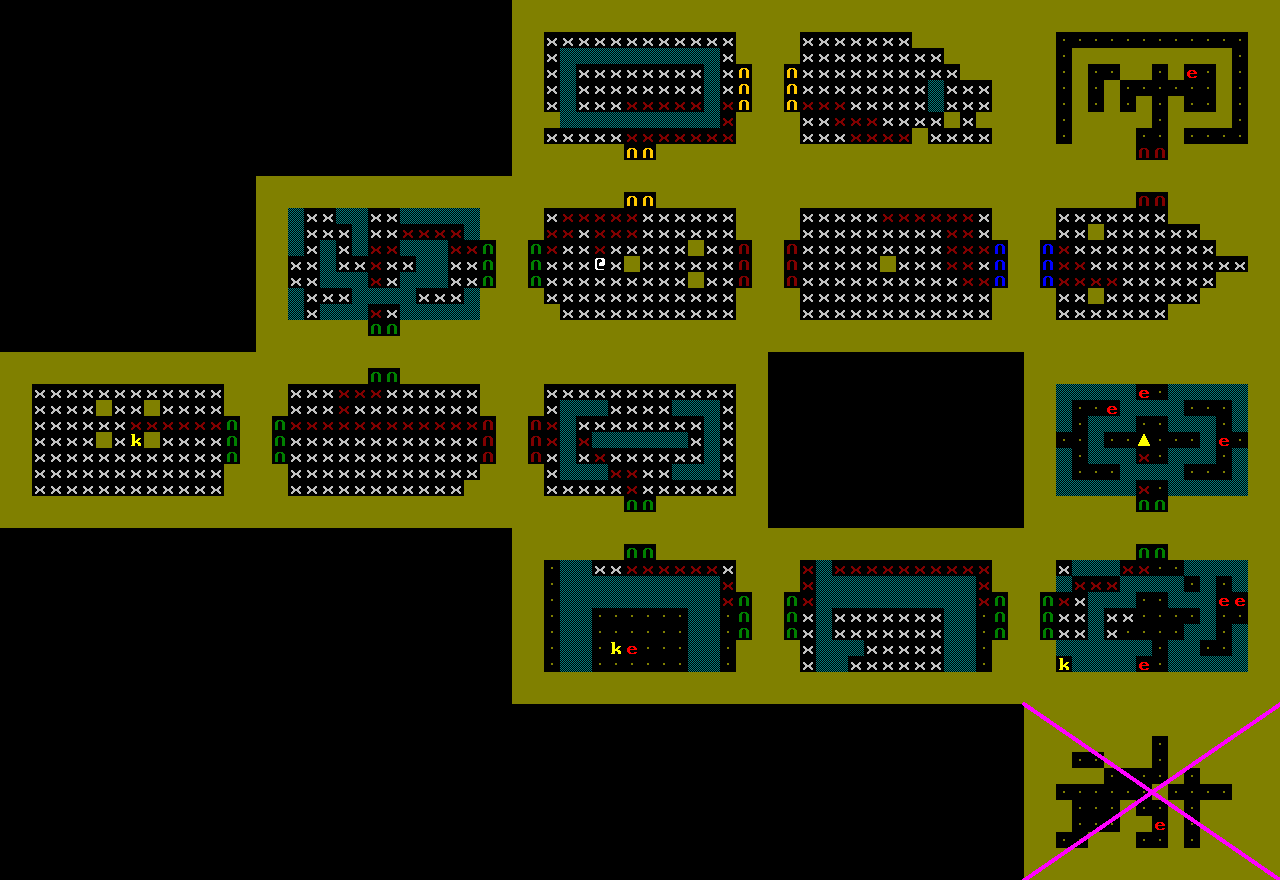}
    \caption{Direct2GAN with \textbf{WWR}. \hl{Wall: 0, Water: 2, Reachable Rooms:~14.}}
    \label{fig:zeldaWWRDirect}
\end{subfigure}\\

\begin{subfigure}{0.49\textwidth} 
    \includegraphics[width=1.0\textwidth]{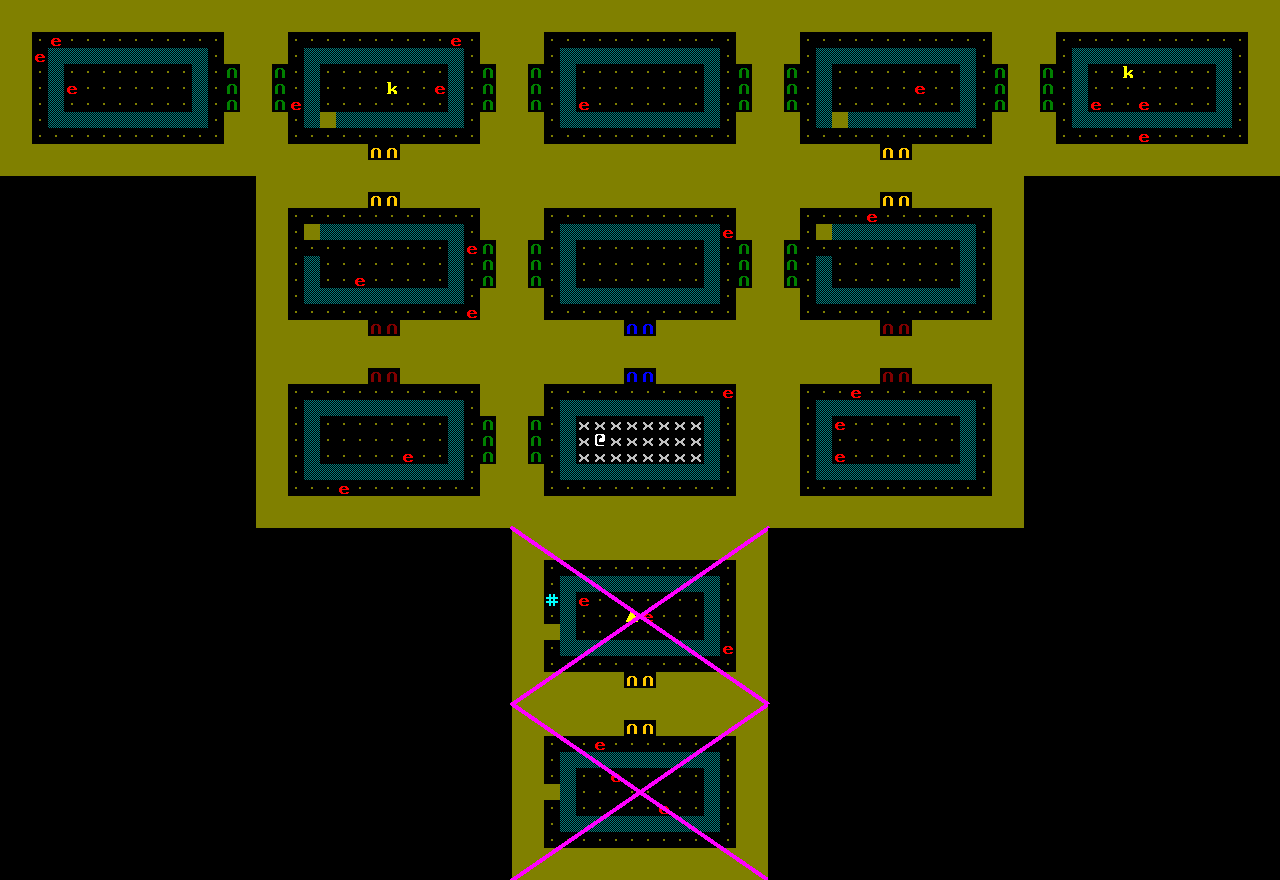}
    \caption{CPPN2GAN with \textbf{Distinct BTR}. \hl{Distinct: 1, Backtracked: 0, Reachable:~11.}}
    \label{fig:zeldaDBRCPPN}
\end{subfigure}
\begin{subfigure}{0.49\textwidth} 
    \includegraphics[width=1.0\textwidth]{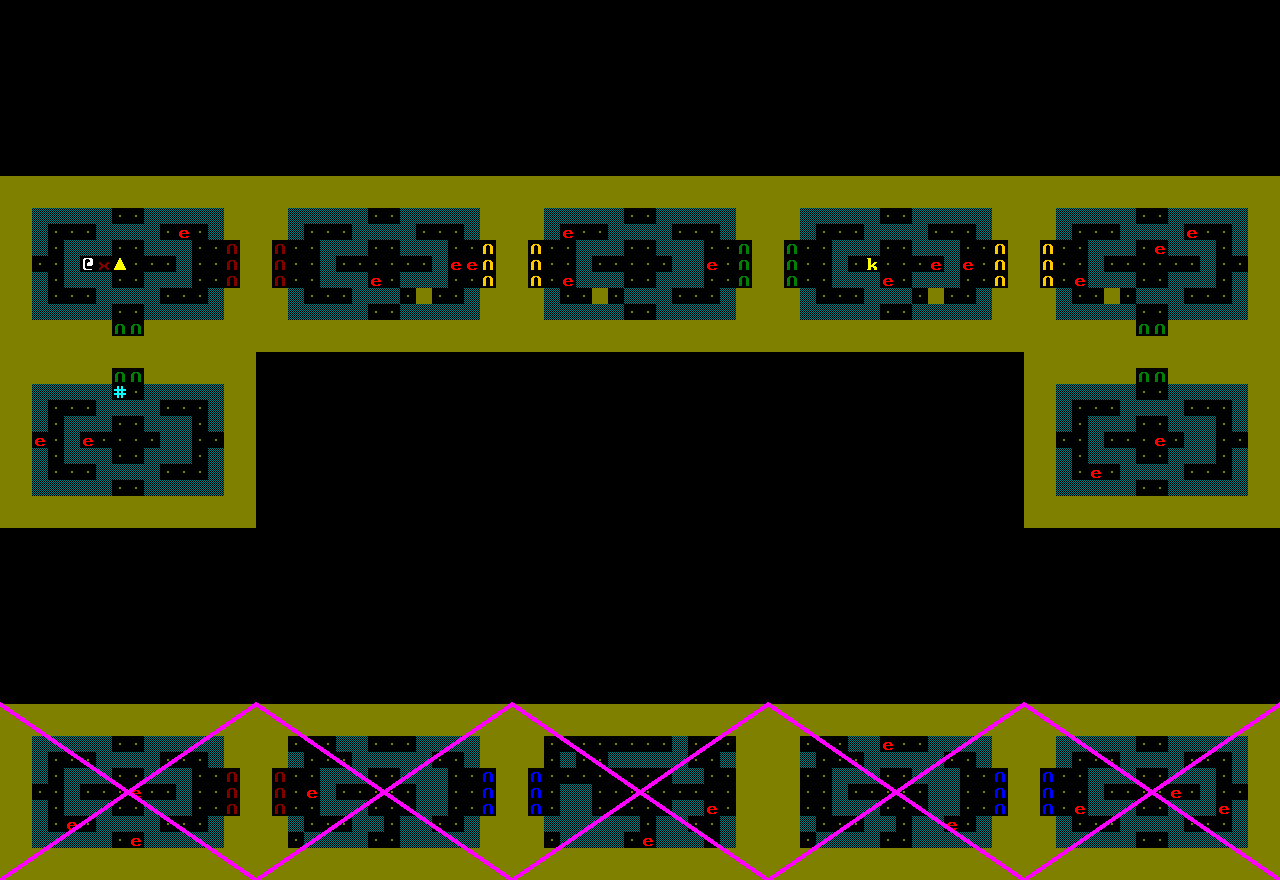}
    \caption{CPPN2GAN with \textbf{WWR}. \hl{Wall: 0, Water: 6, Reachable Rooms:~7.~~~~}} % Extra spacing here helps vertical alignment
    \label{fig:zeldaWWRCPPN}
\end{subfigure} \\

\begin{subfigure}{0.49\textwidth} 
    \includegraphics[width=1.0\textwidth]{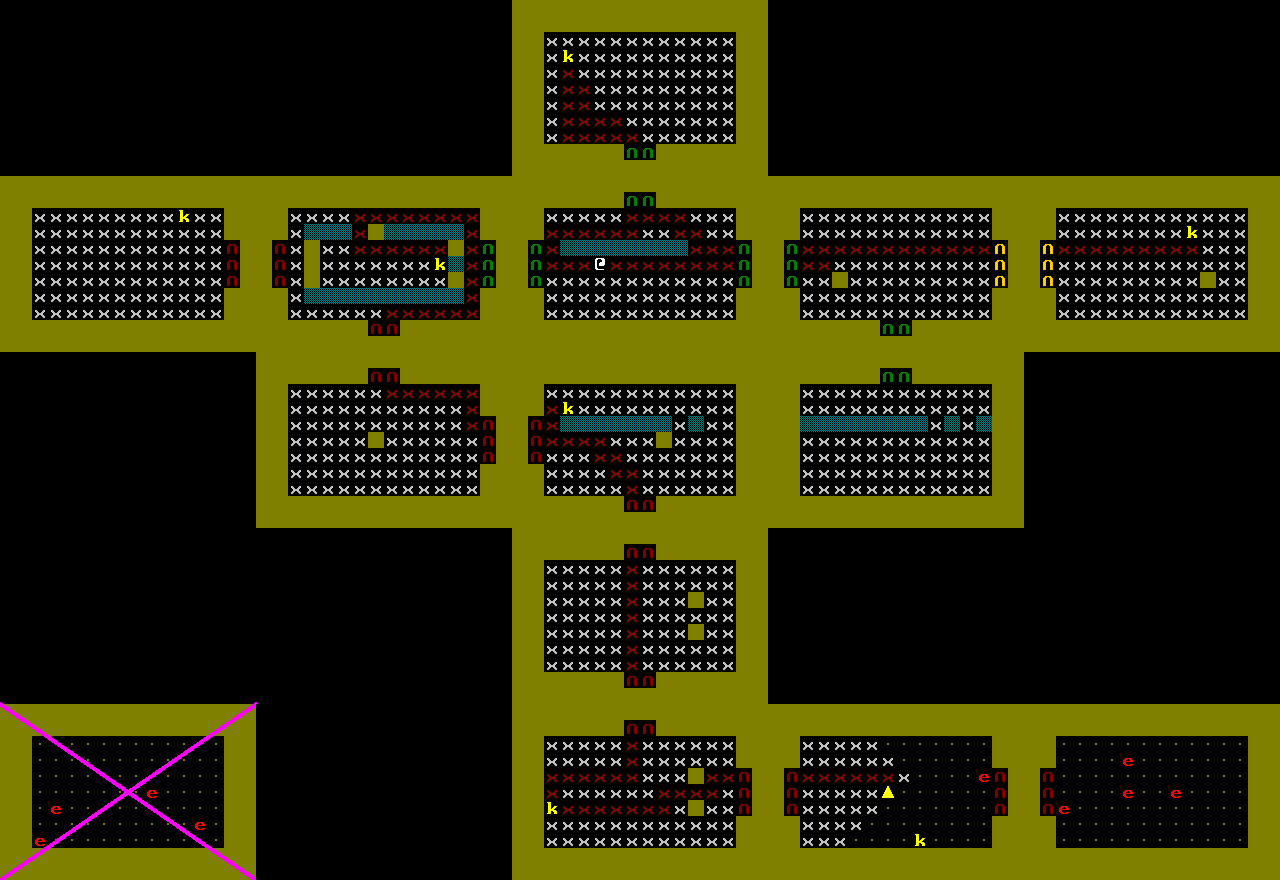}
    \caption{CPPNThenDirect2GAN \hl{(direct)} with \textbf{Distinct BTR}. \hl{Distinct: 7, Backtracked: 13, Reachable:~13.}}
%    \includegraphics[width=1.0\textwidth]{CPPNThenDirectDistinctRooms[6]BackTrackedRooms[10]Rooms[20]-Direct-0.70000-57215.png}
%    \caption{CPPNThenDirect2GAN \hl{(direct)} with \textbf{Distinct BTR}. \hl{Distinct: 6, Backtracked: 10, Reachable:~20.}}
    \label{fig:zeldaDBRCPPNThenDirect}
\end{subfigure} 
\begin{subfigure}{0.49\textwidth} 
    \includegraphics[width=1.0\textwidth]{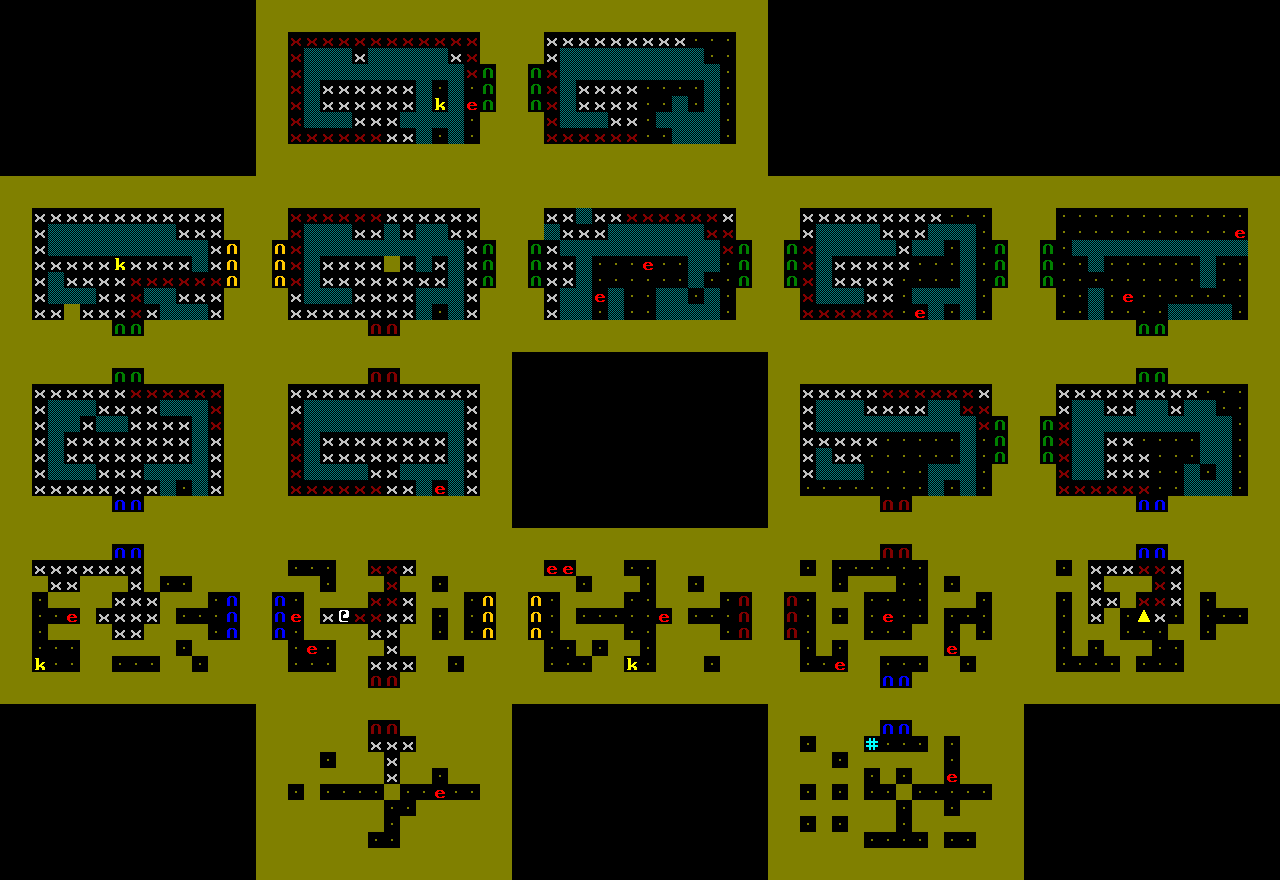}
    \caption{CPPNThenDirect2GAN \hl{(direct)} with \textbf{WWR}. \hl{Wall: 2, Water: 2, Reachable Rooms:~18.}}
    \label{fig:zeldaWWRCPPNThenDirect}
\end{subfigure}  \\

\caption{\textbf{Dungeons From Each Binning Scheme and Encoding for Zelda}.}
\label{fig:zeldaBinLevels}
\end{figure*}

\begin{figure*}[tp]
\centering

\begin{subfigure}{1.0\textwidth} 
    \includegraphics[width=1.0\textwidth]{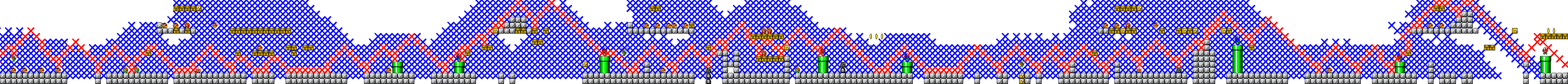}
    \caption{Direct2GAN with \textbf{Distinct ASAD}. \hl{Distinct: 10, Space Coverage: 5, Decoration: 7.}}
    \label{fig:marioDNSDDirect}
\end{subfigure}\\

\begin{subfigure}{1.0\textwidth} 
    \includegraphics[width=1.0\textwidth]{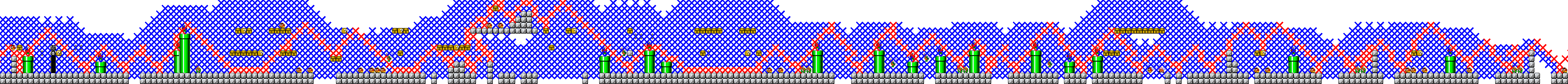}
    \caption{Direct2GAN with \textbf{Sum DSL}. \hl{Decoration: 9, Space Coverage: 9, Leniency: -5.}}
    \label{fig:marioDNSLDirect}
\end{subfigure}\\

\begin{subfigure}{1.0\textwidth} 
    \includegraphics[width=1.0\textwidth]{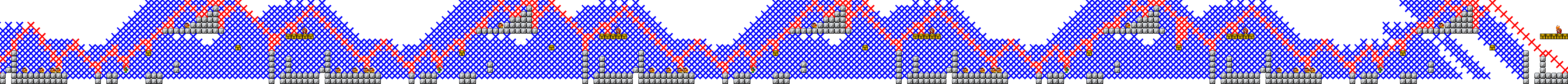}
    \caption{CPPN2GAN with \textbf{Distinct ASAD}. \hl{Distinct: 2, Space Coverage: 6, Decoration: 0.}}
    \label{fig:zmarioDNSDCPPN}
\end{subfigure}\\

\begin{subfigure}{1.0\textwidth} 
    \includegraphics[width=1.0\textwidth]{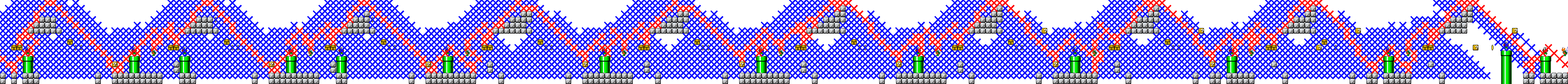}
    \caption{CPPN2GAN with \textbf{Sum DSL}. \hl{Decoration: 5, Space Coverage: 9, Leniency: -1.}}
    \label{fig:marioDSLCPPN}
\end{subfigure} \\

\begin{subfigure}{1.0\textwidth} 
    \includegraphics[width=1.0\textwidth]{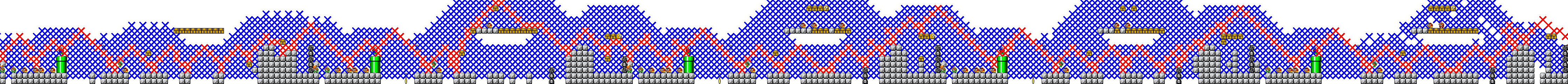}
    \caption{CPPNThenDirect2GAN \hl{(direct)} with \textbf{Distinct ASAD}. \hl{Distinct: 6, Space Coverage: 8, Decoration: 4.}}
    \label{fig:marioDNSDCPPNThenDirect}
\end{subfigure} \\

\begin{subfigure}{1.0\textwidth} 
    \includegraphics[width=1.0\textwidth]{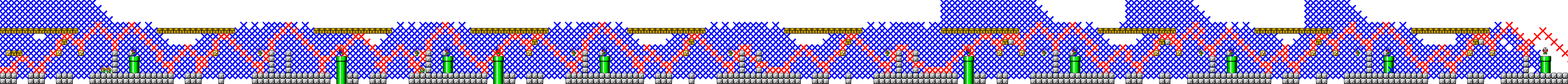}
    \caption{CPPNThenDirect2GAN \hl{(direct)} with \textbf{Sum DSL}. \hl{Decoration: 9, Space Coverage: 9, Leniency: 0.}}
    \label{fig:marioDNSLCPPNThenDirect}
\end{subfigure}  \\

\caption{\textbf{Levels From Each Binning Scheme and Encoding for Mario}.}
\label{fig:marioBinLevels}
\end{figure*}

\end{document}